\documentclass{egpubl}
\usepackage{sca2026}
 
\SpecialIssuePaper         %

\CGFccby

\usepackage[T1]{fontenc}
\usepackage{dfadobe}  

\usepackage{cite}  %
\BibtexOrBiblatex
\electronicVersion
\PrintedOrElectronic

\ifpdf \usepackage[pdftex]{graphicx} \pdfcompresslevel=9
\else \usepackage[dvips]{graphicx} \fi

\usepackage{egweblnk} 

\usepackage{graphicx}
\usepackage{amsmath}
\usepackage{amssymb}
\usepackage{booktabs}
\usepackage{upgreek}
\usepackage[dvipsnames]{xcolor}

\usepackage{caption}
\usepackage{subcaption}

\usepackage{tikz}
\usepackage{ulem}
\usepackage{array}
\usepackage{multirow}
\usepackage{makecell}

\definecolor{amethyst}{rgb}{0.6, 0.4, 0.8}
\definecolor{darkpastelgreen}{rgb}{0.01, 0.75, 0.24}
\definecolor{amber}{rgb}{1.0, 0.75, 0.0}
\definecolor{cadmiumorange}{rgb}{0.93, 0.53, 0.18}
\definecolor{lawngreen}{rgb}{0.49, 0.99, 0.0}
\definecolor{limegreen}{rgb}{0.2, 0.8, 0.2}
\definecolor{neongreen}{rgb}{0.22, 0.88, 0.08}
\definecolor{amethyst}{rgb}{0.6, 0.4, 0.8}
\definecolor{darkpastelgreen}{rgb}{0.01, 0.75, 0.24}
\definecolor{greenbest}{RGB}{88,137,15}
\definecolor{redworst}{RGB}{137,15,27}
\definecolor{royalazure}{rgb}{0.25, 0.41, 0.88}

\newcommand{\new}[1]{{{#1}}}
\newcommand{\conditioning}[1]{\textcolor{Mulberry}{{#1}}}
\newcommand{\reaction}[1]{\textcolor{ForestGreen}{{#1}}}

\newcommand{\pose}[1]{\mathbf{p}_{#1}}
\newcommand{\joints}[1]{\boldsymbol{\theta}_{#1}}
\newcommand{\rot}[1]{\mathbf{r}_{#1}}
\newcommand{\trans}[1]{\mathbf{t}_{#1}}
\newcommand{\predpose}[1]{\hat{\mathbf{p}}_{#1}}
\newcommand{\predjoints}[1]{\hat{\boldsymbol{\theta}}_{#1}}
\newcommand{\offjoints}[1]{\Delta\hat{\boldsymbol{\theta}}_{#1}}
\newcommand{\predrot}[1]{\hat{\mathbf{r}}_{#1}}
\newcommand{\predtrans}[1]{\hat{\mathbf{t}}_{#1}}
\newcommand{\encoder}{\mathcal{E}}
\newcommand{\decoder}{\mathcal{D}}
\newcommand{\capfix}{\mathcal{CF}}
\newcommand{\Emean}{\boldsymbol{\mu}}
\newcommand{\Esd}{\boldsymbol{\sigma}}
\newcommand{\Evar}{\boldsymbol{\sigma^2}}
\newcommand{\latent}{\mathbf{z}}

\newcommand{\lossVAE}{\mathcal{L}_{\text{VAE}}}
\newcommand{\lossrec}{\mathcal{L}_{\text{rec}}}
\newcommand{\lossKL}{\mathcal{L}_{\text{KL}}}

\newcommand{\lossCF}{\mathcal{L}_{\capfix}}
\newcommand{\losscol}{\mathcal{L}_{\text{col}}}
\newcommand{\lossdelta}{\mathcal{L}_{\offjoints{}}}
\newcommand{\lossjoints}{\mathcal{L}_{\text{joints}}}

\title[GNOCHI: Generative Neural mOdel for Close Human-Human Interactions]%
      {GNOCHI: Generative Neural mOdel for Close Human-Human Interactions}

\author[Gómez-Nogales et al.]
{\parbox{\textwidth}{\centering 
        Gonzalo Gómez-Nogales\thanks{Equal contribution}$^1$\orcid{0009-0000-1309-8383
        }~~~~~~~~~~~~~~
		Marc Comino-Trinidad\footnotemark[1]$^1$\orcid{0000-0001-5621-7565}~~~~~~~~~~~~~~   
        Andrés Casado-Elvira$^1$\orcid{0000-0003-0021-3930
        }~~~~~~~~~~~~~~
		Dan Casas\thanks{Work done prior to joining Amazon}$^1$\orcid{0000-0002-3664-089X}
        }
        \\
	{\parbox{\textwidth}{\centering $^1$Universidad Rey Juan Carlos, Madrid, Spain.}
	}
}

\begin{document}

\teaser{
    \vspace{-0.8cm}
    \includegraphics[width=0.8\linewidth]{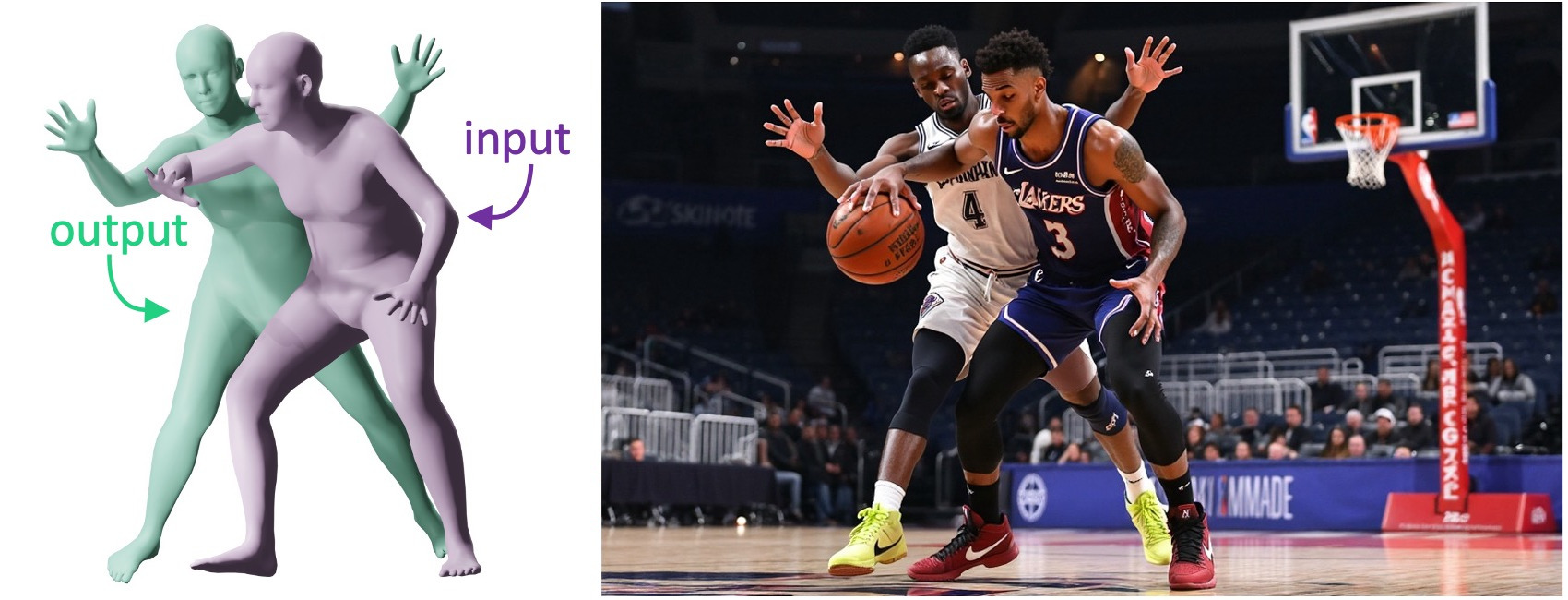}
    \centering
    \caption{Given an input 3D pose (\textit{i.e.}, the conditioning pose, in \conditioning{purple}), our generative model infers a 3D pose of a human in close interaction (\textit{i.e.}, the reaction pose, in \reaction{green}).
    This enables the conditional generation of 3D humans in close interaction, which can be used fine-grain control signal for image generative methods (right).}
    \label{fig:teaser}
}

\maketitle
\begin{abstract}
Creating realistic 3D human-human interactions in virtual environments is challenging due to the high degrees of freedom in human body and the need for physically accurate poses that do not collide with each other.
Traditional methods for human-human interaction are based on motion tracking or 3D body reconstruction, but lack generative capabilities.
Recent generative methods enable the synthesis of individual or interacting motions via text or image input, but generally fall short in modeling close interactions.
This paper introduces a novel generative model for close 3D human-human interactions using a conditional variational autoencoder (cVAE), which generates poses for one human conditioned on the pose of another, allowing for controlled and diverse interaction synthesis. 
To train our model, we address two underlying long-standing challenges in the field of human-human interaction: 
data scarcity, for which we propose an automated supervised data augmentation strategy that generates synthetic yet realistic interaction poses;
and collision awareness in generative approaches, for which we propose a self-supervised loss based on a collision resolution technique using volumetric proxies to ensure physically correct interactions.
We extensively evaluate the capabilities of our model, and demonstrate a wide variety of plausible and physically correct interactions, not possible to generate with current state-of-the-art methods.


\end{abstract}

\section{Introduction}
\label{sec:intro}
Creating life-like virtual 3D scenes is central to Computer Vision, Graphics, and VR, impacting dataset generation, human tracking, 3D scene understanding, and character animation. 
However, modeling and reconstructing how humans interact in everyday 3D scenes is a highly complex task due to the large number of degrees of freedom involved. 
Additionally, as humans, we are very sensitive to non-physically-correct virtual scene arrangements (\textit{e.g.}, interpenetrations, mesh collisions, or impossible configurations), hence, errors in modeling human interactions automatically produce significant visual discomfort.

Existing research on 3D interaction generally falls into three categories:
human-scene methods \cite{jiang2024scaling,hassan2021populating,guzov24ireplica,zhao2023synthesizing,zhao2022compositional,hassan2021stochastic,mueller2024hsfm}, for large-scale navigation; 
human-object methods \cite{jiang2023chairs,xie2022chore,christen2022dgrasp,hasson2019learning,ye2023ghop,taheri2021goal}, for manipulation and grasping;
and human-human models \cite{mueller2024buddi,ye2023slahmr,fieraru2020three,fieraru2021remips,maluleke2024synergy}. We focus on the latter, specifically addressing the synthesis of natural, close-contact interactions between two highly articulated bodies.

Unfortunately, most existing human-human interaction methods prioritize tracking or reconstruction \cite{fieraru2020three,fieraru2021remips,ye2023slahmr,guo2022expi,yin2023hi4d, sun2021romp,sun2022bev}. These approaches yield complex optimization or learning-based strategies for pose estimation, yet they lack generative capabilities and cannot sample from a learned distribution to synthesize unseen human-human interactions.
Recent works \cite{tanaka2023interaction,Chopin2023interaction,fan2024freemotion,shan2024multiperson,javed2025intermask,liang2024intergen} proposed generative models for human interaction, usually from text prompts, but they do not tackle the specific case of close interactions, nor do they incorporate robust collision awareness in their formulations.
BUDDI \cite{mueller2024buddi} is the notable exception, and it proposes a generative model capable of synthesizing realistic poses of two humans in close interaction.
Despite that BUDDI models both poses in a joint latent space, we show that conditional sampling can be achieved by incorporating the conditioning pose information throughout the diffusion process, similar to the in-painting strategies for conditional image synthesis.
However, our experiments show that this often compounds errors in close interactions, leading to physically non-plausible root placements or orientations, or unresolved colliding 3D meshes.

To mitigate this limitation, we propose a novel generative model for close 3D human-human interaction that is explicitly conditioned on one pose.
Implemented as a conditional variational autoencoder (cVAE) , our model uses the SMPL \cite{loper2015smpl} parameters of a \conditioning{conditioning} human to synthesize the body parameters of a \reaction{reactive agent}.
Since our model is probabilistic, we can generate many different poses given the same conditioning pose, which further increases the applicability of the method.
Additionally, we demonstrate that our model yields a continuous latent space that allows to seamlessly interpolate between reactive poses while keeping the same conditioning pose.

Manual rigging of two-person interactions is challenging; adjusting a single joint in a \conditioning{conditioning} pose typically requires the artist to manually re-balance dozens of degrees of freedom in the \reaction{reacting pose} to maintain contact and avoid penetration. Our model acts as a 3D pose copilot, offering a variety of semantically plausible reactive candidates that serve as a high-quality starting point for artists, effectively shortening the manual iterative loop.

To train our model, we tackle two long-standing problems in data-driven human-human interaction: data scarcity and collision awareness in data-driven models.
Regarding the first problem, despite the few recently introduced datasets for this task 
\cite{fieraru2020three,yin2023hi4d,xu2024interx,zheng2021deepmulticap,guo2022expi}, the coverage and diversity of \textit{close interactions} are still limited because these datasets rely on cumbersome motion capture, manual annotations, or expensive post-processing pipelines that do not scale well.
To mitigate this, we introduce a novel automated strategy to, for the first time, apply data augmentation techniques to 3D human-human interaction data, enabling to create synthetic and physically-plausible poses without requiring any manual work or new recordings.
Our key contribution is an individual-level stochastic process to manipulate the degrees of freedom of captured human-human motions while guaranteeing the correctness of the resulting interaction.
We achieve this by combining a learned human pose subspace \cite{vposer2019cvpr} that guarantees natural (yet unseen) individual poses, with an efficient collision resolve strategy that guarantees physically correct human-human poses in interactions without mesh intersections at the vertex level.

Our second contribution addresses a fundamental limitation in data-driven models: while our augmented dataset provides potentially unlimited collision-free human interaction samples, trained data-driven models (\textit{e.g.}, cVAE) cannot inherently guarantee collision-free states for unseen samples. 
We tackle this limitation by incorporating an additional decoder trained with a collision-aware novel self-supervised loss based on highly-efficient distance computation across volumetric proxies, ensuring physically plausible interactions during generation.

We demonstrate that our generative approach can synthesize pose-conditioned human-human interactions for a wide variety of scenarios, including sports, dancing, fighting, and social communications. 
Additionally, as a forward generative model, our approach can be used to synthesize reactive interaction partners or refine tracking results from current human mesh recovery methods \cite{sun2022bev,mueller2024buddi}, providing a richer variety of plausible responses and enhanced controllability of the final output.

\section{Related Work}
\label{sec:related-work}

\subsection{Datasets}
Obtaining accurate human-human body data is inherently challenging. Various methods exist to transform real-world data into parameters suitable for computational processing~\cite{fieraru2020three, yin2023hi4d, xu2024interx, guo2022expi, zheng2021deepmulticap}.
Fieraru \textit{et al}.~\cite{fieraru2020three} use 2D annotated data to predict 3D contact, resulting in two datasets: CHI3D based on captured data, and FlickrCI3D based on Flickr images.
Using an alternating optimization scheme, Yin \textit{et al.}~\cite{yin2023hi4d} segment the 3D surface of two avatars and fit SMPL \cite{loper2015smpl} model parameters to them.
Inter-X \cite{xu2024interx} is the largest labeled human interaction dataset, featuring around 11,000 sequences and over 8.1 million frames. It provides detailed descriptions, action categories, and annotations for interaction order.
Zheng \textit{et al.}~\cite{zheng2021deepmulticap} addresses self-occlusions through multiview capture. Their MultiHuman dataset consists of 150 static scenes with different levels of occlusions and ground truth 3D human models.
To predict subsequent dance moves, Guo \textit{et al.}~\cite{guo2022expi} created a dataset of dancers in extreme poses, capturing dynamic and complex movements for motion prediction tasks.
In this work, we leverage the Hi4D \cite{yin2023hi4d} dataset which, despite the impressive quality of the reconstructed human interactions, still contains a limited amount of in-contact interactions. 
We apply our novel pose augmentation technique to the subset of in-contact interactions from Hi4D, generating 10$\times$ samples that we use to train our model.

\subsection{Multi-Character Animation}
Generating human-human interaction is a classic problem in computer animation. Early works use optimization-based frameworks \cite{liu2006composition} and spatial relationship descriptors \cite{ho2010spatial} to adapt captured motions to new interactions. 
Later, patch-based methods \cite{kim2012tiling} tiled or concatenated pre-recorded interactions to generate variety, while data-driven generate-and-rank systems \cite{won2014generating} composed and ranked scenes from motion-capture libraries. While effective, these \textbf{data-driven} approaches required dense motion libraries.

More recently, there is increasing use of physics-based simulation  \cite{zhang2023simulation}, alongside strategic planning methods that explicitly model competitive and collaborative goals \cite{shum2007simulating, shum2012simulating}. 
Other recent works aim at learning time-dependent dynamic human interactions \cite{ghosh2025duetgen, liu2025ponimator}.
Instead, we focus our efforts on learning a continuous space generative manifold of static interaction, which is a fundamental problem in keyframe animation. 
\subsection{Reconstruction of People in Close Interaction}

Many previous works~\cite{mueller2024buddi, mueller2021tuch, ye2023slahmr, shuai2023reconstructing, multiply,zheng2021deepmulticap} aim to accurately reconstruct interacting humans. 
SLAHMR~\cite{ye2023slahmr} reconstructs sequences involving multiple people by accurately decoupling human and camera motion. 
The underlying model uses an optimization method to achieve this decoupling, allowing for the reconstruction of complex scenes.
However, while the method is effective in handling multiple people, it falls short in accurately recovering close interactions between them, highlighting an area for potential improvement in future research.
Shuai et al.~\cite{shuai2023reconstructing} tackle the challenge of reconstructing close human interactions from multiple views.
Their approach integrates a learning-based pose estimation component that uses multi-view 2D keypoint heatmaps as input and reconstructs the pose of each individual using a 3D conditional volumetric network.
This method significantly improves the accuracy of pose reconstruction in crowded scenes.
Other works~\cite{mueller2021tuch,mihajlovic2022coap} address the challenge of modeling single avatars with accurate self-contact using collision-specific terms. 
Rather than focusing on interactions between multiple avatars, these approaches emphasize the importance of self-contact, ensuring that the reconstructed avatars maintain realistic and physically plausible poses.

Closest to our work, BUDDI~\cite{mueller2024buddi} focuses on modeling two interactive avatars using a diffusion model, which learns the joint distribution over the poses of two people in close social interaction.
The model is particularly notable for its ability to generate pairs of 3D avatars that exhibit realistic and close interactions, making it highly relevant for applications requiring detailed social dynamics.
Our objective, however, is to explicitly generate an interacting avatar given the pose of another one. Despite that BUDDI can be adapted to pose-conditional sampling, our method yields more accurate contact for close interactions.

\subsection{Collision Aware Reconstruction}

Some works~\cite{fieraru2021remips, ugrinovic2024multiphys, mueller2024buddi} put a special effort into avoiding collisions and mesh intersections between avatars. 
MultiPhys~\cite{ugrinovic2024multiphys} integrates a physics simulator in an autoregressive manner to refine kinematic estimates and ensure physical compliance. The simulator captures coherent spatial placement between individuals and eliminates penetration issues. The pipeline involves feeding motion estimated by a kinematic-based method into the physics simulator, which then corrects any collisions or penetrations by adjusting the avatars' positions and poses.
BUDDI~\cite{mueller2024buddi} employs a collision-checking strategy that primarily relies on a coarse approximation of the avatar mesh. 
This simplified representation facilitates efficient collision detection and resolution by reducing computational complexity.
The model uses ground-truth contact annotations to fit SMPL-X to images, ensuring that avatars maintain realistic social distances and avoid collisions.
Similarly, REMIPS~\cite{fieraru2021remips} also uses a coarse mesh approximation by applying decimation operators to detect and solve self- and interpenetration-collisions. 
This model incorporates self-contact and interaction-contact losses directly into the learning process, ensuring that the reconstructed avatars do not intersect with each other or the environment.
The use of self-supervised losses allows the model to generalize well to in-the-wild scenarios, maintaining physical consistency in multi-person 3D reconstructions.
Instead of using a coarse version of SMPL, we resolve collisions at the vertex level to create a large yet vertex-accurate dataset.
At train time, we use highly-efficient volumetric capsules to compute potential intersections.

\section{Close Contact Dataset}
\label{sec:dataset}
\begin{figure*}[!ht]
\vspace{-0.3cm}
    \centering
    \includegraphics[width=\textwidth]{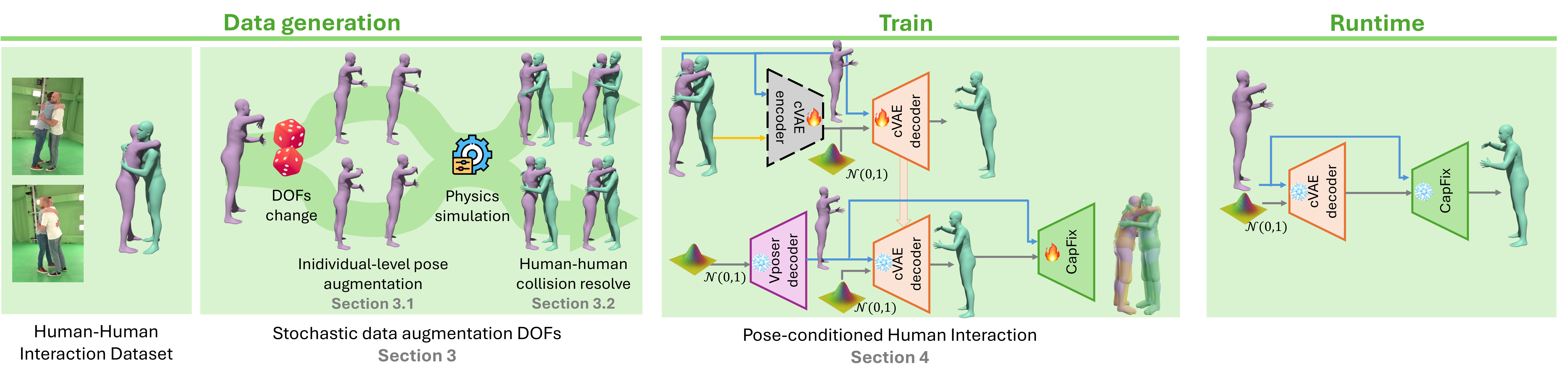}
\vspace{-0.8cm}    
    \caption{From an existing dataset \cite{yin2023hi4d}~(left), we apply individual-level noise at each body joint (Section~\ref{sec:individual-pose-aug}), and jointly resolve the collisions using a physical simulator (Section~\ref{sec:collision-solving}) to create an augmented dataset.
    At train time, our model learns to infer a \reaction{reacting} avatar given a \conditioning{conditioning} avatar (center top, Section~\ref{sec:cVAE}). 
    Then, a second model is trained to fix collisions between two interacting avatars (center bottom, Section~\ref{sec:capfix}). At runtime, we can sample a reacting collision-free avatar given a conditioning pose.
    }
    \label{fig:pipeline}
\vspace{-0.4cm}
\end{figure*}

Capturing and reconstructing 3D humans in close interaction is a tedious task due to the unavoidable occlusions that prevent observing the full human surface, even in multi-camera environments. 
Recent methods \cite{yin2023hi4d} have proposed complex pipelines for accurately reconstructing close interactions in 3D; however, they are expensive and do not scale well.

In data-driven methods, to circumvent the scarcity of training data, data augmentation is a common strategy to increase the number of dataset samples, and it plays a fundamental part in many image-based machine learning methods~\cite{mueller2018ganerated}.
However, augmenting 3D human-human pose data is challenging due to the many nuances that make human interaction realistic.

To this end, we propose a novel approach to build a pseudo-synthetic dataset of human-human interactions.
Starting from a subset of the Hi4D~\cite{yin2023hi4d} dataset, we first apply an individual-level data-augmentation strategy (Section \ref{sec:individual-pose-aug}), and then resolve vertex-level human-human collisions using a physics-based simulation strategy (Section \ref{sec:collision-solving}).
This enables us to obtain a dataset of unseen yet physically-correct \textit{close} (\textit{i.e.} in-contact) human-human interaction poses, which we later use to train a pose-conditioned generative model in Section~\ref{sec:pose-conditioned-interaction}.

\subsection{Individual-level Pose Augmentation}
\label{sec:individual-pose-aug}
We begin our dataset creation by selecting a subset of frames from the Hi4D~\cite{yin2023hi4d} dataset that contain contact interactions, namely poses where both interacting meshes are at less than 1 cm from each other.
For each body pose of each selected frame, we first apply random noise to the rotation of specific body joints, and then project the resulting pose into a subspace of feasible poses using the VPoser~\cite{vposer2019cvpr} autoencoder, ensuring the final pose is anatomically plausible and introducing additional noise.  

To generate random rotations, we employ the axis-angle representation for its intuitive and mathematically simple nature, \new{the full details of which are provided in the Supplementary Material.}
We carefully define a specific rotation distribution for each joint.
For example, knees only rotate around the horizontal axis, while for other joints, such as hips, the distribution is weighted for all three spatial axes.
For each joint, we define a set of weights $w_x, w_y, w_z$ and a maximum angle $\alpha_M$. We then sample a vector
\begin{equation}
        \overrightarrow{v_r} = \frac{w_x r_x + w_y r_y + w_z r_z}{\| w_x r_x + w_y r_y + w_z r_z \|},
\end{equation}
where $r_x, r_y, r_z \sim \mathcal{U}_{[-0.5, 0.5]}$.
Finally, the angle $ \alpha $ is similarly obtained 
\begin{gather}
	\alpha = \alpha_M w_\alpha, 
\end{gather}
where $w_\alpha \sim \mathcal{U}_{[-1, 1]}$.
\new{This symmetric uniform distribution effectively places the rest pose at the center of each joint's maximum sweep angle.} The final random axis-angle is obtained by multiplying $\alpha \cdot \overrightarrow{v_r}$. To apply the noise to the joint, we convert both the joint and the random axis-angle rotation into matrix rotation representation and multiply them.
Setting two weights $w_i$ to zero means selecting only one axis for rotation. If one or none of the weights are zero, controlling the random rotation becomes more challenging. 
\new{Because our perturbation is based strictly on these maximum sweep bounds, the naive random rotation can potentially result in an anatomically unfeasible configuration, such as joints bending backward. To explicitly address this lack of joint constraints, we pass the randomized pose through the VPoser autoencoder \cite{vposer2019cvpr}. VPoser projects the perturbed pose back into a learned subspace of human poses, ensuring that the final output is always anatomically valid regardless of the initial random noise.}

\subsection{Human-Human Collision Resolve}
\label{sec:collision-solving}

Our individual-level pose augmentation described in Section~\ref{sec:individual-pose-aug} is a naive strategy that can lead to new, albeit physically incorrect, poses that may intersect with nearby individuals. To mitigate this, we employ a highly efficient strategy to accurately resolve collisions between interacting humans, which is crucial in building our large dataset.

Approximating humans with coarse volumetric proxies is a common technique to resolve collisions because it allows for fast inter-proxy distance computation. However, in the specific case of close human-human contact, this strategy produces unrealistic interactions due to the coarse approximation of the human surface. To circumvent this problem, we leverage a pre-existing solver that implements a method based on accurate Signed Distance Fields (SDFs) to resolve human-human intersections.

\new{It is important to note that the initial SMPL fits of the Hi4D data are not inherently penetration-free.} For each pair of humans interacting in the original dataset, the solver computes the SDF of each human in the scene using OpenVDB. This effectively creates two grids (one for each human) that can be efficiently used to query distances to human surfaces from any 3D world coordinate. The solver then checks the distance of all surface points of one human (i.e., all the vertices) against the other human in the frame, and vice versa.
For each surface vertex that is \textit{inside} the other body \new{or its own body}, a penalty force $E_\text{col}(\updelta) = \frac{1}{2} \, k_{\updelta} \, \updelta^2$ is applied in the direction of the closest surface point normal. Here, $\updelta$ represents the distance from the vertex position to the surface of the penetrated body, and this force is applied to the bones that influence the vertex.

{
\newcommand{\colwidth}{0.17}
\newcommand{\cropL}{200mm}
\newcommand{\cropR}{200mm}
\newcommand{\cropB}{36mm}
\newcommand{\cropT}{36mm}
	
\begin{figure}
\vspace{-0.4cm}
\centering
\begin{tabular}{ccccc}
    \rotatebox[origin=tl]{90}{\footnotesize{\centering\hspace{0.6cm}{Original sample}}}
    & 
    \begin{subfigure}{\colwidth \linewidth}
        \includegraphics[width=\textwidth]{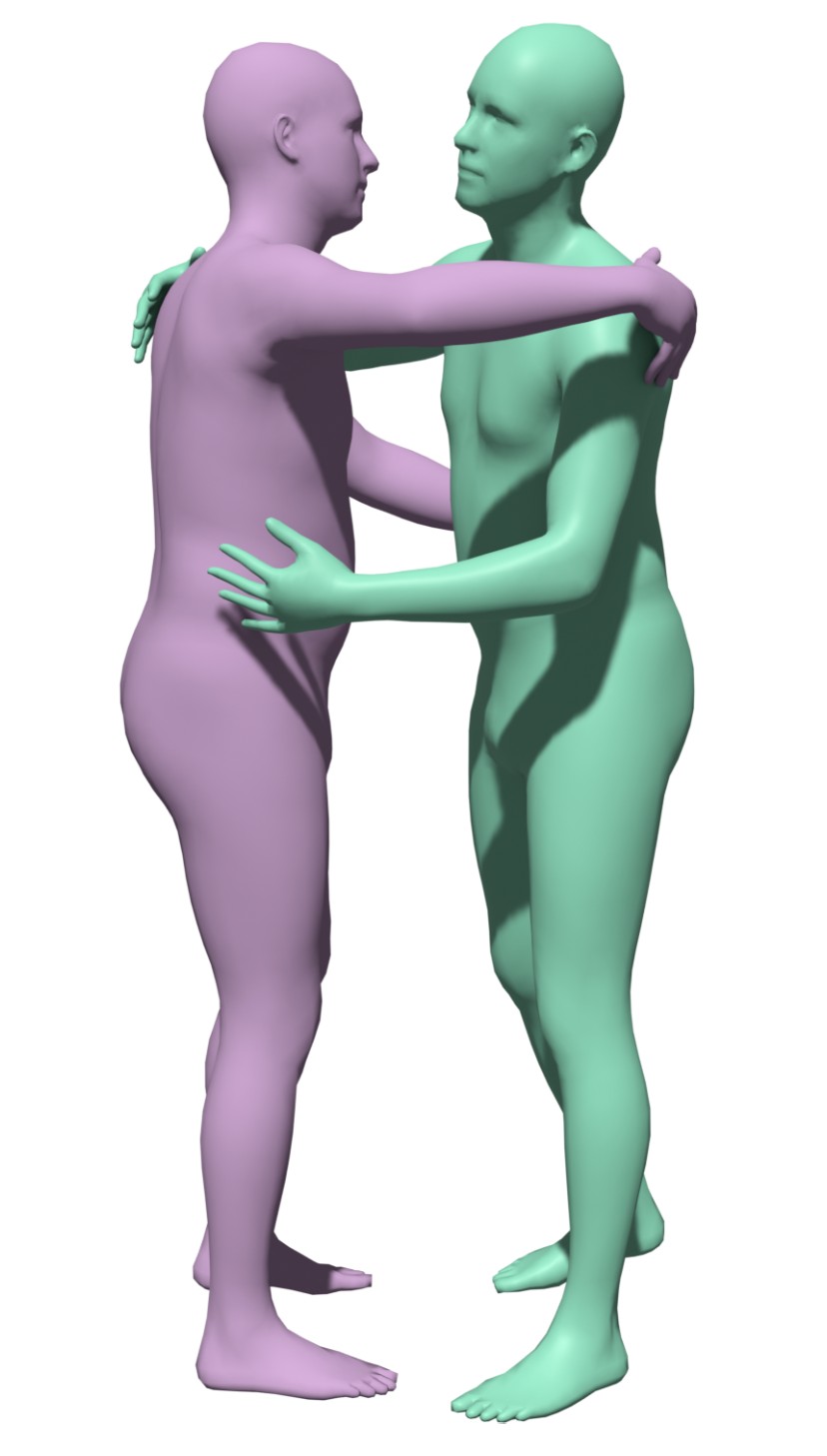}
        \caption{}
        \label{fig:subfig:sim:original}
    \end{subfigure}
    &
    \rotatebox[origin=tl]{90}{\footnotesize{\begin{tabular}{c}Augmented samples\\\vspace{4pt}(Section \ref{sec:individual-pose-aug})\end{tabular}}}
    & 
    \begin{subfigure}{\colwidth \linewidth}
        \includegraphics[width=\textwidth]{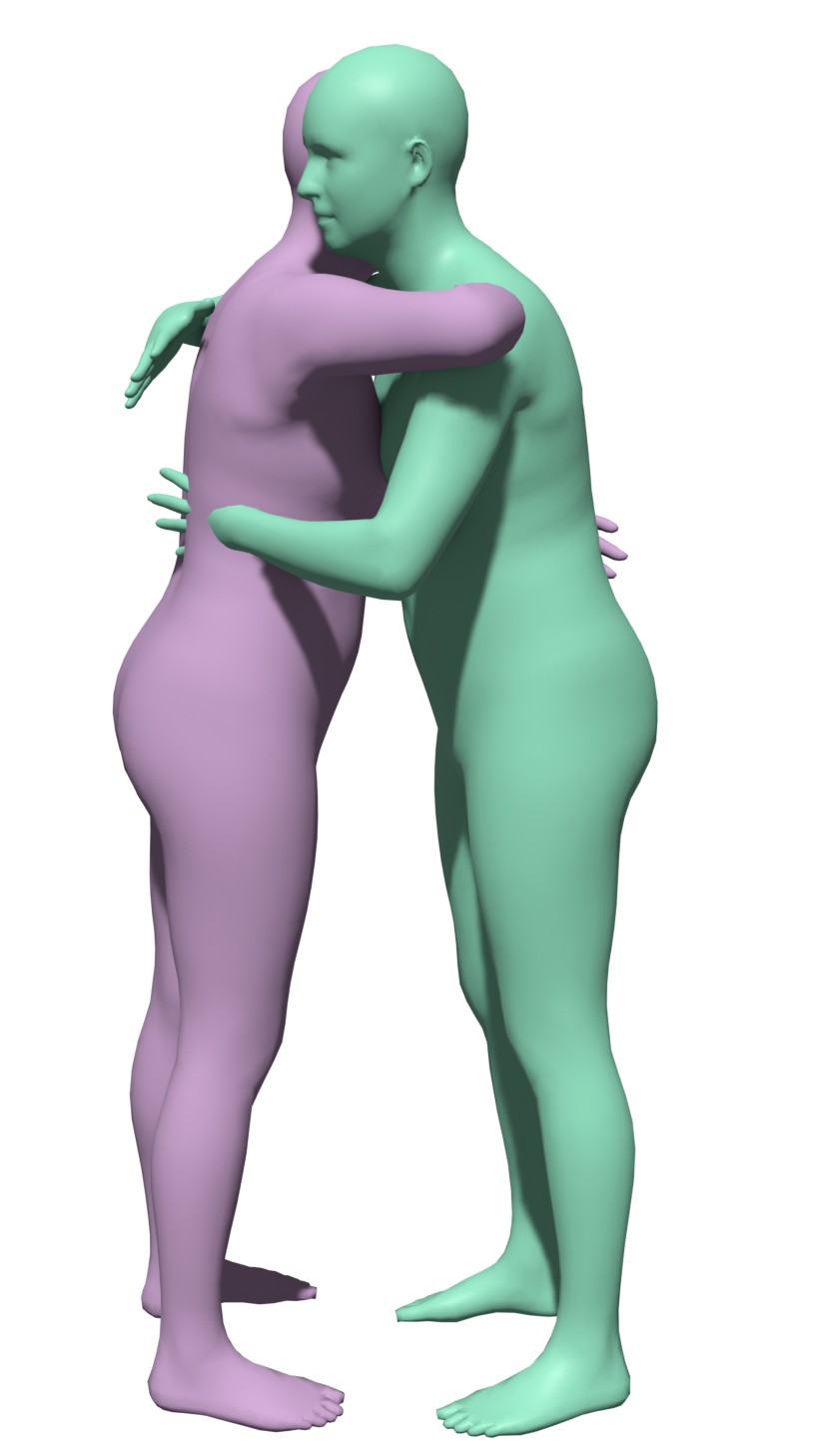}
        \caption{}
        \label{fig:subfig:sim:augmented1}
    \end{subfigure}
    \begin{subfigure}{\colwidth \linewidth}
        \includegraphics[width=\textwidth]{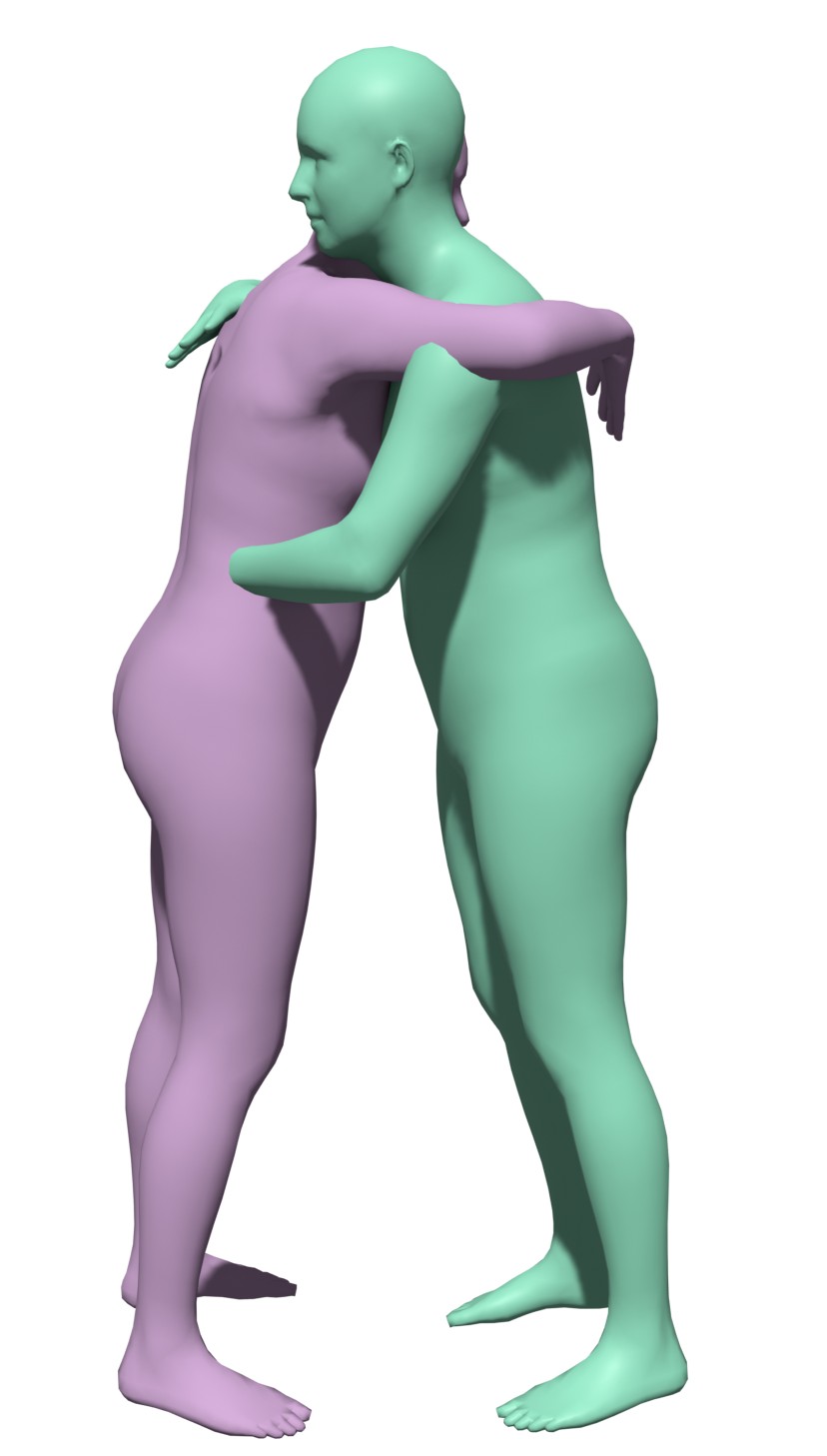}
        \caption{}
        \label{fig:subfig:sim:augmented2}
    \end{subfigure}
    \begin{subfigure}{\colwidth \linewidth}
        \includegraphics[width=\textwidth]{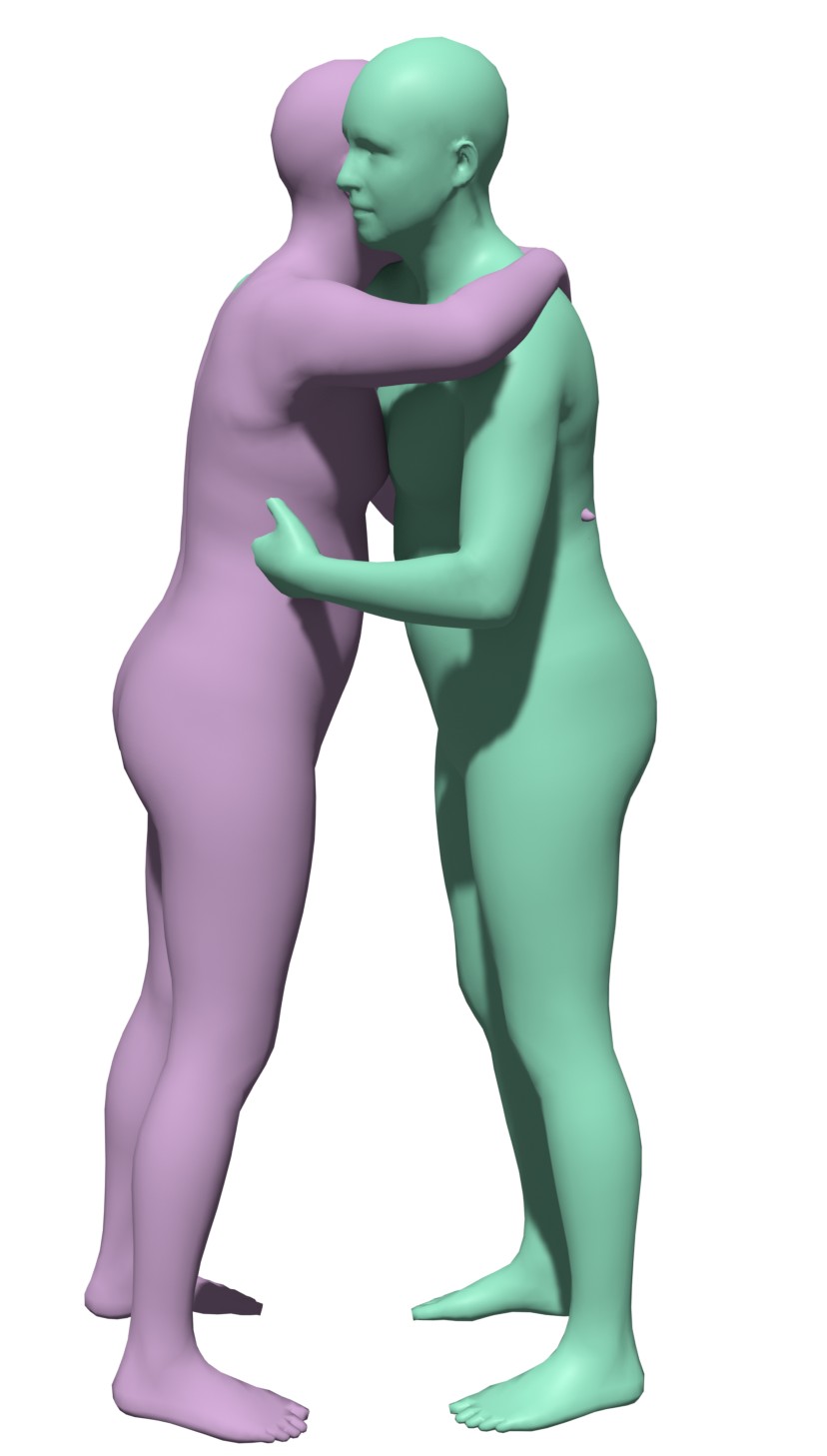}
        \caption{}
        \label{fig:subfig:sim:augmented3}
    \end{subfigure}
    \\
    &
    &
        \rotatebox[origin=tl]{90}{\footnotesize{\begin{tabular}{c}Corrected samples\\\vspace{4pt}(Section \ref{sec:collision-solving})\end{tabular}}}
    &
    \begin{subfigure}{\colwidth \linewidth}
        \includegraphics[width=\textwidth]{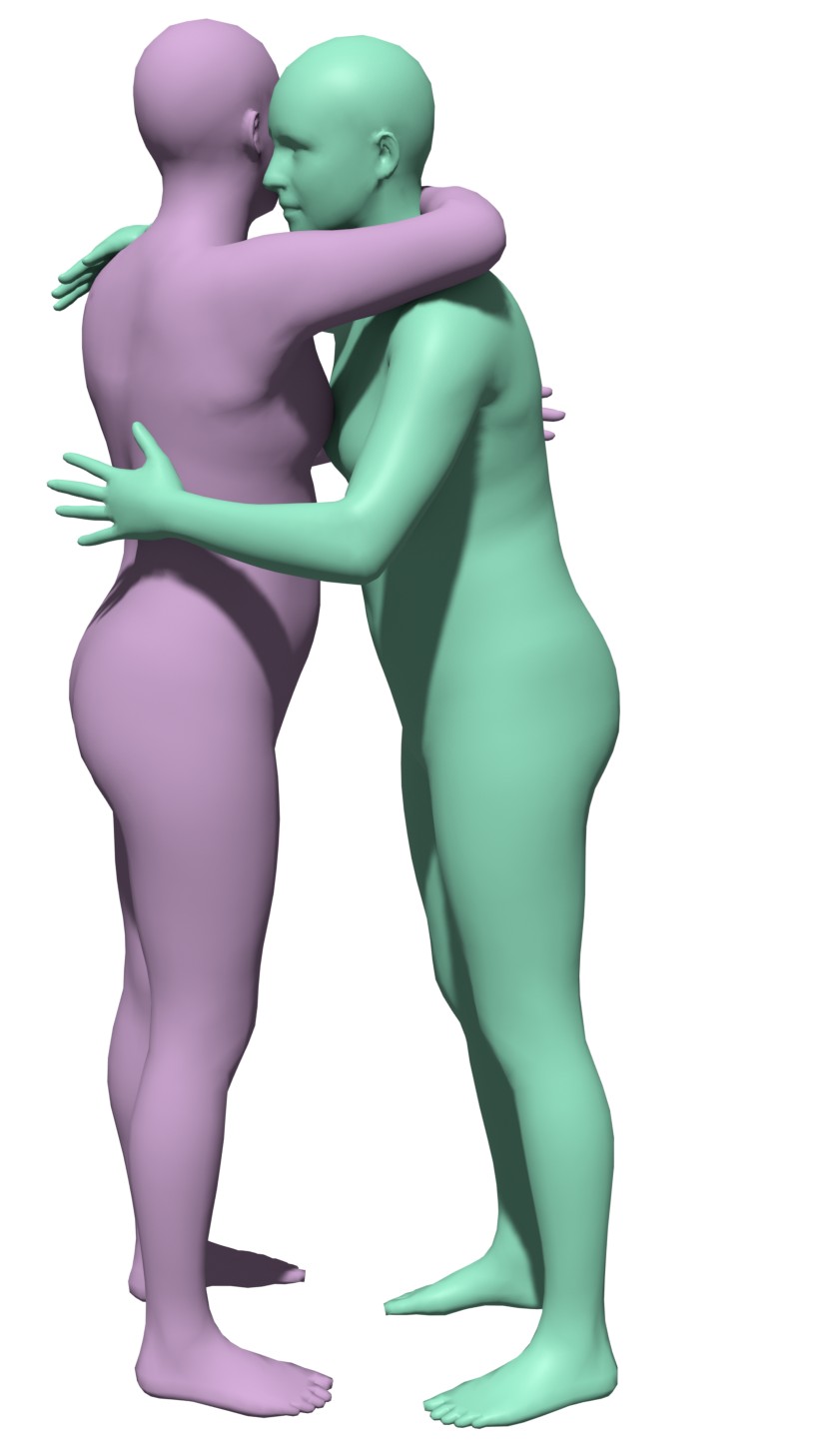}
        \caption{}
        \label{fig:subfig:sim:simulated1}
    \end{subfigure}
    \begin{subfigure}{\colwidth \linewidth}
        \includegraphics[width=\textwidth]{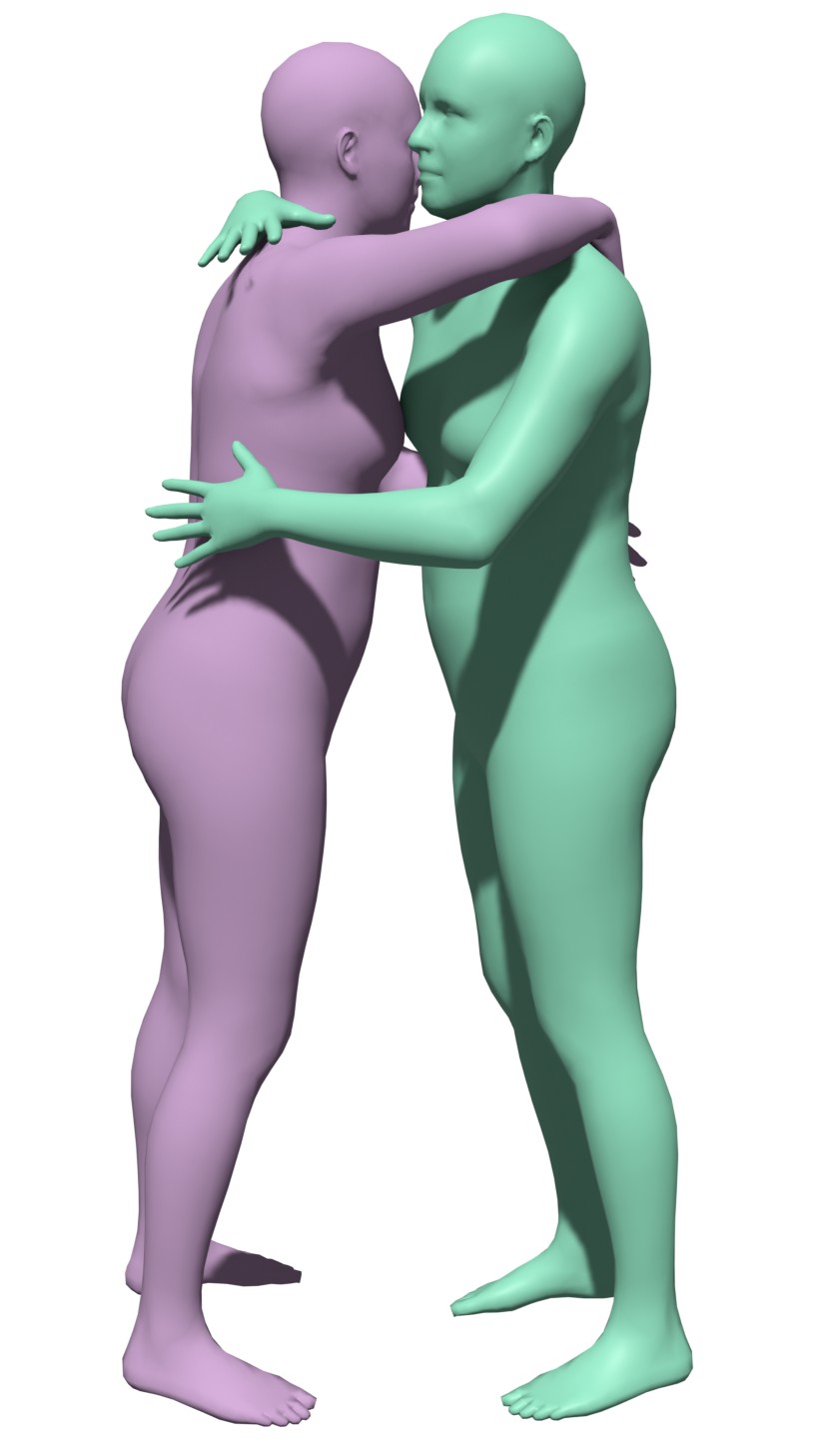}
        \caption{}
        \label{fig:subfig:sim:simulated2}
    \end{subfigure}
    \begin{subfigure}{\colwidth \linewidth}
        \includegraphics[width=\textwidth]{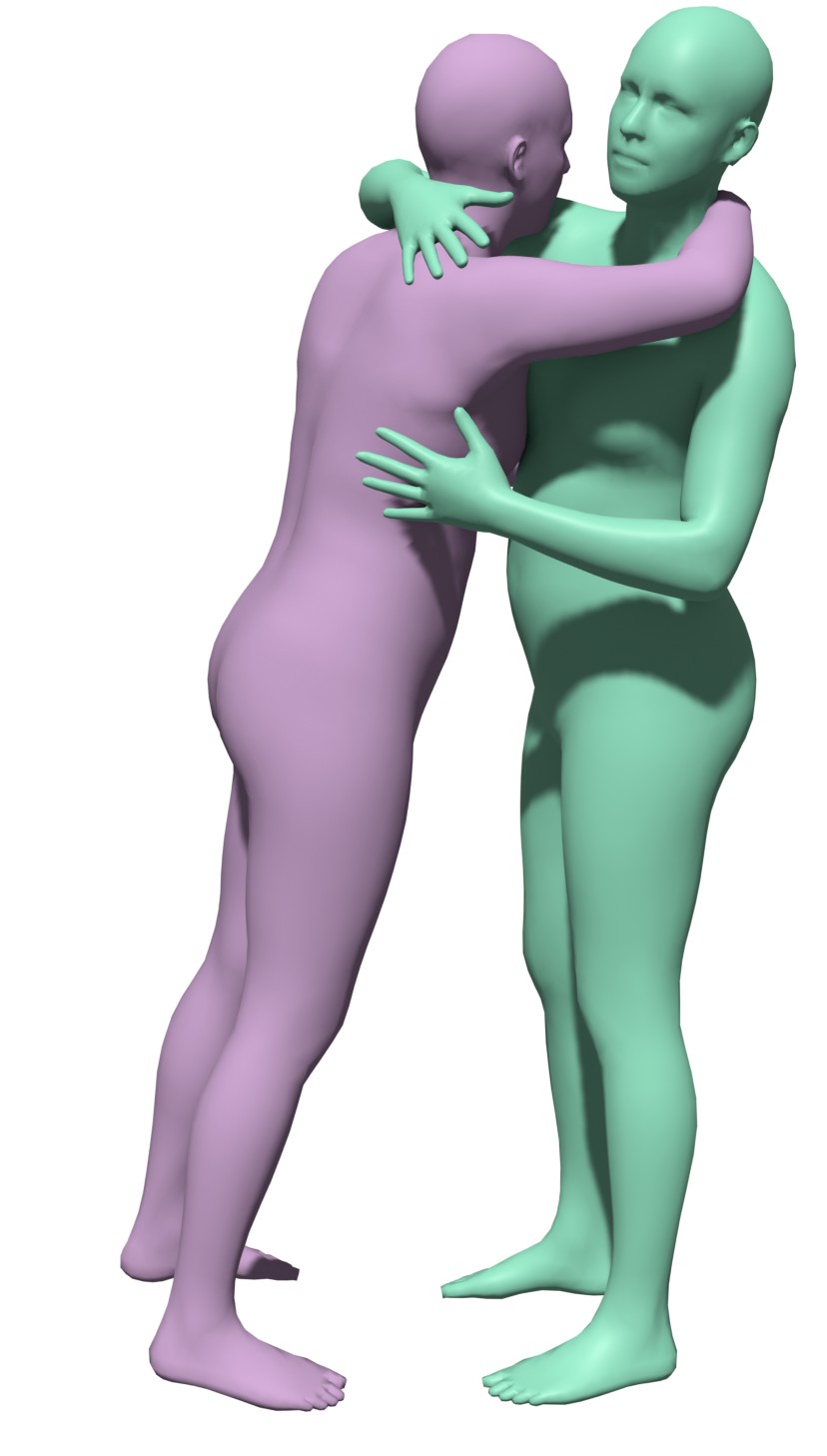}
        \caption{}
        \label{fig:subfig:sim:simulated3}
    \end{subfigure}
\end{tabular}
\vspace{-0.5cm}
    \caption{Data augmentation steps. We add noise to the original poses (\ref{fig:subfig:sim:original}), yielding augmented poses (\ref{fig:subfig:sim:augmented1}, \ref{fig:subfig:sim:augmented2}, and \ref{fig:subfig:sim:augmented3}) which may contain collisions. Our physics solver fixes these, yielding physically-correct interactions (\ref{fig:subfig:sim:simulated1}, \ref{fig:subfig:sim:simulated2}, and \ref{fig:subfig:sim:simulated3}).
    }
    \label{fig:fix-simulator}
\vspace{-0.5cm}
\end{figure}
}

We integrate the penalty equations using the popular optimization formulation of backward Euler \cite{gast2015optimization, kane2000variational}.
To solve it, we use Newton's method with analytical computation of gradients and Hessians. 
Just one Newton iteration works well for our disentanglement simulations. 
Each Newton iteration yields a sparse linear system of size $6 \times 24 \times 2$, where $2$ is the number of humans interacting, $24$ is the number of bones per body, and $6$ is the number of degrees of freedom per bone.
We solve this linear system using the conjugate gradient method. 

We present results of our collision resolution approach in Figure~\ref{fig:fix-simulator}. Notably, our method effectively eliminates mesh interpenetrations while preserving close contact between the avatars.

\subsection{Final Dataset Details}
For each avatar pair in the Hi4D dataset, we generated 10 variations. Because the random noise introduced during pose augmentation can cause the avatars to drift apart, we filtered out any samples that no longer met our contact threshold, resulting in a final dataset comprising 55,897 examples. We later merged this dataset with the 5,744 poses from Hi4D interaction scans, yielding a full augmented dataset of 61,641 poses. 
\new{To prevent any information leakage between sets, we implemented a sequence-heldout train-test split rather than a random per-sample partition. Augmented variants are strictly assigned to the same split as their base sequence. Detailed sample distributions for the training and testing sets are provided in Section~\ref{sec:ablation}.}

\section{Pose-Conditioned Human Interactions}
\label{sec:pose-conditioned-interaction}

Our goal is to model physically-correct and realistic 3D human poses in close interaction.
Specifically, we address the problem of \textit{pose-conditioned pose} inference, where we are given a pose (the \conditioning{conditioning} pose) and aim to infer poses (the \reaction{reaction} poses) that realistically interact with the input.

To this end, we first introduce the representation used to describe poses (Section~\ref{sec:representations}). We then present our compact yet expressive generative latent space tailored for humans in interaction (Section \ref{sec:cVAE}), which we train using a conditional Variational Autoencoder (cVAE) that leverages the pose-augmented human-human dataset introduced in Section~\ref{sec:dataset}.
Next, we describe a self-supervised strategy (Section~\ref{sec:capfix}) to learn to resolve the residual human-human collisions in our latent space.
At inference time (Section~\ref{sec:human-interaction-inference}), our method enables the generation of human-human 3D pose interactions conditioned on the pose of one of the avatars.
Figure~\ref{fig:pipeline} depicts the detailed architecture of this pipeline. The full architecture details of the model can be found in the Supplementary Material.
\subsection{Pose Representation}
\label{sec:representations}

We define the pose of an avatar $X$ as
\begin{equation}
    \pose{X} = \left\lbrace \joints{X}, \rot{X}, \trans{X} \right\rbrace,
\end{equation}
where $\joints{X}$ are the joint rotations, $\rot{X}$ is the global rotation, and $\trans{X}$ is the global translation.
We use the original SMPL~\cite{loper2015smpl} 23 joints but, to facilitate training, we leverage the continuous 6D representation for each joint rotation by Zhou \textit{et al.}~\cite{zhou2019continuity}.
Consequently, the joint rotations $\joints{X}$ is a tensor in $\mathbb{R}^{23 \times 3 \times 2}$ and the global rotation matrix $\rot{X}$ is $\mathbb{R}^{3 \times 2}$. The global translation $\trans{X}$ is represented by a 3D position vector $\trans{X}$ in $\mathbb{R}^{3}$.

Each sample consists of two poses: a \conditioning{conditioning pose} $\pose{C}$ and a \reaction{reaction pose} $\pose{R}$.
We use pose coordinates relative to the conditioning avatar $\pose{C}$, therefore, translation $\trans{C}$ and the rotation $\rot{C}$ of avatar $C$ are the origin of coordinates and a unit matrix, respectively, and $\pose{C}$ can be defined by the  $\joints{C}$ alone.

\subsection{Latent Space for Human Interaction}
\label{sec:cVAE}

\begin{figure}[t]
    \centering
    \includegraphics[width=1.0\linewidth, trim={12 0pt 60pt 10pt}]{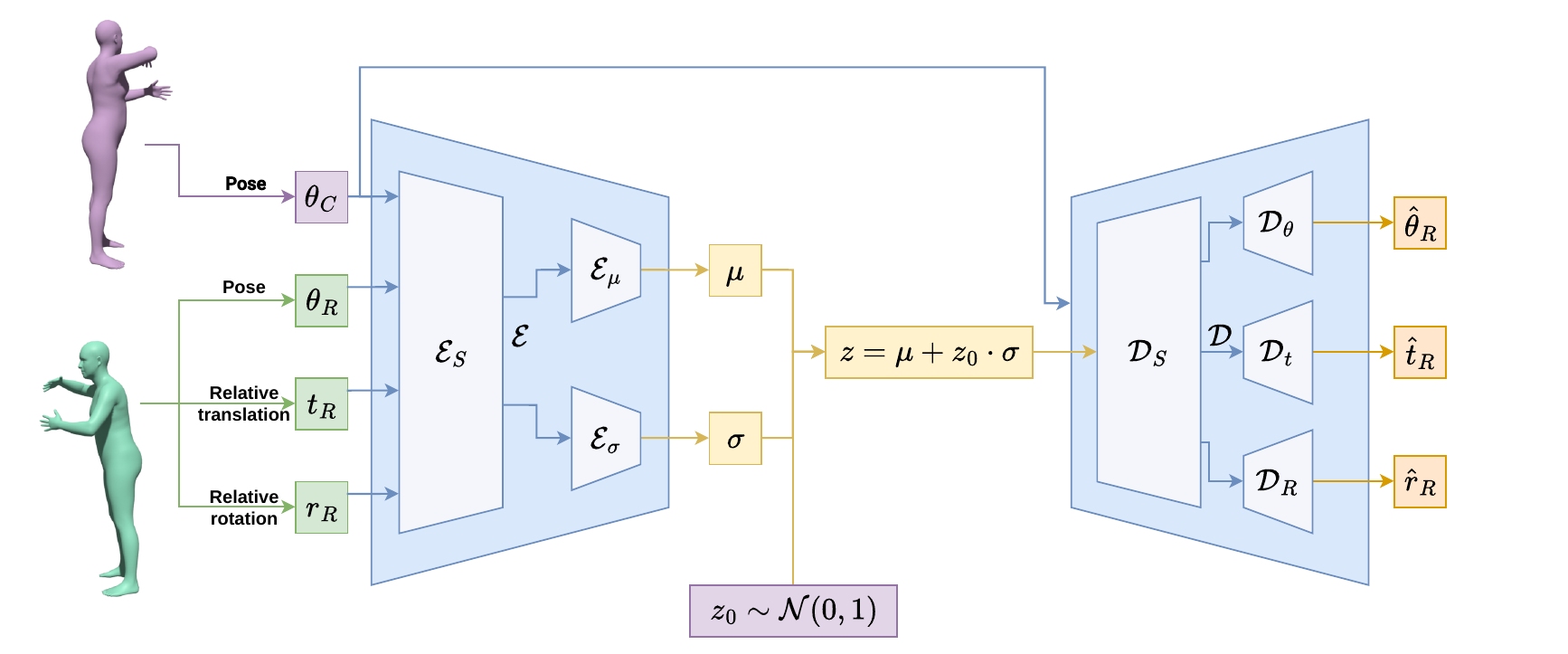}
    \caption{Detailed cVAE model architecture. The input consists of the conditioning pose $\joints{C}$ and the reaction pose $\joints{R}, \trans{R}, \rot{R}$, where $\trans{R}$ and $\rot{R}$ are the relative translation and rotation of the \reaction{reacting avatar} with respect to the \conditioning{conditioning avatar}. These four tensors are fed into the encoder, which has shared weights $\encoder_{S}$ and specialized weights $\encoder_{\Emean}, \encoder_{\Esd}$. The interaction is then encoded into two tensors, ${\Emean}$ and ${\Esd}$, that define a normal distribution. 
    To sample a \reaction{reacting pose} from the distribution, we first sample $\latent_0$ from the standard normal distribution and then apply the reparameterization trick to obtain the sample $\latent$. To decode this sample, the decoding process uses shared weights $\decoder_{S}$ and specialized weights $\decoder_{\joints{}}, \decoder_{\trans{}}, \decoder_{\rot{}}$, which return the corresponding body parameters. Note that the \conditioning{conditioning pose} $\joints{C}$ is also an input to the decoder. Different $\latent$ values will generate different reactions for a given pose $\joints{C}$. Similarly, the same $\latent$ value will generate different avatars depending on the pose $\joints{C}$ it has to react to.}
    \label{fig:cvae-architecture}
    \vspace{-0.3cm}
\end{figure}

To learn a generative model for human-human interaction, we use a conditional Variational Autoencoder (cVAE)~\cite{sohn2015learning} consisting of an encoder $\encoder$ and decoder $\decoder$ implemented as multilayer perceptron (MLP) networks.
We use a custom cVAE architecture and train the encoder-decoder network end to end. 
The encoder $\encoder: (\mathbb{R}^{23 \times 3 \times 2}, \mathbb{R}^{23 \times 3 \times 2}, \mathbb{R}^{3}, \mathbb{R}^{3 \times 2}) \to (\mathbb{R}^{128}, \mathbb{R}^{128})$ takes as input $\lbrace \joints{C}, \joints{R}, \trans{R}, \rot{R} \rbrace$ and predicts the latent variables that define the normal distribution (\textit{i.e.}, mean $\Emean$ and variance parameters $\Evar$) of the latent space.
The decoder $\decoder: (\mathbb{R}^{23 \times 3 \times 2}, \mathbb{R}^{128}) \to (\mathbb{R}^{23 \times 3 \times 2}, \mathbb{R}^{3 \times 2}, \mathbb{R}^{3})$ takes as input a concatenation of a sample $\latent = \Emean + \latent_0 \cdot \Esd \quad | \quad \latent_0 \sim \mathcal{N}(0, 1)$ and the \conditioning{conditioning pose} $\joints{C}$, and it decodes the \reaction{reaction pose} $\predpose{R} = \lbrace \predjoints{R}, \predrot{R}, \predtrans{R} \rbrace$.
As depicted in Figure~\ref{fig:cvae-architecture}, the encoder $\encoder$ and decoder $\decoder$ are implemented using shared and specialized components. For example, the decoder $\decoder_{S}: \mathbb{R}^{128} \to \mathbb{R}^{128}$ first converts our latent representation $\latent$ into a shared decoded pose representation. Then, we extract the joint rotations $\predjoints{R}$, global translation $\predtrans{R}$, and global rotation $\predrot{R}$ using task-specific decoder networks $\decoder_{\joints{}}$, $\decoder_{\trans{}}$, and $\decoder_{\rot{}}$ as follows:
\begin{equation}
    \predjoints{R} = \decoder_{\joints{}}(\decoder_{S}( \left\lbrace \joints{C}, \latent \right\rbrace ))
\end{equation}
\begin{equation}
    \predrot{R} = \decoder_{\rot{}}(\decoder_{S}( \left\lbrace \joints{C}, \latent \right\rbrace ))
\end{equation}
\begin{equation}
    \predtrans{R} = \decoder_{\trans{}}(\decoder_{S}( \left\lbrace \joints{C}, \latent \right\rbrace ))
\end{equation}
where $\lbrace \joints{C}, \latent \rbrace \in \mathbb{R}^{266}$ denotes the concatenation and flattening of the tensors.

We train our cVAE network end-to-end using the loss
\begin{equation}
    \lossVAE =  \lambda_\text{KL}\lossKL + \lambda_\text{rec}\lossrec
\end{equation}
where $\lossKL$ is the Kullback–Leibler divergence term that enforces the latent spaces to follow a Gaussian distribution, and $\lossrec$ is a body surface term defined as
\begin{equation}
    \lossrec = \begin{Vmatrix} f_\text{SMPL}(\pose{R}) - f_\text{SMPL}(\predpose{R}) \end{Vmatrix}_2^2 
\end{equation}
that enforces the output \reaction{reaction pose} body parameters $\predpose{R}$ to generate SMPL body surface vertices close to ground truth vertices. \new{We empirically set the loss weights to $\lambda_\text{KL} = 1 \times 10^{-3}$ and $\lambda_\text{rec} = 7.5$. The network is trained using the Adam optimizer with a learning rate of $1 \times 10^{-4}$ and a batch size of 50.}

Once our cVAE is trained, the decoder $\decoder$ is be able to generate realistic \reaction{reactive poses} $\pose{R}$ for any arbitrary \conditioning{conditioning pose} $\pose{C}$.

\subsection{Learning to Resolve Human-Human Contact}
\label{sec:capfix}
\begin{figure}[t]
\vspace{-0.2cm}
    \centering
    \includegraphics[width=1\linewidth]{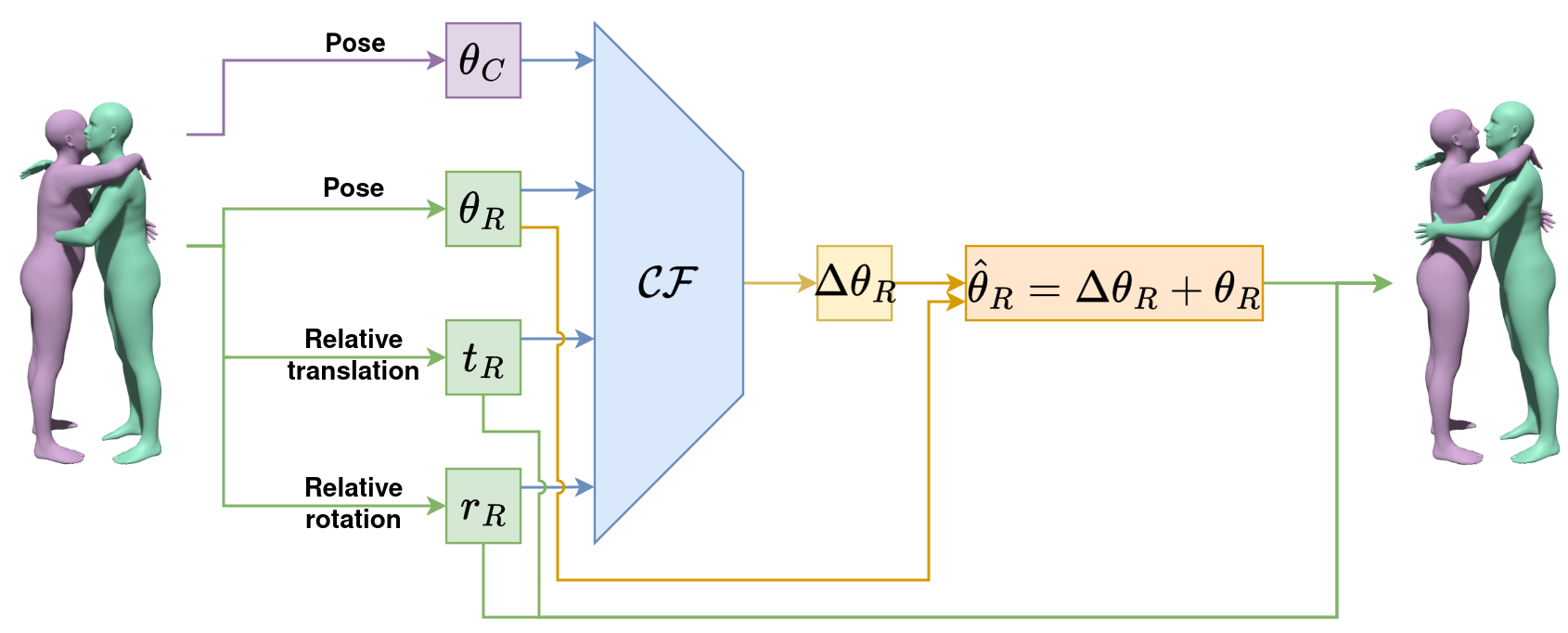}
    \caption{Detailed architecture collision resolve module. 
    The output is an offset $ \Delta \joints{R} $ that, once applied to the original reacting pose $ \joints{R} $, results in a pose $ \predjoints{R} $ that avoids colliding with the conditioning avatar with pose $ \joints{C} $.}
    \label{fig:capfix-architecture}
    \vspace{-0.2cm}
\end{figure}

Despite training on collision-free data, our cVAE cannot guarantee intersection-free samples at test-time.
To mitigate this, we propose an additional module $\capfix$, which is trained in a self-supervised manner. 
Our main idea is to infer joint rotation offsets that slightly modify the \reaction{reaction pose} to resolve collisions with the \conditioning{conditioning pose}.
We do not modify the global translation and rotation to avoid the trivial solution of moving one avatar far from the other. 
This module is illustrated in Figure~\ref{fig:capfix-architecture}.

More specifically, the $\capfix$ module is a MLP network that takes $\joints{Cv}$ and $\lbrace \predjoints{Rv}, \predrot{Rv}, \predtrans{Rv} \rbrace$ as input and predicts joint rotation offsets $\offjoints{Rv}$ such that the SMPL mesh rigged using $\lbrace \predjoints{Rv} + \offjoints{Rv}, \predrot{Rv}, \predtrans{Rv} \rbrace$ does not intersect with the SMPL mesh rigged using $\joints{Cv}$:
\begin{equation}
\begin{split}    
    \capfix: (\mathbb{R}^{23 \times 3 \times 2}, \mathbb{R}^{23 \times 3 \times 2}, \mathbb{R}^{3 \times 2}, \mathbb{R}^{3}) &\to \mathbb{R}^{23 \times 3 \times 2} \\
    (\joints{Cv}, \predjoints{Rv}, \predrot{Rv}, \predtrans{Rv}) &\mapsto \offjoints{Rv}
\end{split}
\end{equation}
To train $\capfix$ in a self-supervised manner, we first generate random SMPL poses $\joints{Cv}$ by sampling the VPoser~\cite{vposer2019cvpr} decoder. 
These poses, together with a random latent space vector $\latent \sim \mathcal{N}(0, 1)$, are fed into our decoder $\decoder$ generating the predicted poses $\decoder(\left \lbrace \joints{Cv}, \latent \right \rbrace) = \lbrace \predjoints{Rv}, \predrot{Rv}, \predtrans{Rv} \rbrace$. 

We train the $\capfix$ network end-to-end, while keeping the $\decoder$ parameters frozen, using the loss:
\begin{equation}
    \lossCF = \lambda_\text{col}\losscol + \lambda_{\offjoints{}}\lossdelta + \lambda_\text{joints}\lossjoints
\end{equation}
where $\lossdelta = ||\Delta\hat{\theta}||^2_2$ is a regularizer to penalize the joint rotation offsets from growing too large, $\lossjoints$ is a regularizer to penalize the 3D joint positions from moving too far from the original ones, and $\losscol$ is a collision penalty between two bodies, explained next.

To efficiently detect and prevent mesh intersections, we approximate each human body using a set of 24 capsules rigidly attached to the skeleton joints (see Supplementary Material for rigging details). 
Using these capsules, for a pair of avatars, $C$ and $R$, we compute a distance matrix $\mathbf{D}_{C \times R} \in \mathbb{R}^{24 \times 24}$, representing the penetration distance between a pair of capsules across both bodies.
Positive values indicate no intersection, while negative values indicate capsule overlap.
Since capsules provide a loose approximation of the avatar geometry, we establish a tolerance threshold $\mathbf{D_T}$ from our training data.
This threshold is computed as the 95th percentile of distance values across all training poses, allowing for natural proximity between avatars without triggering the collision penalty. The collision loss $\losscol$ is then defined as
\begin{equation}
    \losscol = \left\| \min(\mathbf{D}_{C \times R} - \mathbf{D_T}, 0) \right\|
\end{equation}

This formulation penalizes only the distances below the learned tolerance threshold $\mathbf{D_T}$, effectively preventing unrealistic intersections while allowing natural close interactions.

\new{We empirically set the loss weights to $\lambda_\text{col} = 30$, $\lambda_\text{joints} = 0.5$, and $\lambda_{\offjoints{}} = 0.1$. Similar to the cVAE, this module is optimized using Adam with a learning rate of $1 \times 10^{-4}$ and a batch size of 50.}

\subsection{Inference}
\label{sec:human-interaction-inference}

As hinted in Figure~\ref{fig:pipeline}, at inference time, only a \conditioning{conditioning avatar} is needed. More specifically, given a conditioning body pose $ \joints{C} $ and after sampling a latent space vector $ \latent \sim \mathcal{N}(0, 1) $, the generated \reaction{reacting avatar} can be computed as:

\begin{equation}
    \pose{R} = \capfix ( \decoder(\lbrace \joints{C}, \latent \rbrace)).
\end{equation}

\section{Results and Evaluation}
\label{sec:results-and-eval}

\begin{table*}[h]
    \centering
    \begin{tabular}{@{} c c c@{\hspace{4pt}}c@{\hspace{4pt}}c@{\hspace{4pt}}c@{\hspace{4pt}}c@{\hspace{4pt}}c@{\hspace{12pt}} c@{\hspace{4pt}}c@{\hspace{4pt}}c @{}}
        \toprule
        \multirow{2}{*}{\textbf{Method}} & \multirow{2}{*}{\makecell{\textbf{Data Aug.}}} & \multicolumn{6}{c}{\textbf{In-Distribution (ID)}} & \multicolumn{3}{c}{\textbf{Out-of-Distribution (OOD)}} \\ 
        \cmidrule(lr{12pt}){3-8} \cmidrule{9-11}
        & & $\lossrec{}\!\downarrow$ & $\mathcal{E}_{\rot{R}}$ $\!\!\downarrow$ & $\mathcal{E}_{\trans{R}}\!\!\downarrow$ & \textbf{Inter} $\!\downarrow$ & \textbf{IoU}$\downarrow$ & \textbf{VPoser}$\downarrow$ & \textbf{Inter} $\!\downarrow$ & \textbf{IoU}$\downarrow$ & \textbf{VPoser}$\downarrow$ \\ \midrule
        \makecell{cVAE $(\lossrec{}, \lossKL{})$} & No & 32.731 & \underline{151.168} & \underline{15.428} & 299.790 & 2.135 & 54.070 & 815.900 & 6.191 & 60.887 \\
        \makecell{cVAE $(\lossrec{}, \lossKL{})$} & Yes & \textbf{7.462} & \textbf{10.132} & \textbf{3.127} & 206.810 & 1.441 & \textbf{30.836} & 205.057 & 1.390 & \textbf{26.391} \\
        \makecell{cVAE $(\lossrec{}, \lossKL{}, \losscol{})$} & Yes & 66.725 & 431.798 & 41.218 & \textbf{26.090} & \textbf{0.229} & 239.468 & \textbf{15.571} & \textbf{0.138} & 233.418 \\
        \makecell{cVAE $(\lossrec{},  \lossKL{}) + \capfix{}$} & Yes & \underline{9.679} & \textbf{10.132} & \textbf{3.127} & \underline{121.690} & \underline{0.846} & \underline{31.499} & \underline{120.629} & \underline{0.833} & \underline{28.970} \\
        \bottomrule
    \end{tabular}
    \caption{In-Distribution (ID) and Out-of-Distribution (OOD) metrics for the different cVAE variants evaluated on the baseline heldout test set and standalone Mixamo keyframes respectively. Values $\lossrec{}$, $\mathcal{E}_{\rot{R}}$, $\mathcal{E}_{\trans{R}}$ and IoU have been scaled by $10^3$. The average volume intersection Inter is represented in $\text{cm}^3$. Best results are highlighted in \textbf{bold}, and second-best results are \underline{underlined}.}
    \label{table:combined-metrics}
\vspace{-0.3cm}
\end{table*}

\subsection{Ablation Study}
\label{sec:ablation}

In this section, we present several metrics to justify our architectural choices for conditional human pose generation. Specifically, we compare four distinct model variants. The first is a baseline vanilla cVAE trained exclusively with $\lossrec{}$ and $\lossKL{}$ losses (as detailed in Section~\ref{sec:cVAE}) on a subset of Hi4D~\cite{yin2023hi4d} poses identified as containing close-contact interactions. The second variant maintains this vanilla architecture but is trained on our augmented dataset. The third model builds upon this augmented version by incorporating the additional $\losscol{}$ loss during training. Finally, we evaluate our complete proposed architecture, which applies the $\capfix{}$ module to the augmented-trained vanilla cVAE.

\paragraph*{Training:}
We train our models using a sequence-heldout train-test split, keeping all frames from a given base interaction sequence strictly within either the training or testing set. \new{Specifically, the data is organized into 100 sequence groups.} The baseline split contains 5,744 poses from Hi4D interaction scans, divided into \new{76 training groups} (4,616 poses) and \new{24 testing groups} (1,128 poses). The augmented setting merges these baseline poses with 55,897 augmented poses, totaling 61,641 poses partitioned into 49,202 for training and 12,439 for testing.

\paragraph*{Metrics:} We utilize six key metrics to thoroughly evaluate our models. The first three assess pose reconstruction accuracy: the test-split reconstruction loss $\lossrec{}$, the mean error on the generated \reaction{reaction avatar} global orientation $\mathcal{E}_{\rot{R}}$, and the mean error on the generated \reaction{reaction avatar} global translation $\mathcal{E}_{\trans{R}}$. The next two metrics evaluate physical plausibility by measuring collisions. Specifically, we compute the mean intersected volume (\textbf{Inter}) $V(C \cap R)$ between each pair of \conditioning{conditioning avatar} $C$ and \reaction{reaction avatar} $R$, and the mean intersection over union (\textbf{IoU}), where the individual $\text{IoU}_{C,R}$ is defined as:

\begin{equation}
    \text{IoU}_{C,R} = \frac{V(C \cap R)}{V(C) + V(R) - V(C \cap R)}.
\end{equation}

Finally, to assess the overall naturalness of the generated poses, we compute the \textbf{VPoser} plausibility score. We take the generated body pose, isolate the first 21 body joints, encode this pose using VPoser, and calculate the latent prior energy as $\frac{1}{2} \sum (\exp(\text{logvar}) + \mu^2 - 1 - \text{logvar})$. Poses that lie closer to the learned human-pose prior yield lower energy values, indicating that lower scores reflect more plausible and realistic human poses.

\paragraph*{Evaluation:} We evaluate our models on in-distribution (ID) and out-of-distribution (OOD) benchmarks, sampling 10 generations per \conditioning{conditioning pose} for both. For the ID setup, we use 100 \conditioning{conditioning poses} from the baseline test split (yielding 1,000 generations). Because these samples possess ground-truth interacting partners, we evaluate them using the complete suite of reconstruction, collision, and plausibility metrics defined above. For the OOD setting, we select 70 keyframes from 35 Mixamo animations (yielding 700 generations). Since we lack ground-truth \reaction{reaction poses} for this out-of-distribution set, our evaluation relies strictly on measuring the physical plausibility (VPoser) and collision (IoU, Inter) metrics of the generated \reaction{reaction poses}.

\paragraph*{Results}
As shown in Table~\ref{table:combined-metrics}, the performance trends remain remarkably consistent across both the In-Distribution (ID) and Out-of-Distribution (OOD) evaluations. First, we observe a clear improvement when training the baseline cVAE on our augmented dataset (second row) compared to the non-augmented baseline (first row), resulting in a drastic reduction across all the metrics. The augmented baseline establishes a strong foundation, yielding the best overall reconstruction loss ($\lossrec{}$) as well as the lowest global orientation ($\mathcal{E}_{\rot{R}}$) and translation ($\mathcal{E}_{\trans{R}}$) errors.

To address the remaining mesh intersections, incorporating the additional capsule loss term $\losscol{}$ (third row) proves highly effective at avoiding collisions, yielding the absolute lowest Inter and IoU values. However, this aggressive collision resolution comes at a severe cost. Because the loss penalty artificially pushes bodies apart to clear intersections, the reconstruction loss ($\lossrec{}$) degrades significantly, and the human pose plausibility prior is broken, as evidenced by a drastic spike in the VPoser energy.

Finally, applying our proposed $\capfix{}$ module (fourth row) offers the most balanced trade-off, consistently securing first or second best across all evaluated metrics. Because $\capfix{}$ resolves collisions by exclusively modifying the local body pose, the global trajectory errors ($\mathcal{E}_{\rot{R}}$ and $\mathcal{E}_{\trans{R}}$) remain entirely unaffected, maintaining the best-in-class performance of the augmented baseline. While moving vertices to prevent collisions inevitably incurs a slight penalty to the reconstruction loss compared to the unconstrained augmented model, $\capfix{}$ preserves a significantly better $\lossrec{}$ than the capsule loss. Ultimately, our module successfully minimizes collisions while maintaining natural and plausible body structures, yielding a VPoser energy that is nearly identical to the augmented-only approach.

\subsection{Quantitative evaluation}
To further quantitatively evaluate our method, we compare our results with  BUDDI~\cite{mueller2024buddi} in terms of mesh intersections.
To enable this comparison, we adapt BUDDI to generate \reaction{reaction poses} conditioned on a \conditioning{conditioning pose} using an in-painting strategy:
at each step of the BUDDI sampling process, we pass the \conditioning{conditioning pose} prior to the diffusion and denoising steps.

Table~\ref{table:quantitative-evaluation} presents two metrics (\textbf{Inter} and \textbf{IoU}) as described in the previous section.
To ensure a fair comparison and avoid dataset bias, we compute these metrics not on randomly sampled outputs from our model, but on a specific split of the Hi4D dataset.
While both our model and BUDDI were trained on instances from the full Hi4D dataset, we ensure that the selected split contains only poses that were unseen by our model during training.
For additional fairness and to obtain a more informative evaluation, we only evaluate test samples that generate \reaction{reaction poses} that are in-contact with the \conditioning{conditioning pose}. %
This is to avoid naive samples that are in interaction but are not in contact, which, of course, do not suffer from mesh intersections.

Our results in Table~\ref{table:quantitative-evaluation} show that BUDDI samples exhibit significantly more mesh intersections, which is reflected in the higher \textbf{Inter} and \textbf{IoU} values. 
In contrast, our method maintains a marginal intersection error, showcasing that it is able to generate interacting meshes that are in close contact but do not intersect with each other.

\begin{table}[htbp]
\vspace{-0.1cm}
    \centering
    \renewcommand{\arraystretch}{1.5}
    \begin{tabular}{@{}c@{\hspace{4pt}}c@{\hspace{4pt}}c@{\hspace{4pt}}}
        \toprule
        \textbf{Method} &\textbf{Inter}$\downarrow$ & \textbf{IoU} $\!\downarrow$ \\ \midrule
        \makecell{Ours} & \textbf{182.45} & \textbf{1.25}  \\ 
        \makecell{BUDDI \cite{mueller2024buddi}} & 2283.90 & 16.87 \\
        \bottomrule
    \end{tabular}
    \caption{Quantitative comparison between our method and BUDDI~\cite{mueller2024buddi}. IoU values have been scaled by $10^3$. The average volume intersection Inter is represented in $\text{cm}^3$}
    \label{table:quantitative-evaluation}
\vspace{-0.5cm}
\end{table}

We further illustrate these differences in the qualitative evaluation Section \ref{sec:human-interaction-qualitative-evaluation}.

\vspace{-0.2cm}
\subsection{Qualitative evaluation}
\label{sec:human-interaction-qualitative-evaluation}
In this section, we present qualitative results of our approach and compare with BUDDI~\cite{mueller2024buddi}. 
To this end, we generated conditioned \reaction{reaction avatars} with BUDDI as explained in the previous section.

\begin{figure*}[t]
    \centering
    \resizebox{\linewidth}{!}
    {%
\begin{tikzpicture}

            \node (img0) at (-1,8.8) 
            {\includegraphics[width=0.15\linewidth]{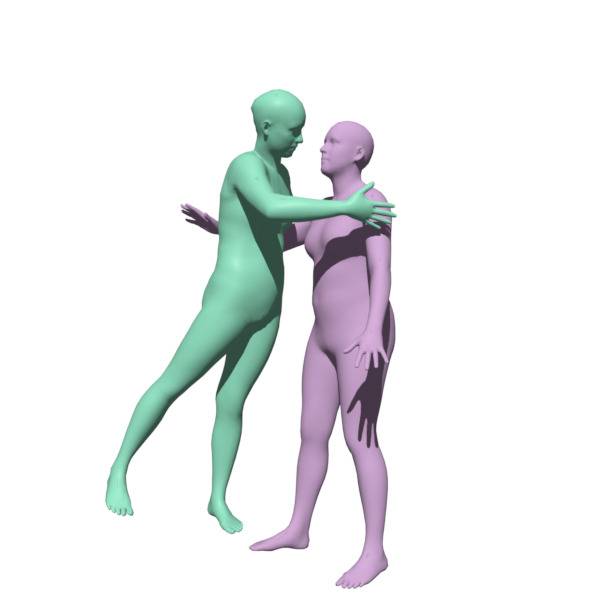}};
            \node (img1) at (1.2,8.8) 
            {\includegraphics[width=0.15\linewidth]{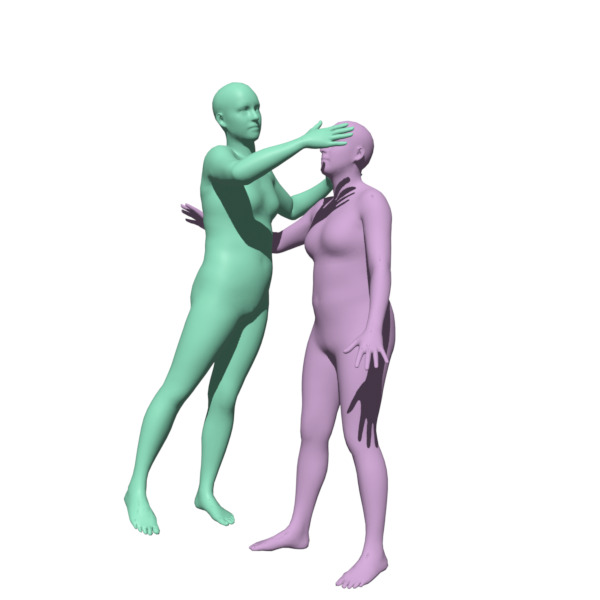}};
            \node (img2) at (3.4,8.8) 
            {\includegraphics[width=0.15\linewidth]{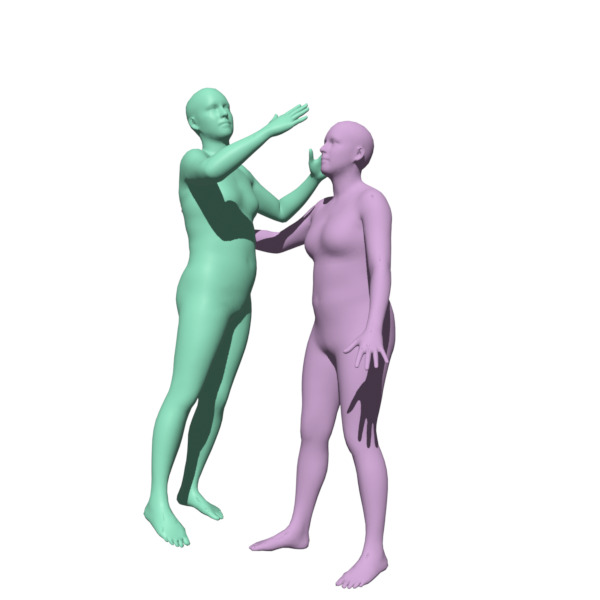}};
            \node (img3) at (5.6,8.8) 
            {\includegraphics[width=0.15\linewidth]{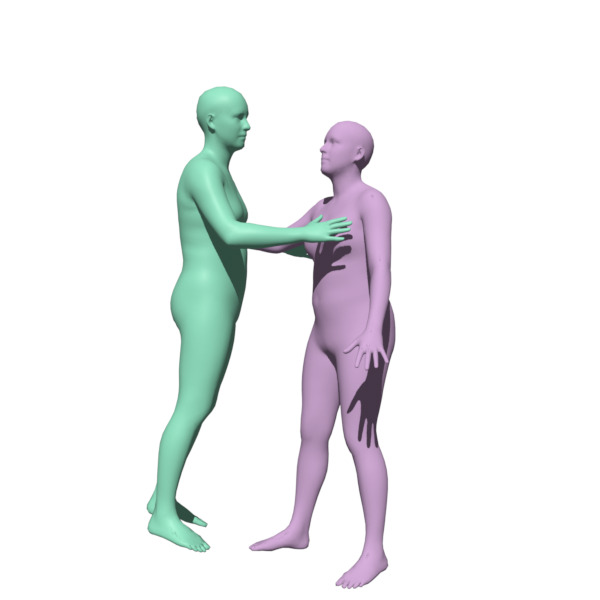}};
            \node (img5) at (7.8,8.8) 
            {\includegraphics[width=0.15\linewidth]{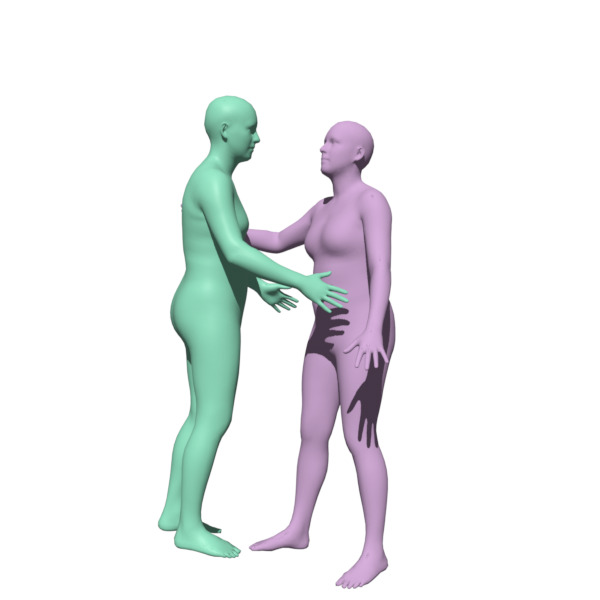}};
            \node (img6) at (10,8.8) 
            {\includegraphics[width=0.15\linewidth]{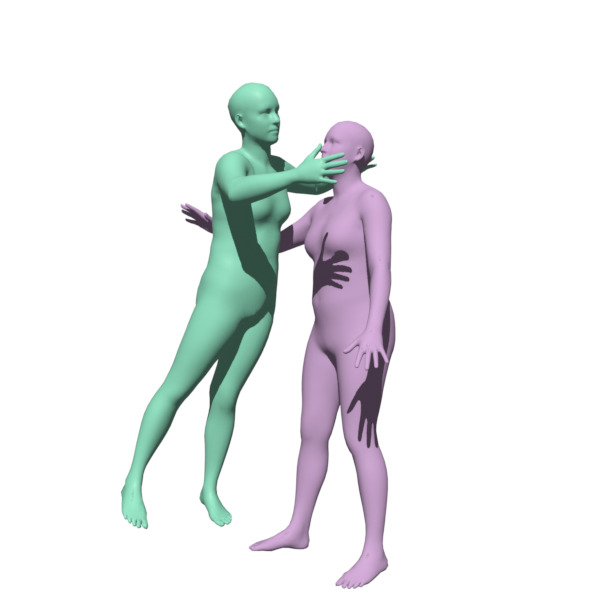}};
            \node (img7) at (12.2,8.8) 
            {\includegraphics[width=0.15\linewidth]{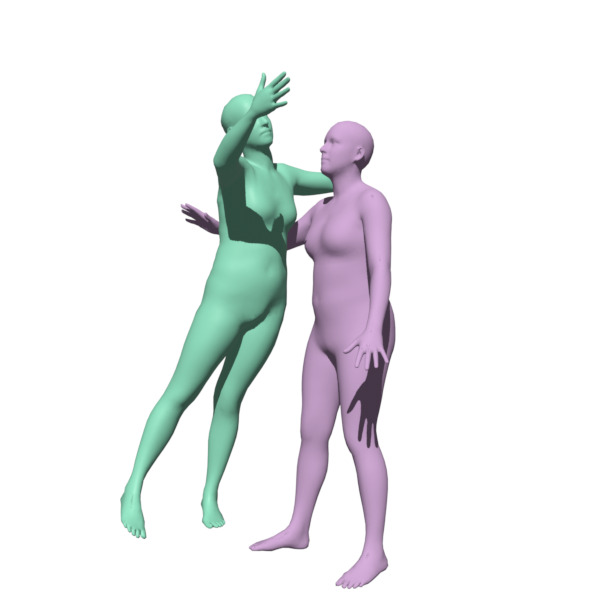}};
            \node (img8) at (14.4,8.8) 
            {\includegraphics[width=0.15\linewidth]{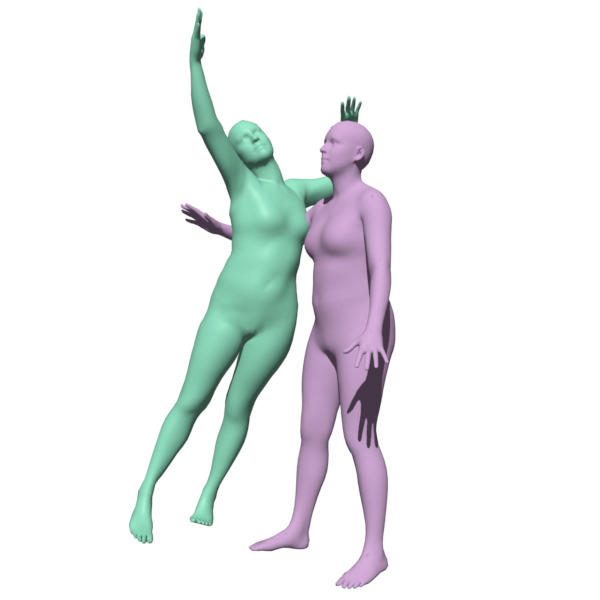}};
            \node (input)[align=center, rotate=90, xshift = -45pt, yshift = 35pt] at (img0.north){Ours};
            \node (input)[align=center, rotate=90, xshift = -95pt, yshift = 50pt] at (img0.north){Interpolation 1};
            \node (img10) at (-1,6.2) 
            {\includegraphics[width=0.15\linewidth]{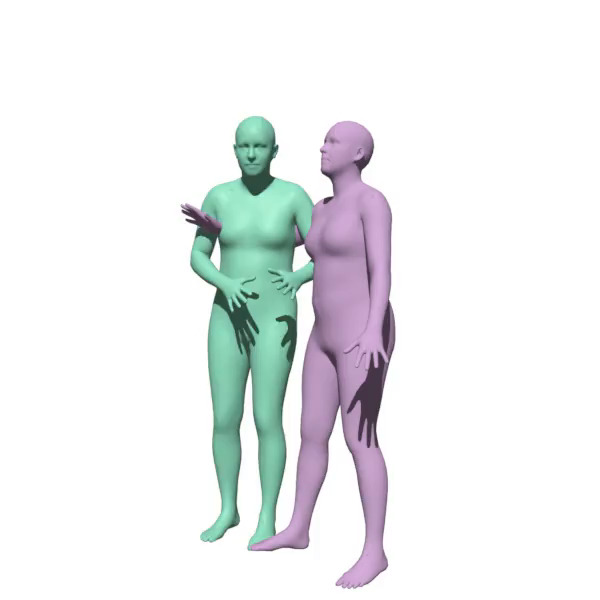}};
            \node (img11) at (1.2,6.2) 
            {\includegraphics[width=0.15\linewidth]{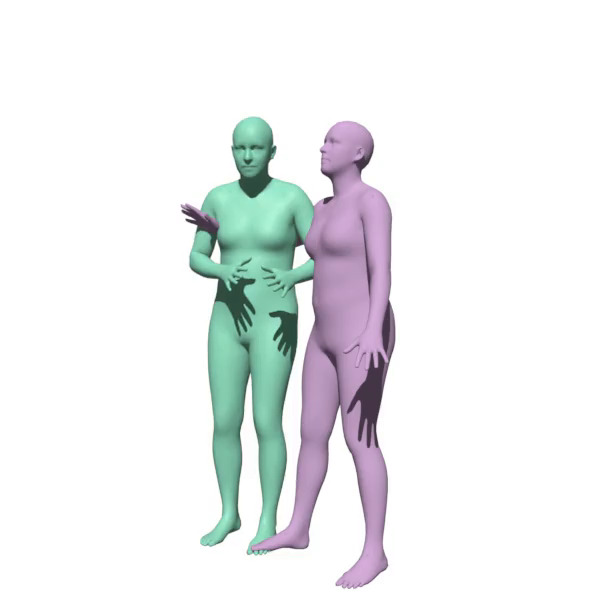}};
            \node (img12) at (3.4,6.2) 
            {\includegraphics[width=0.15\linewidth]{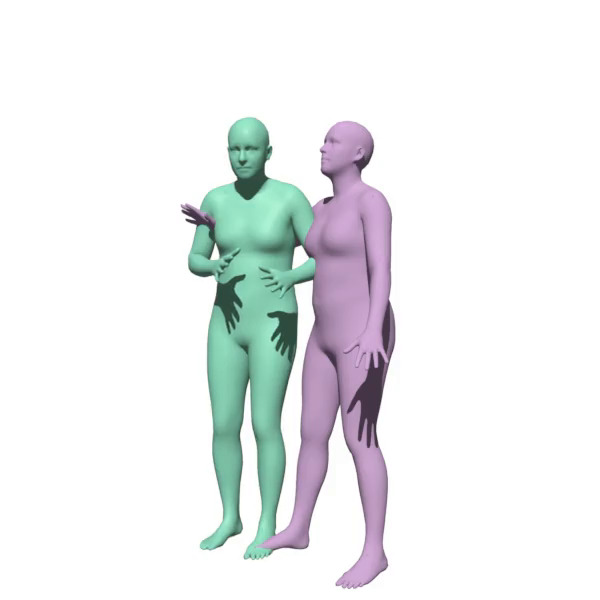}};
            \node (img13) at (5.6,6.2) 
            {\includegraphics[width=0.15\linewidth]{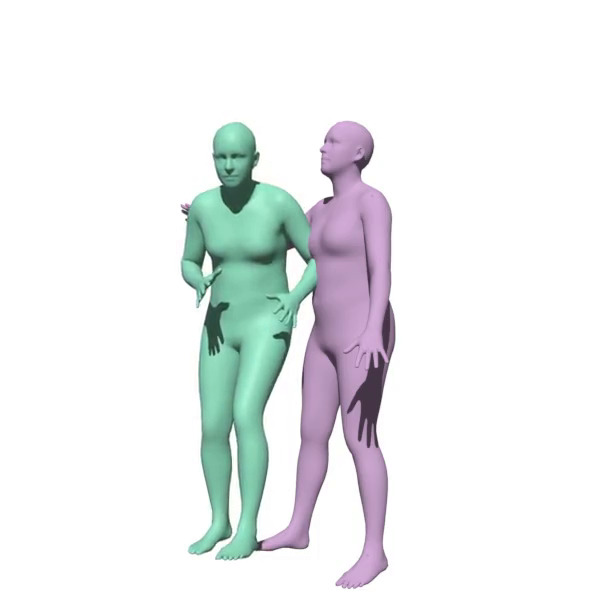}};
            \node (img14) at (7.8,6.2) 
            {\includegraphics[width=0.15\linewidth]{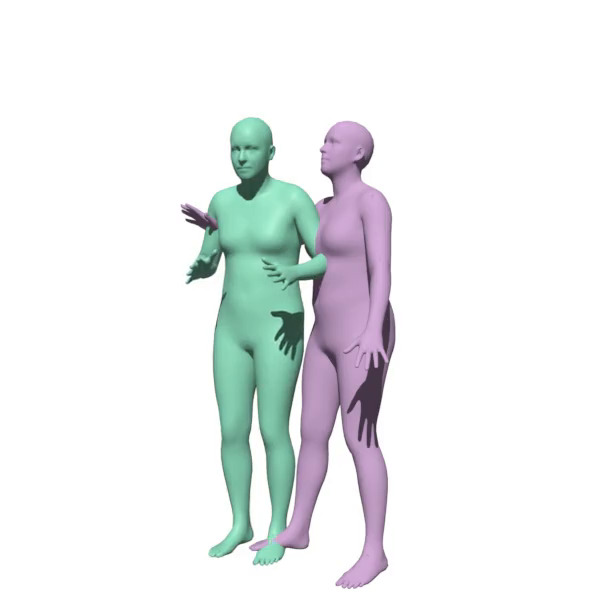}};
            \node (img15) at (10,6.2) 
            {\includegraphics[width=0.15\linewidth]{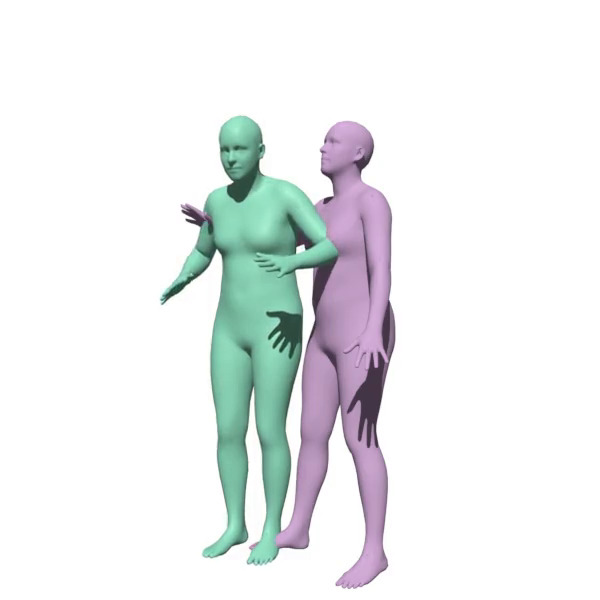}};
            \node (img16) at (12.2,6.2) 
            {\includegraphics[width=0.15\linewidth]{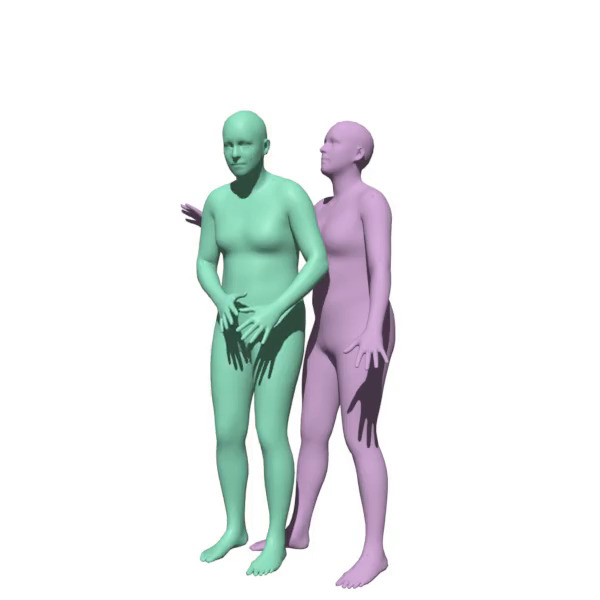}};
            \node (img17) at (14.4,6.2) 
            {\includegraphics[width=0.15\linewidth]{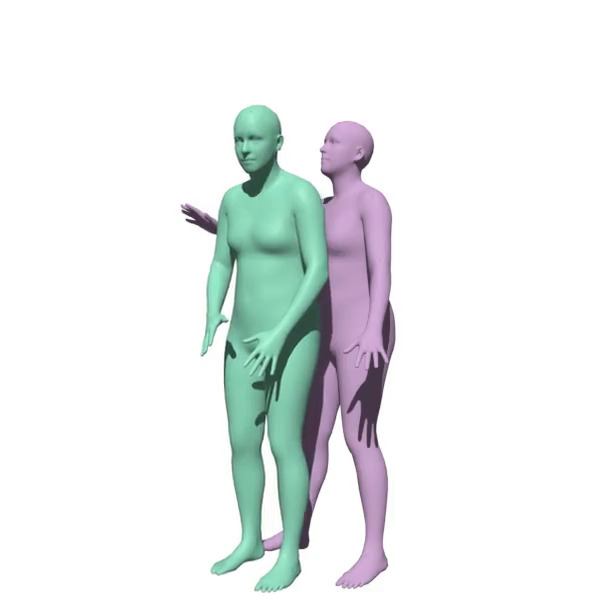}};
            \node (input)[align=center, rotate=90, xshift = -45pt, yshift = 35pt] at (img10.north){BUDDI};
            \node (img20) at (-1,2.6) 
            {\includegraphics[width=0.15\linewidth]{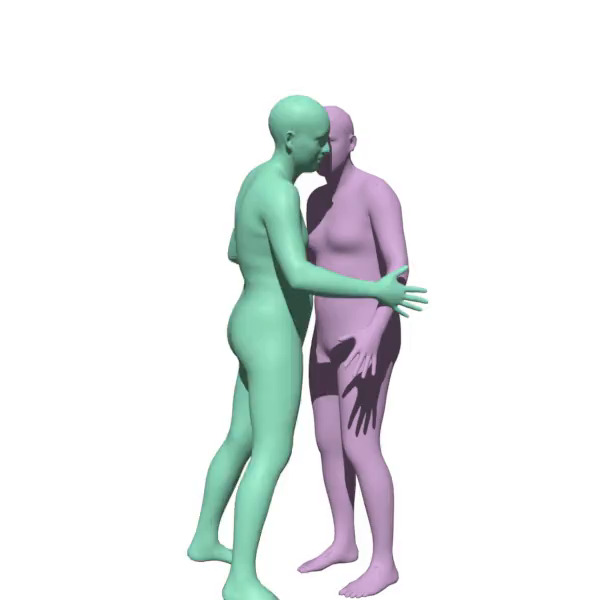}};
            \node (img21) at (1.2,2.6) 
            {\includegraphics[width=0.15\linewidth]{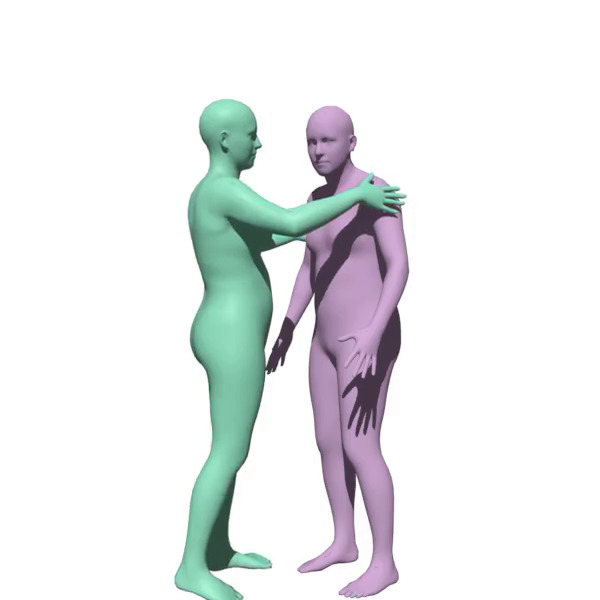}};
            \node (img22) at (3.4,2.6) 
            {\includegraphics[width=0.15\linewidth]{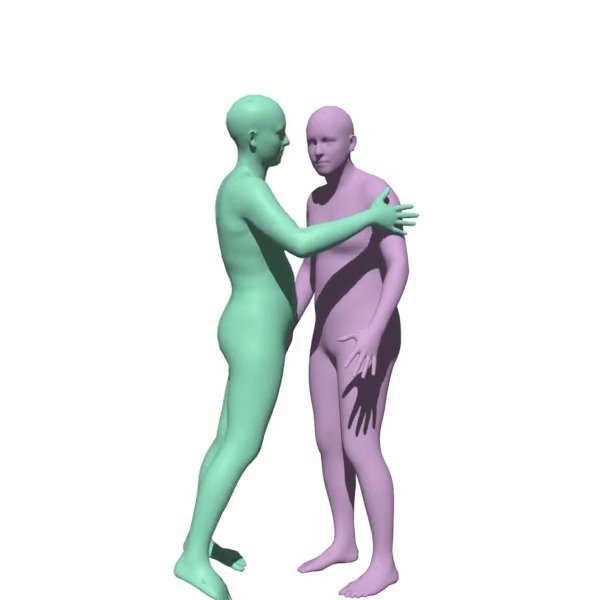}};
            \node (img23) at (5.6,2.6) 
            {\includegraphics[width=0.15\linewidth]{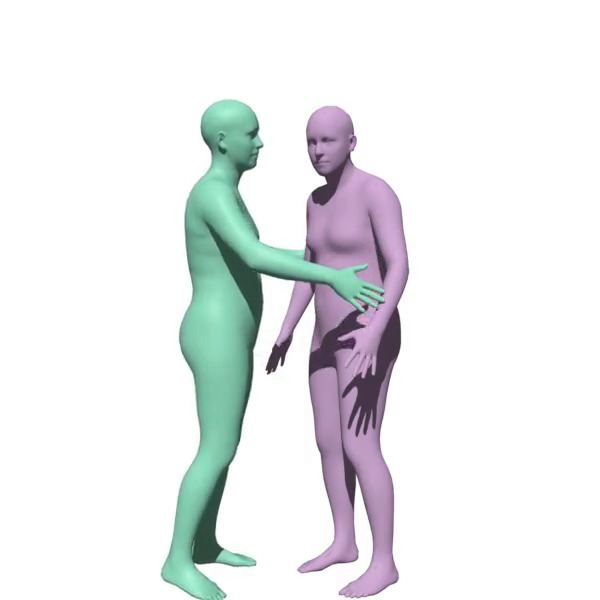}};
            \node (img24) at (7.8,2.6) 
            {\includegraphics[width=0.15\linewidth]{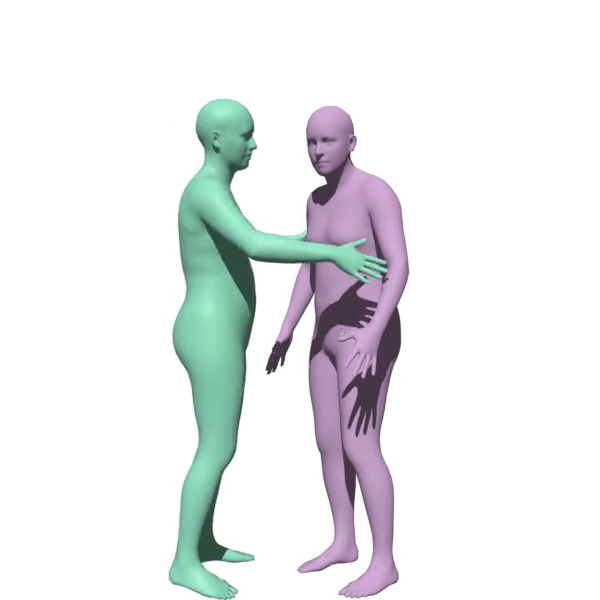}};
            \node (img25) at (10,2.6) 
            {\includegraphics[width=0.15\linewidth]{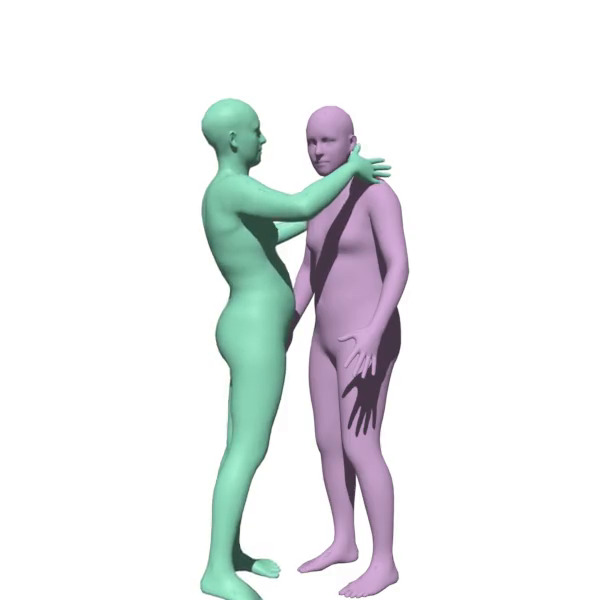}};
            \node (img26) at (12.2,2.6) 
            {\includegraphics[width=0.15\linewidth]{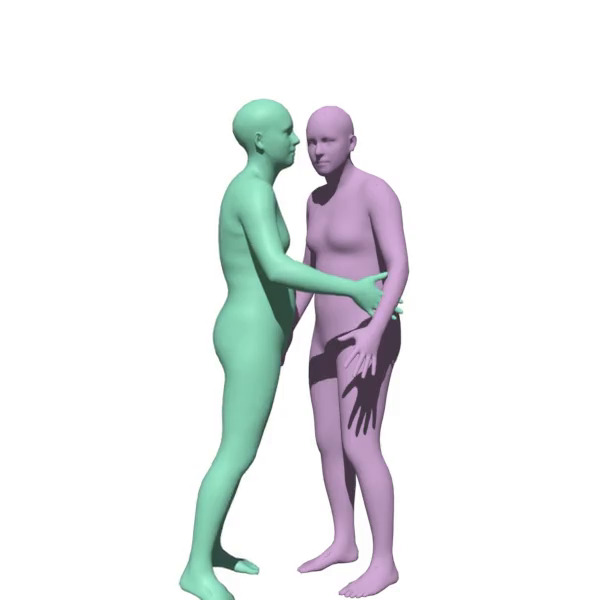}};
            \node (img27) at (14.4,2.6) 
            {\includegraphics[width=0.15\linewidth]{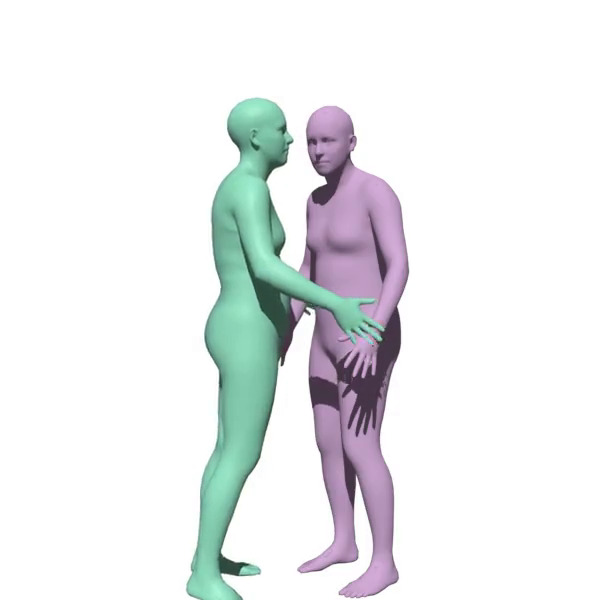}};
            \node (input)[align=center, rotate=90, xshift = -45pt, yshift = 35pt] at (img20.north){Ours};
            \node (input)[align=center, rotate=90, xshift = -95pt, yshift = 50pt] at (img20.north){Interpolation 2};
                        
            \node (img47) at (14.4,0) 
            {\includegraphics[width=0.15\linewidth]{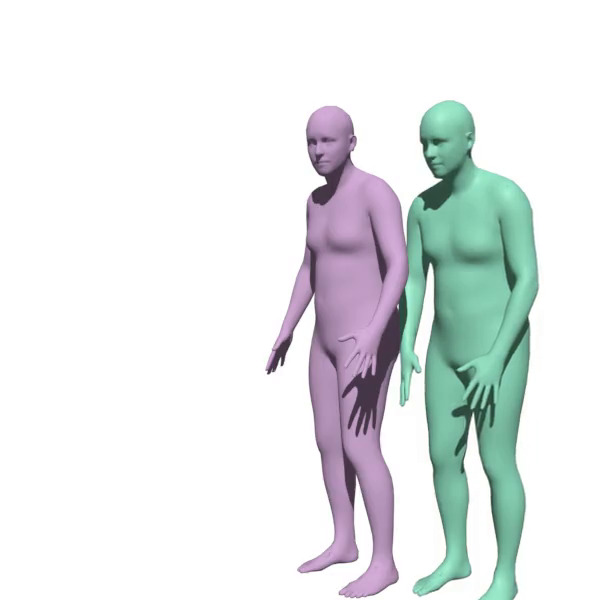}};
            \node (img46) at (12.2,0) 
            {\includegraphics[width=0.15\linewidth]{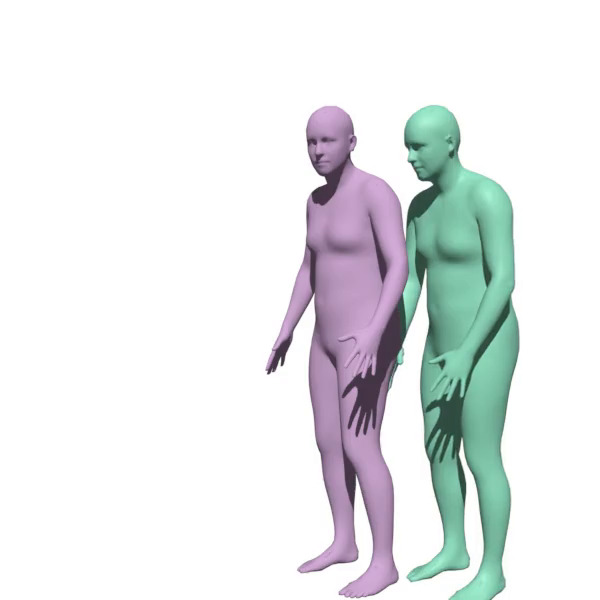}};
            \node (img45) at (10,0) 
            {\includegraphics[width=0.15\linewidth]{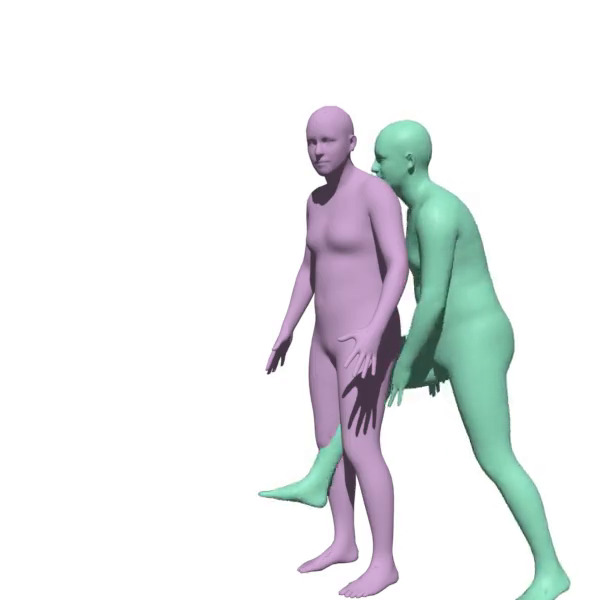}};
            \node (img44) at (7.8,0) 
            {\includegraphics[width=0.15\linewidth]{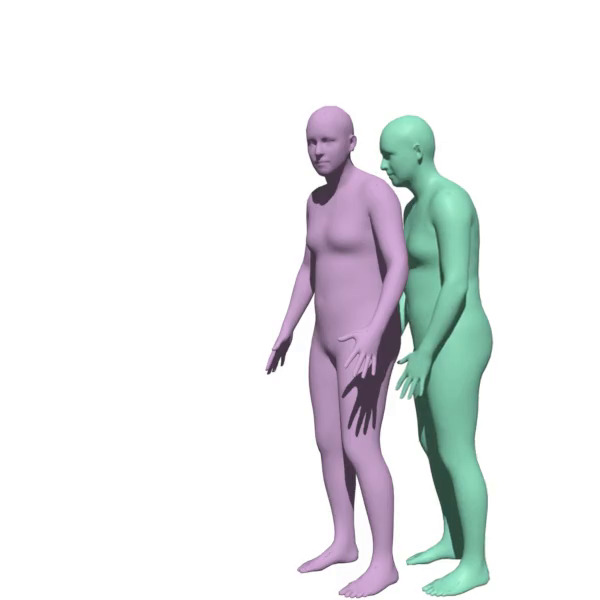}};
            \node (img43) at (5.6,0) 
            {\includegraphics[width=0.15\linewidth]{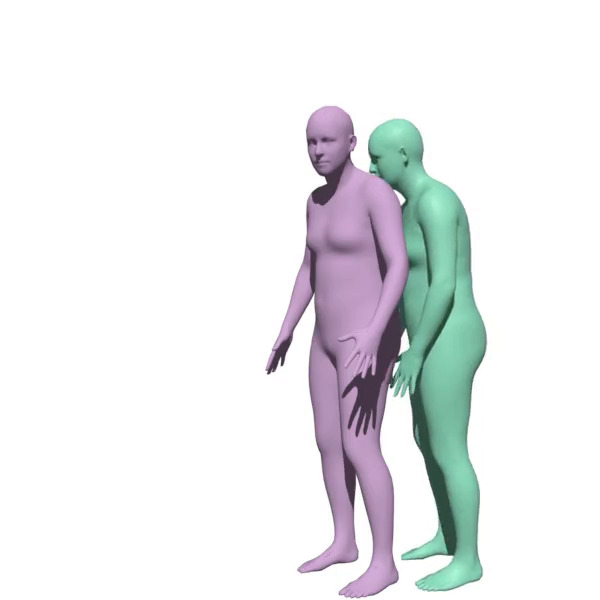}};
            \node (img42) at (3.4,0) 
            {\includegraphics[width=0.15\linewidth]{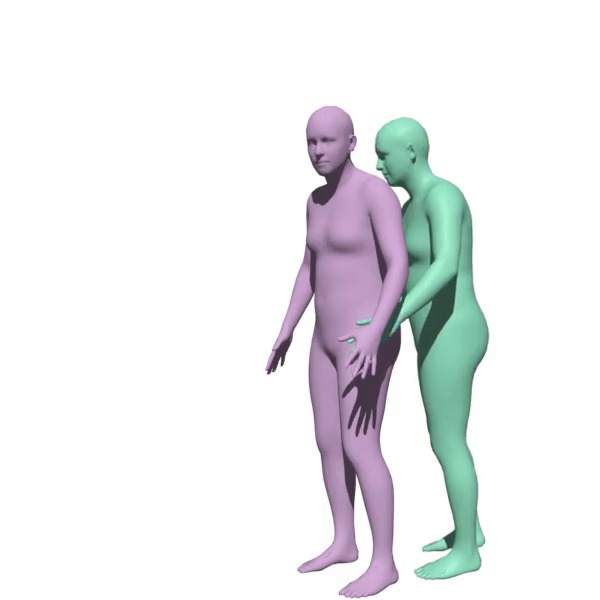}};
            \node (img41) at (1.2,0) 
            {\includegraphics[width=0.15\linewidth]{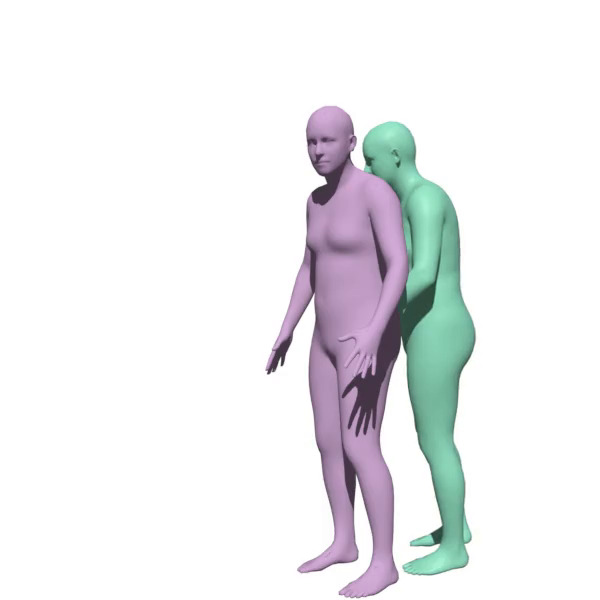}};
            \node (img40) at (-1,0) 
            {\includegraphics[width=0.15\linewidth]{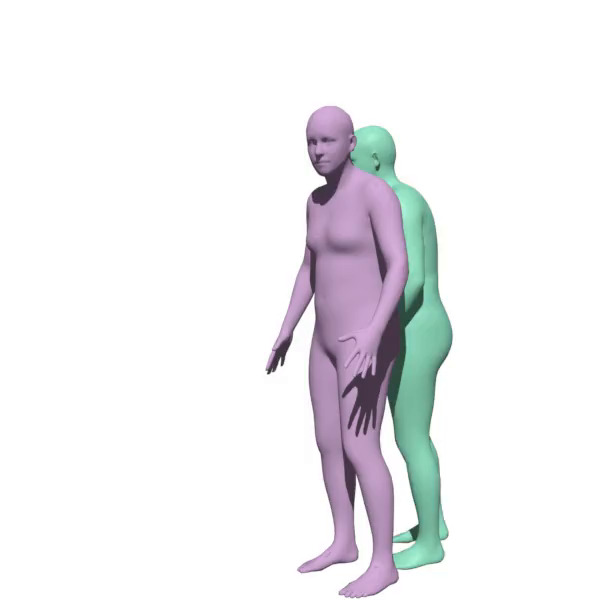}};
            \node (input)[align=center, rotate=90, xshift = -45pt, yshift = 35pt] at (img40.north){BUDDI};
\end{tikzpicture}
    }
    \caption{Two examples (Interpolation 1 and Interpolation 2) of the \reaction{reacting poses} yielded by interpolating two latent vectors (left and right columns) for a constant \conditioning{conditioning pose}. The \conditioning{conditioning poses} were originally generated with BUDDI, hence were never seen at train time by our model.
    Our model generates smoother and more expressive poses, notice seamless and smooth changes of the \reaction{reacting poses} along the horizontal axis. In contrast, BUDDI \reaction{reacting poses} are less expressive and exhibit pose discontinuities.}
    \label{fig:latent-space-interpolation}
\vspace{-0.5cm}
\end{figure*}

We first evaluate the smoothness of the latent space of each model by interpolating between two random latent vectors $z$, using a fixed \conditioning{conditioning pose}.
For a fair comparison and to avoid any dataset bias, we specifically select \conditioning{conditioning poses} that are originally generated with BUDDI, hence never seen at train time by our model.
In Figure~\ref{fig:latent-space-interpolation}, we present for both Ours and BUDDI, 8 equally spaced steps of interpolation between two random samples. 
Our results demonstrate that the interpolations generated by our method exhibit smoother transitions, a greater diversity of poses, and fewer mesh collisions compared to those produced by BUDDI. For further insights, please refer to the Supplementary Video and Material.

\begin{figure}[h]
\vspace{-0.2cm}
    \centering
    \includegraphics[width=\linewidth]{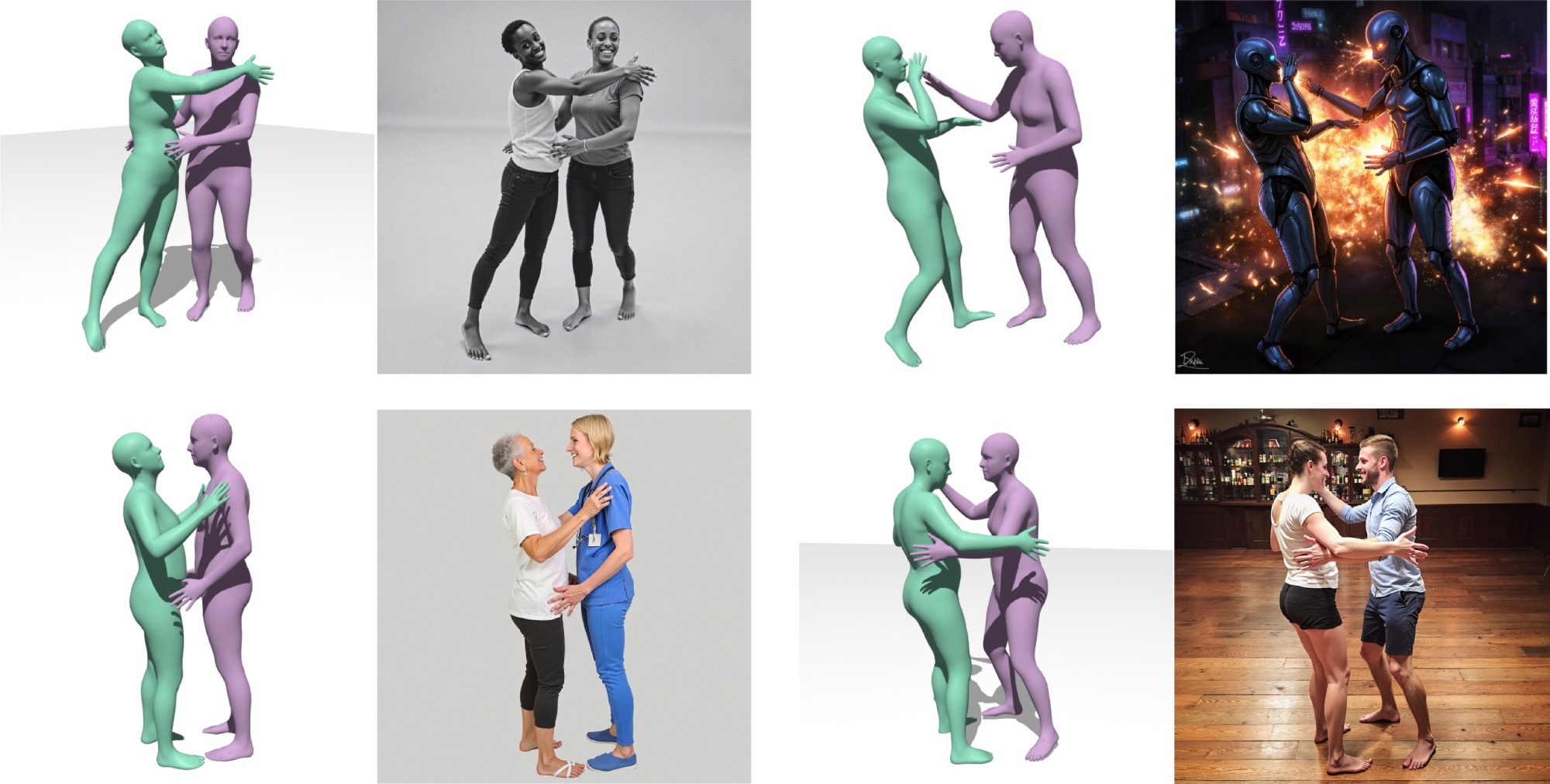}
    \caption{Our results (i.e., renders with \conditioning{conditioning} and \reaction{reaction} poses) can be used as accurate driving signal to control image-to-image models \cite{flux2024}. This enables fine-control synthesis of photorealistic images with complex human-human interactions.}
    \label{fig:controlnet}
\vspace{-0.4cm}
\end{figure}

\begin{figure*}[t]
    \centering
    \resizebox{\linewidth}{!}
    {%
\begin{tikzpicture}
            \node (img0) at (-1,16) 
            {\includegraphics[width=0.15\linewidth]{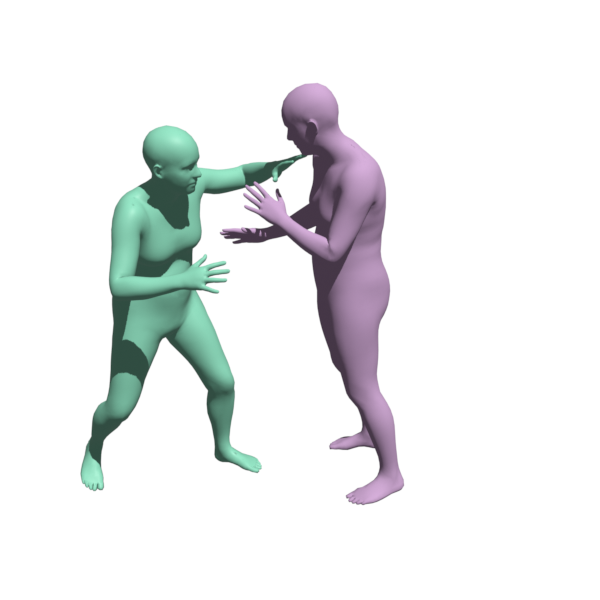}};
            \node (img1) at (1,16) 
            {\includegraphics[width=0.15\linewidth]{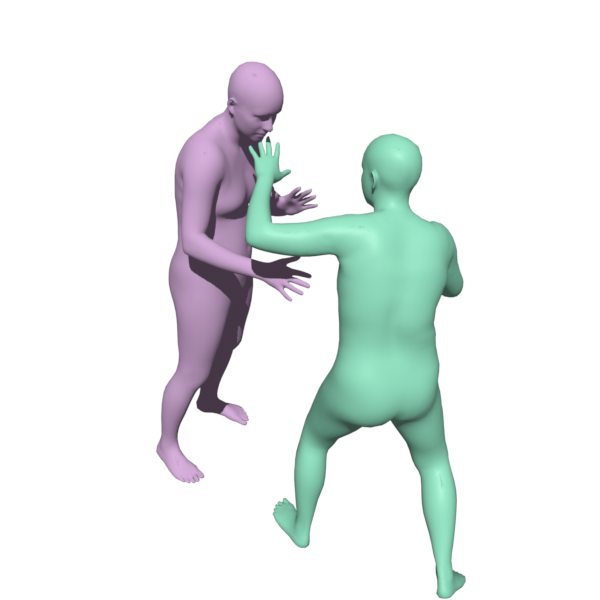}};
            \node (img2) at (3.7,16) 
            {\includegraphics[width=0.15\linewidth]{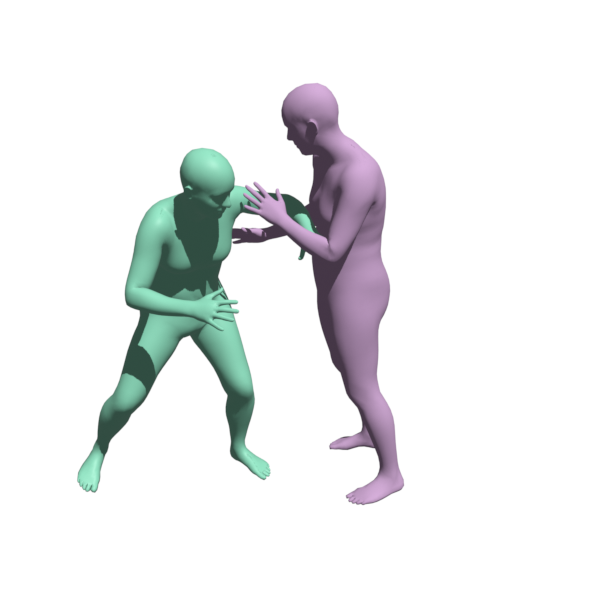}};
            \node (img3) at (5.7,16) 
            {\includegraphics[width=0.15\linewidth]{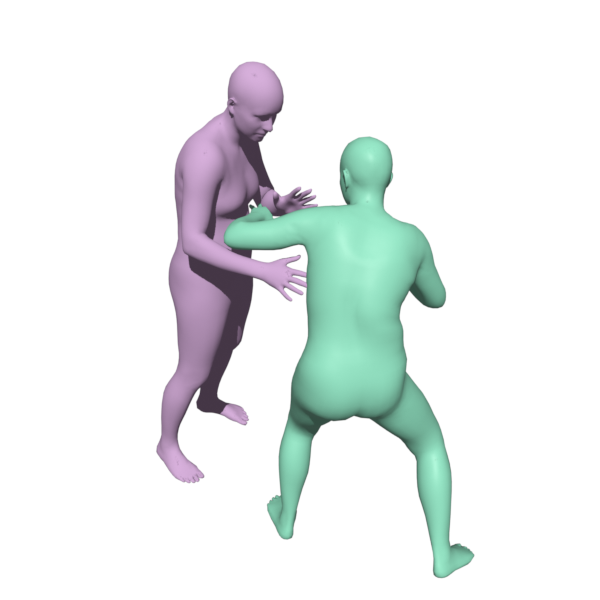}};
            \node (img5) at (8.4,16) 
            {\includegraphics[width=0.15\linewidth]{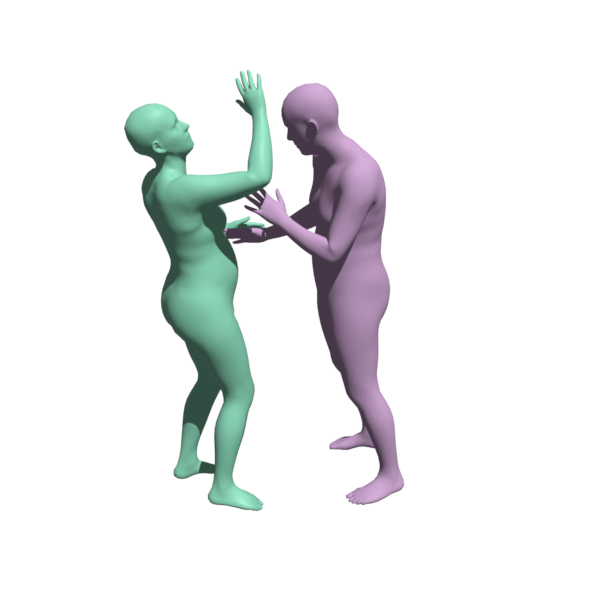}};
            \node (img6) at (10.4,16) 
            {\includegraphics[width=0.15\linewidth]{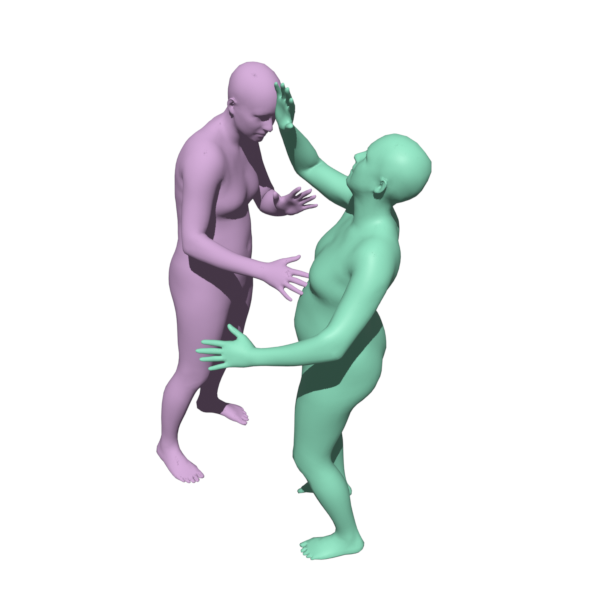}};
            \node (img7) at (13.1,16) 
            {\includegraphics[width=0.15\linewidth]{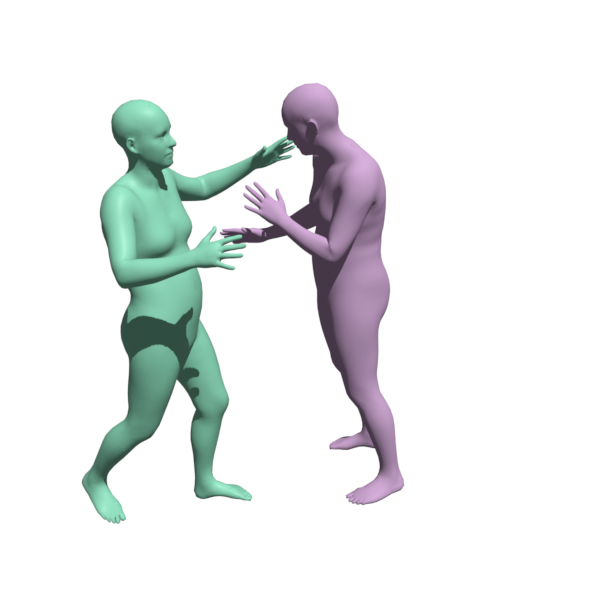}};
            \node (img8) at (15.1,16) 
            {\includegraphics[width=0.15\linewidth]{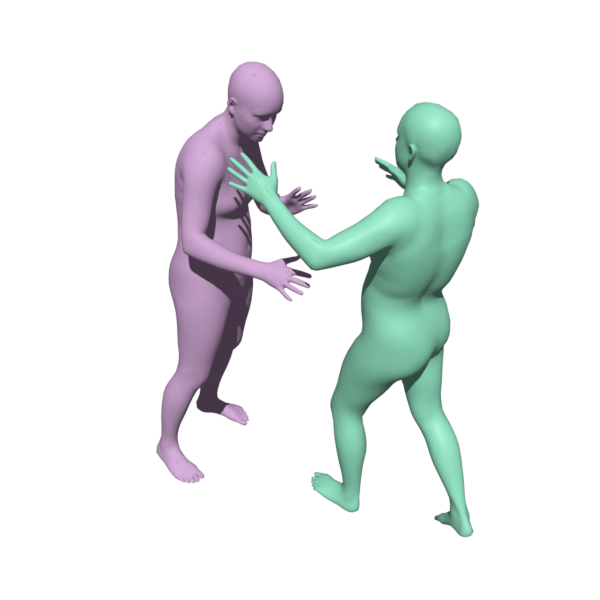}};
            \node (input)[align=center, rotate=90, xshift = -45pt, yshift = 35pt] at (img0.north){Ours};
            \node (input)[align=center, rotate=90, xshift = -95pt, yshift = 50pt] at (img0.north){Sampling (Basketball)};
            \node (img10) at (-1,13) 
            {\includegraphics[width=0.15\linewidth]{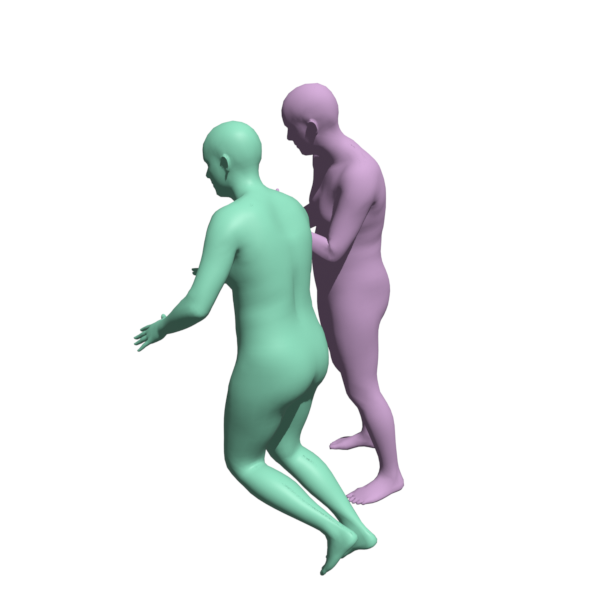}};
            \node (img11) at (1,13) 
            {\includegraphics[width=0.15\linewidth]{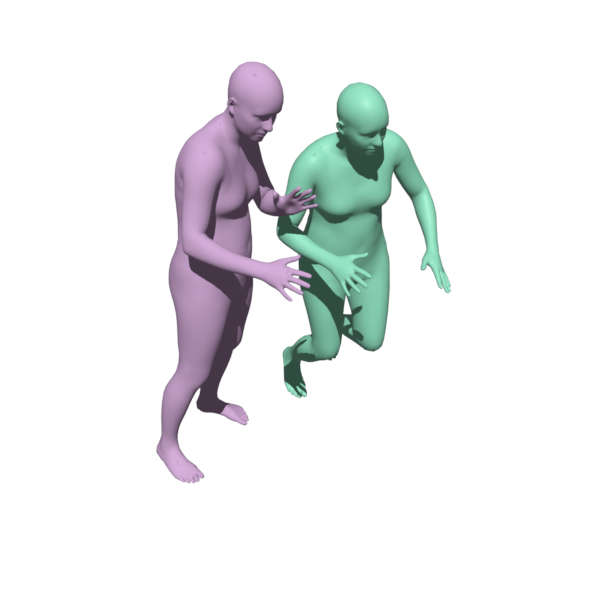}};
            \node (img12) at (3.7,13) 
            {\includegraphics[width=0.15\linewidth]{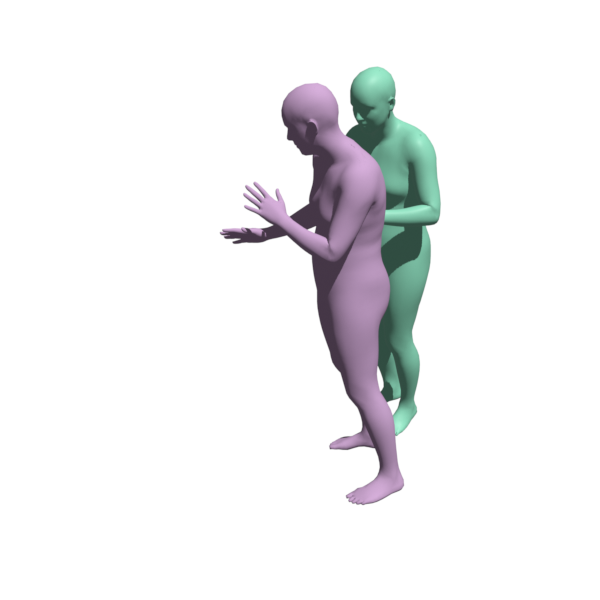}};
            \node (img13) at (5.7,13) 
            {\includegraphics[width=0.15\linewidth]{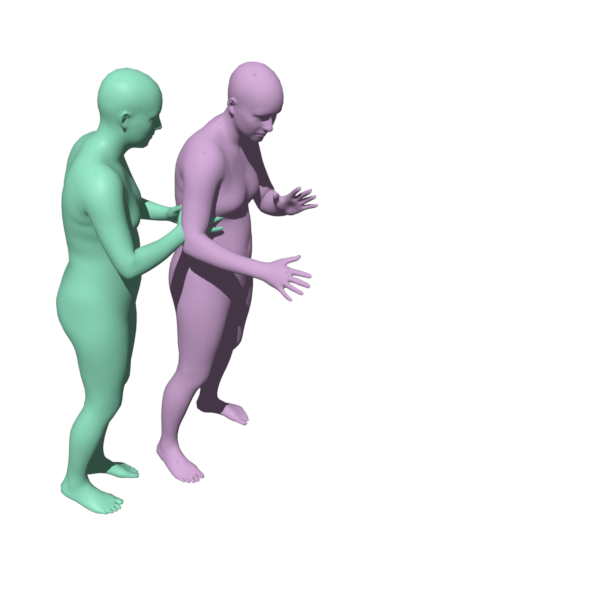}};
            \node (img14) at (8.4,13) 
            {\includegraphics[width=0.15\linewidth]{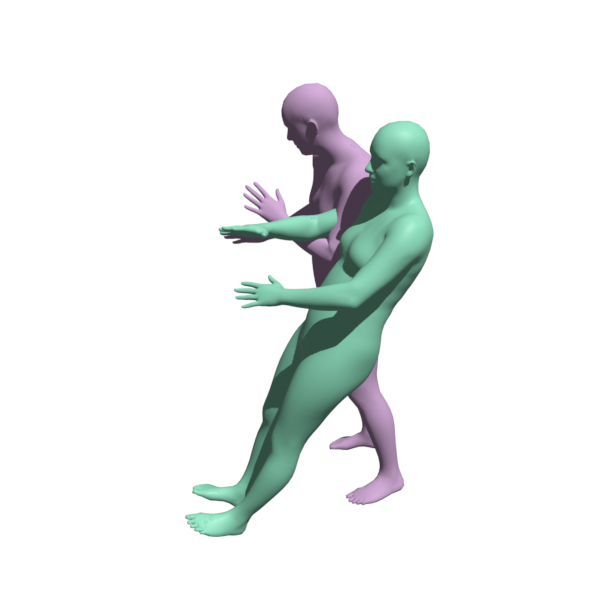}};
            \node (img15) at (10.4,13) 
            {\includegraphics[width=0.15\linewidth]{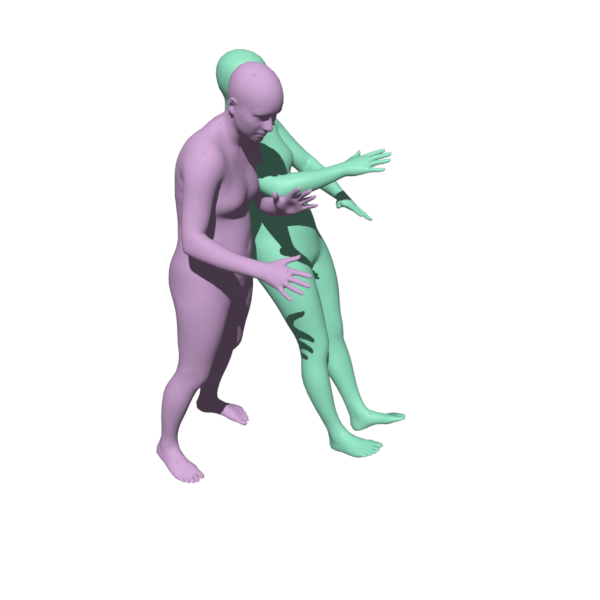}};
            \node (img16) at (13.1,13) 
            {\includegraphics[width=0.15\linewidth]{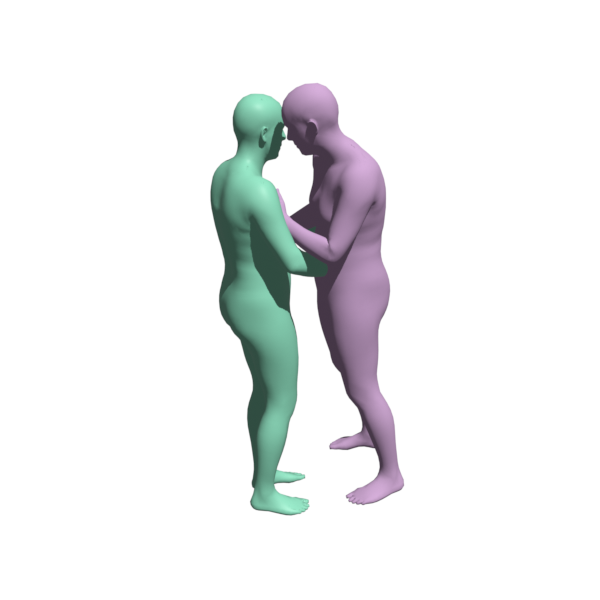}};
            \node (img17) at (15.1,13) 
            {\includegraphics[width=0.15\linewidth]{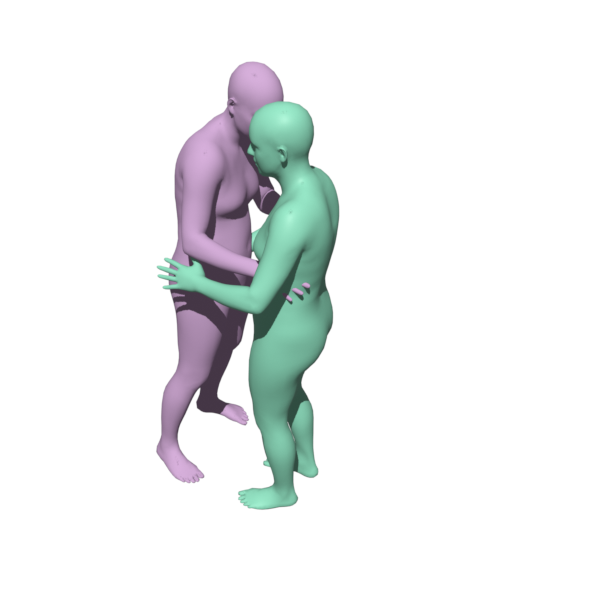}};
            \node (input)[align=center, rotate=90, xshift = -45pt, yshift = 35pt] at (img10.north){BUDDI};
            \node (img0) at (-1,9.5) 
            {\includegraphics[width=0.15\linewidth]{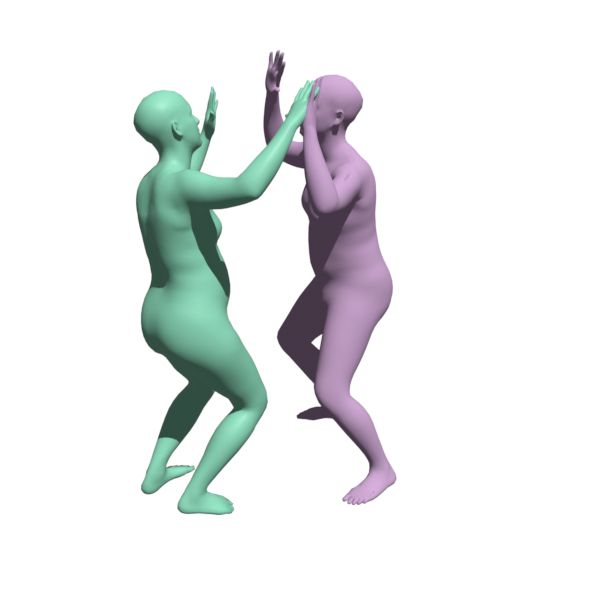}};
            \node (img1) at (1,9.5) 
            {\includegraphics[width=0.15\linewidth]{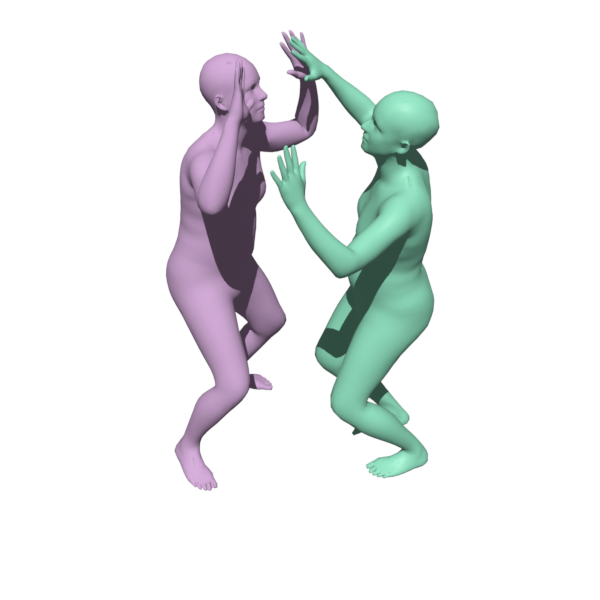}};
            \node (img2) at (3.7,9.5) 
            {\includegraphics[width=0.15\linewidth]{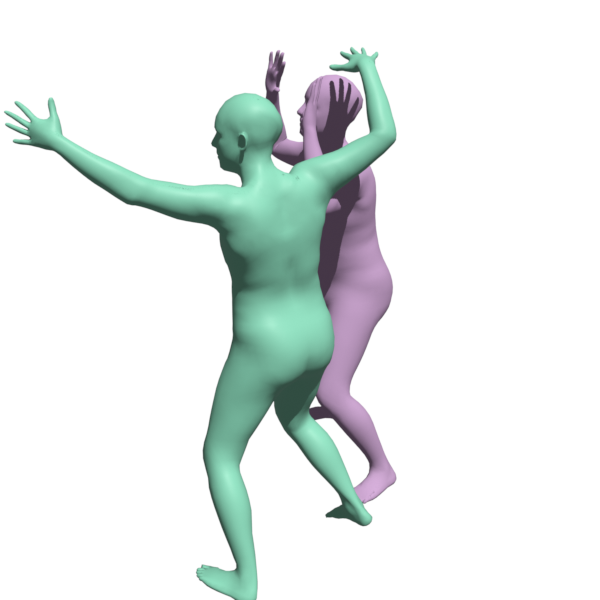}};
            \node (img3) at (5.7,9.5) 
            {\includegraphics[width=0.15\linewidth]{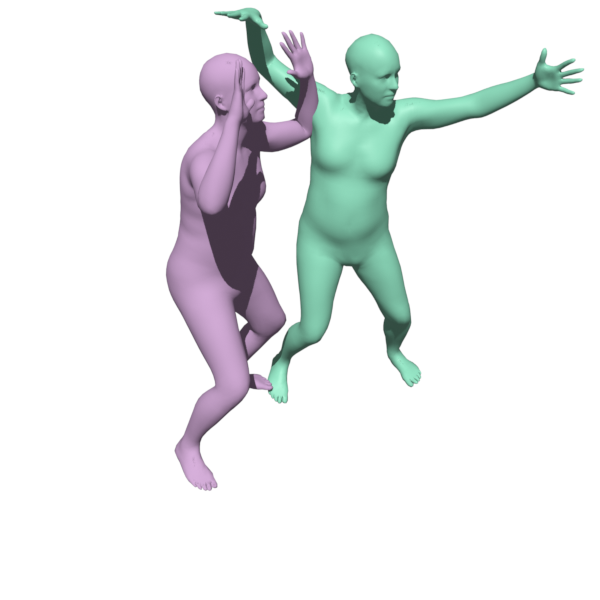}};
            \node (img5) at (8.4,9.5) 
            {\includegraphics[width=0.15\linewidth]{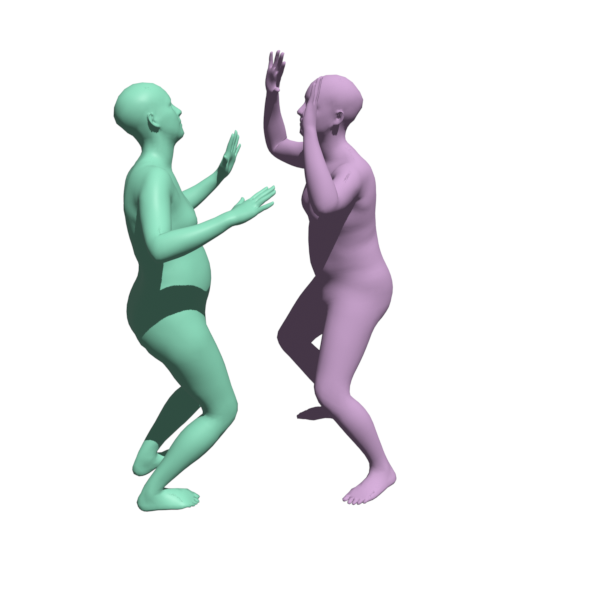}};
            \node (img6) at (10.4,9.5) 
            {\includegraphics[width=0.15\linewidth]{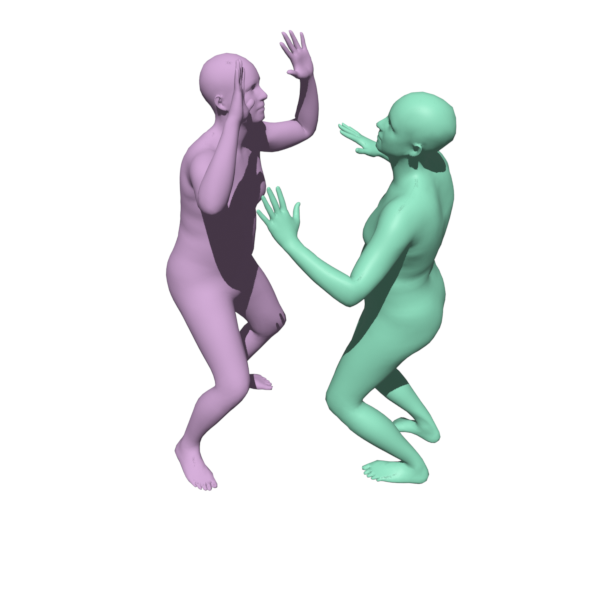}};
            \node (img7) at (13.1,9.5) 
            {\includegraphics[width=0.15\linewidth]{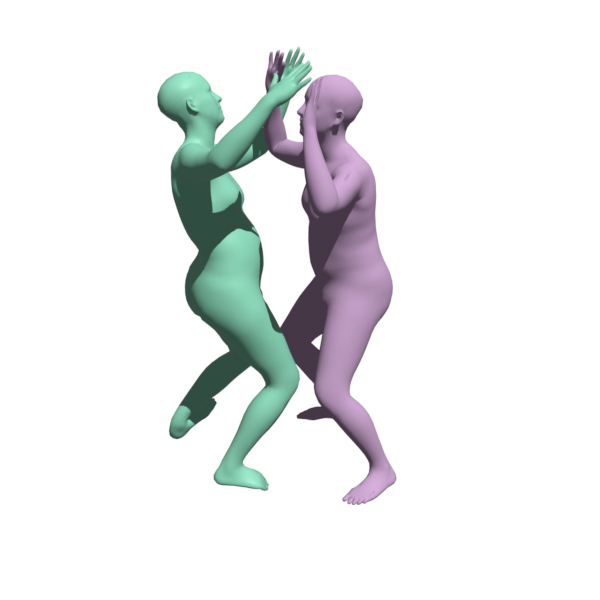}};
            \node (img8) at (15.1,9.5) 
            {\includegraphics[width=0.15\linewidth]{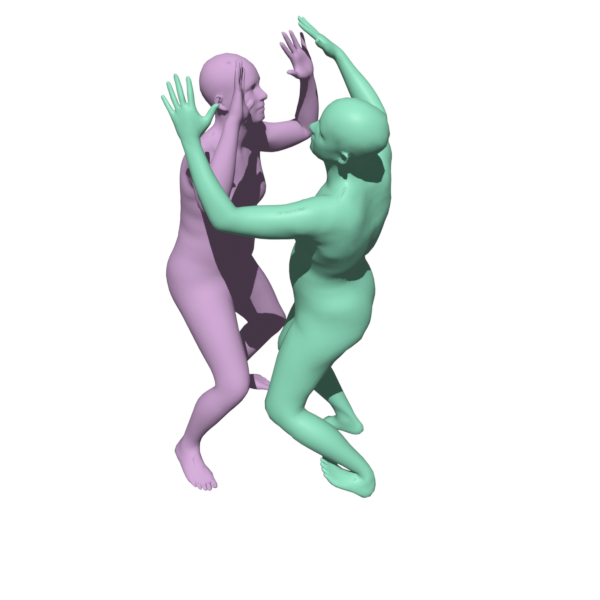}};
            \node (input)[align=center, rotate=90, xshift = -45pt, yshift = 35pt] at (img0.north){Ours};
            \node (input)[align=center, rotate=90, xshift = -95pt, yshift = 50pt] at (img0.north){Sampling (High Five)};
            \node (img10) at (-1,6.5) 
            {\includegraphics[width=0.15\linewidth]{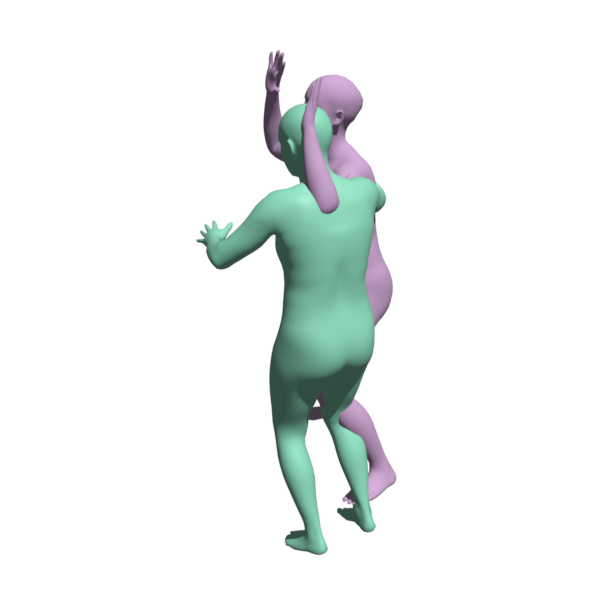}};
            \node (img11) at (1,6.5) 
            {\includegraphics[width=0.15\linewidth]{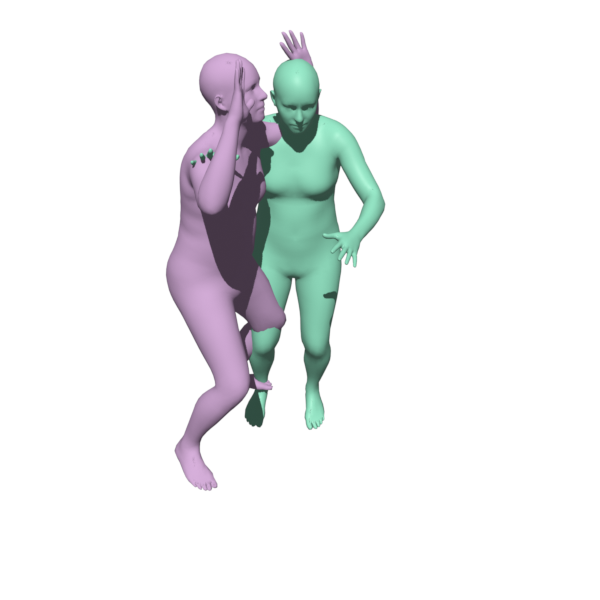}};
            \node (img12) at (3.7,6.5) 
            {\includegraphics[width=0.15\linewidth]{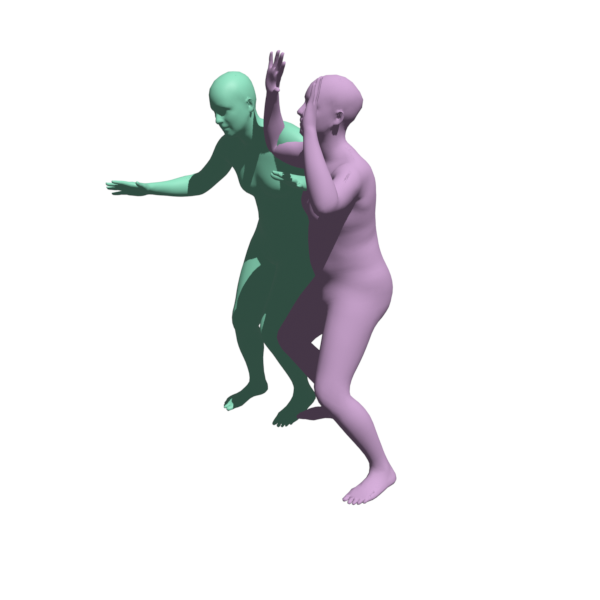}};
            \node (img13) at (5.7,6.5) 
            {\includegraphics[width=0.15\linewidth]{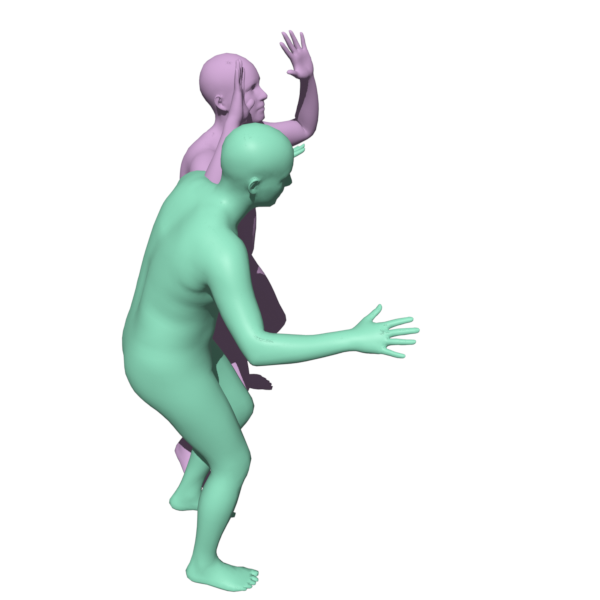}};
            \node (img14) at (8.4,6.5) 
            {\includegraphics[width=0.15\linewidth]{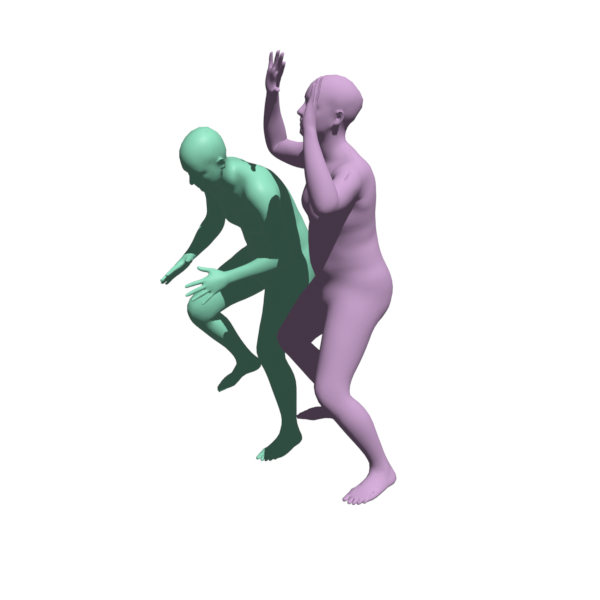}};
            \node (img15) at (10.4,6.5) 
            {\includegraphics[width=0.15\linewidth]{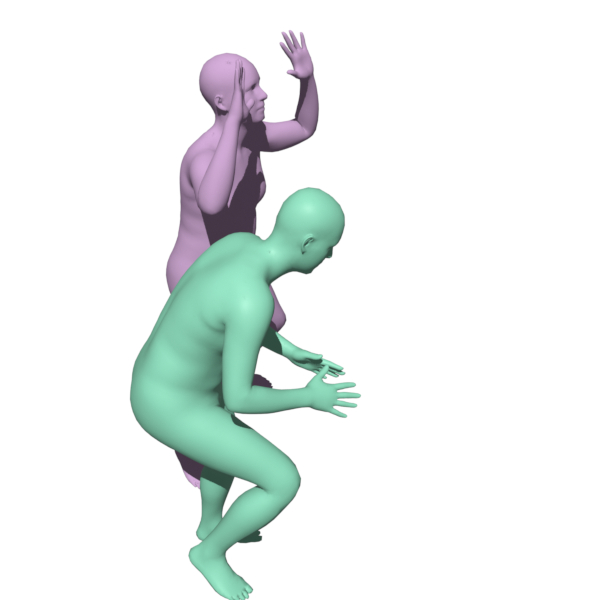}};
            \node (img16) at (13.1,6.5) 
            {\includegraphics[width=0.15\linewidth]{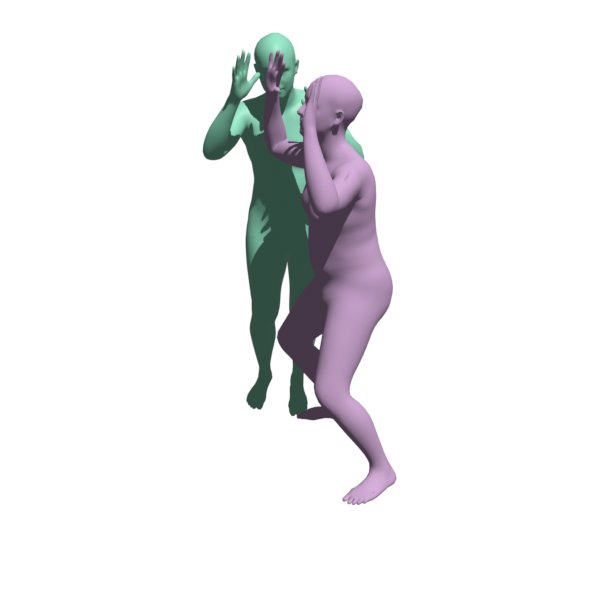}};
            \node (img17) at (15.1,6.5) 
            {\includegraphics[width=0.15\linewidth]{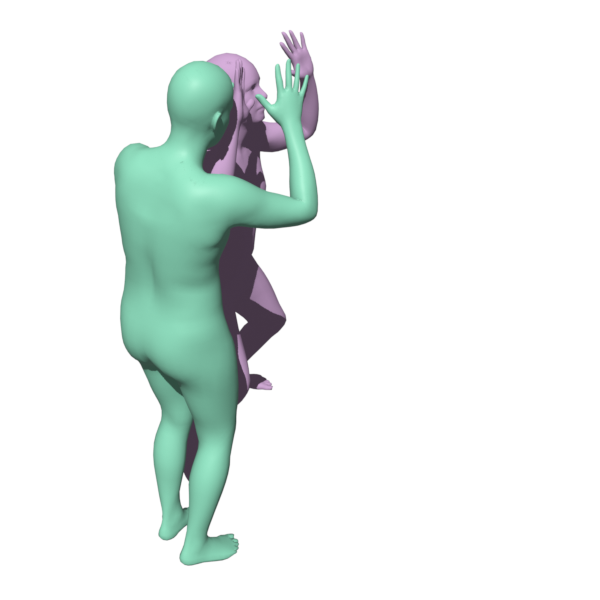}};
            \node (input)[align=center, rotate=90, xshift = -45pt, yshift = 35pt] at (img10.north){BUDDI};
            \node (img20) at (-1,3) 
            {\includegraphics[width=0.15\linewidth]{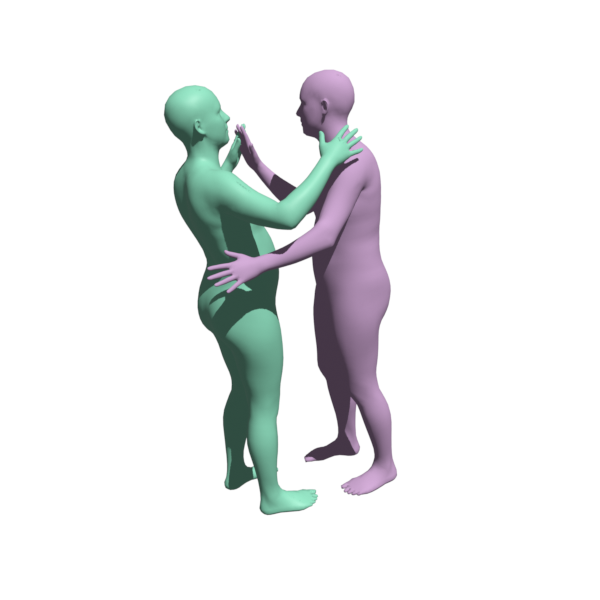}};
            \node (img21) at (1,3) 
            {\includegraphics[width=0.15\linewidth]{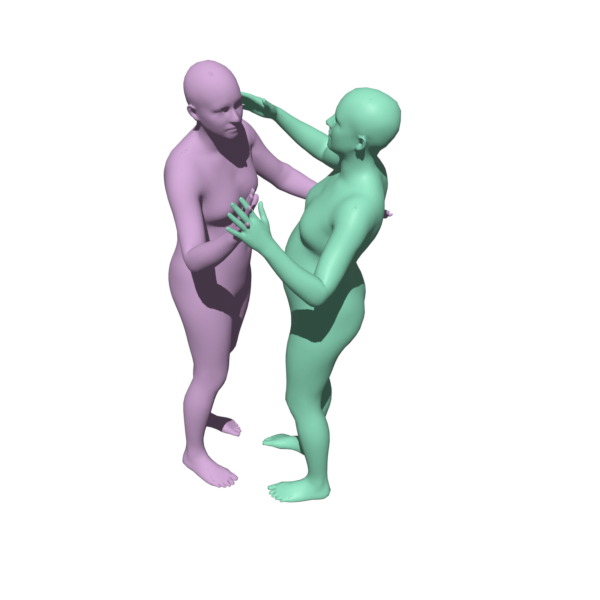}};
            \node (img22) at (3.7,3) 
            {\includegraphics[width=0.15\linewidth]{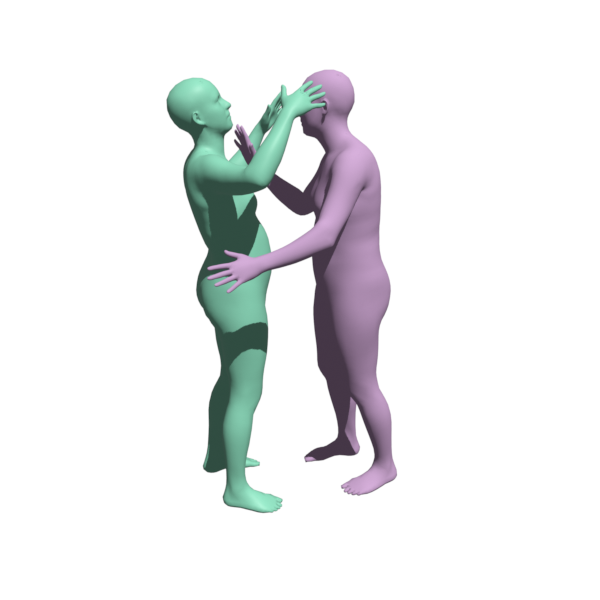}};
            \node (img23) at (5.7,3) 
            {\includegraphics[width=0.15\linewidth]{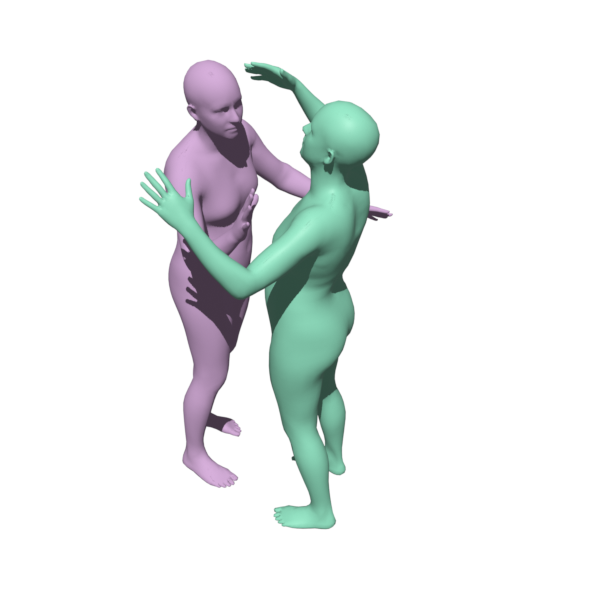}};
            \node (img24) at (8.4,3) 
            {\includegraphics[width=0.15\linewidth]{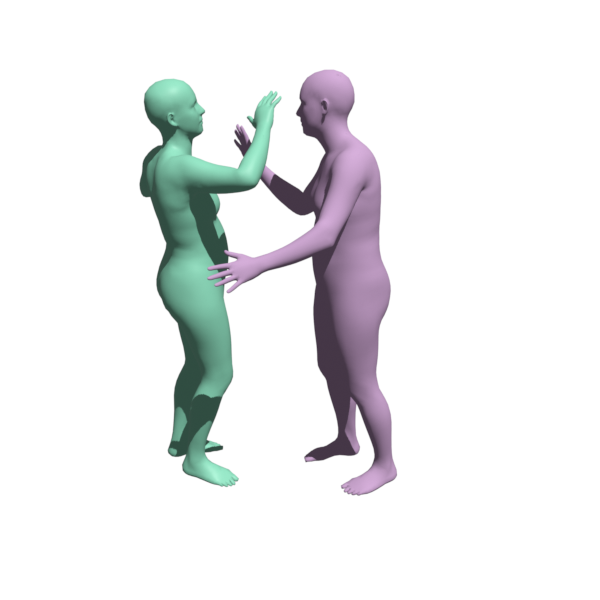}};
            \node (img25) at (10.4,3) 
            {\includegraphics[width=0.15\linewidth]{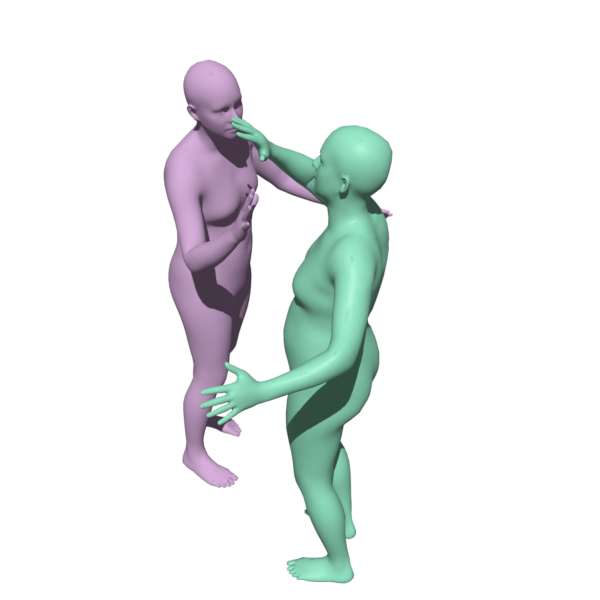}};
            \node (img26) at (13.1,3) 
            {\includegraphics[width=0.15\linewidth]{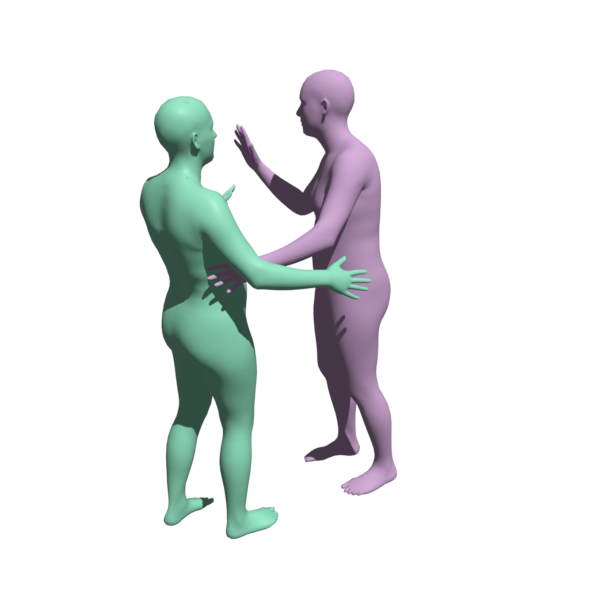}};
            \node (img27) at (15.1,3) 
            {\includegraphics[width=0.15\linewidth]{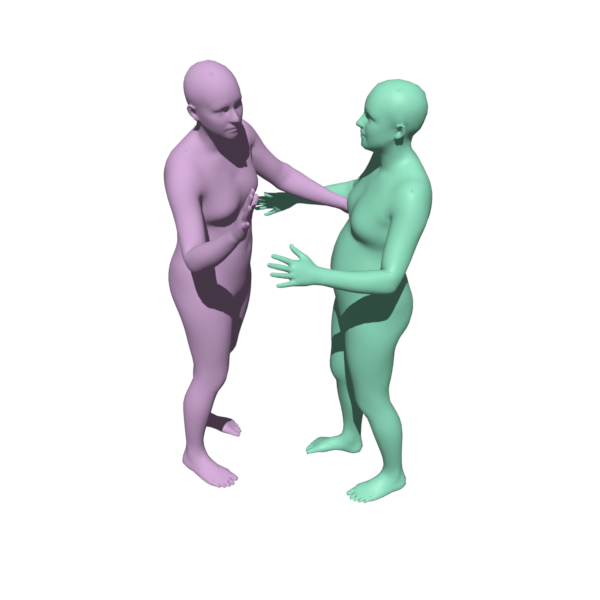}};
            \node (input)[align=center, rotate=90, xshift = -45pt, yshift = 35pt] at (img20.north){Ours};
            \node (input)[align=center, rotate=90, xshift = -95pt, yshift = 50pt] at (img20.north){Sampling (Dance)};
            \node (img20) at (-1,0) 
            {\includegraphics[width=0.15\linewidth]{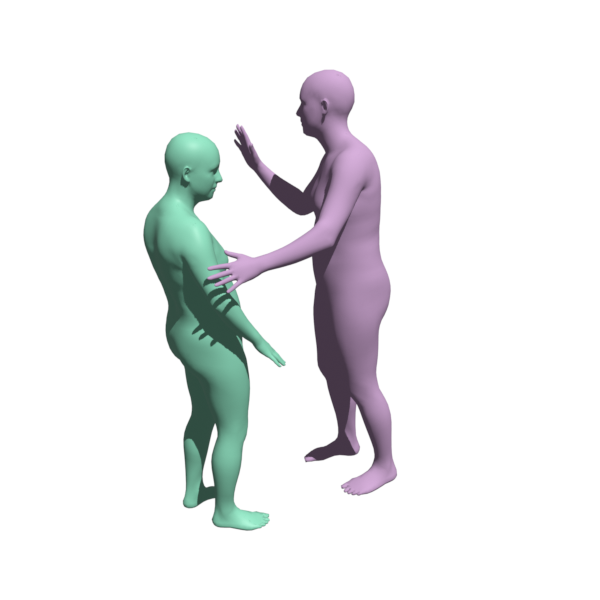}};
            \node (img21) at (1,0) 
            {\includegraphics[width=0.15\linewidth]{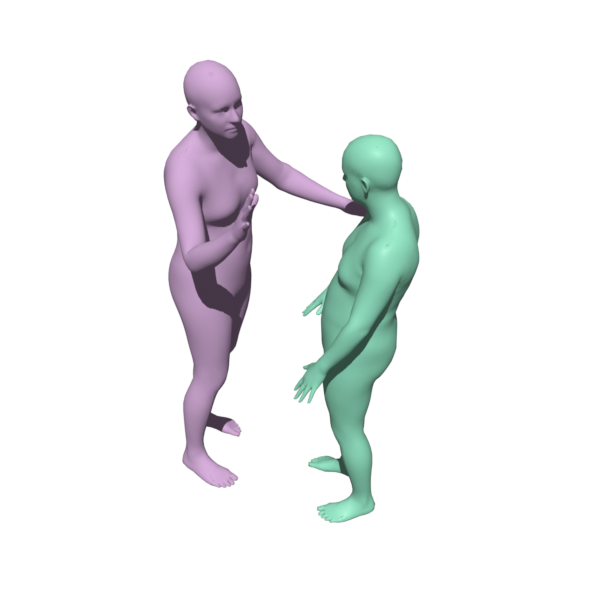}};
            \node (img22) at (3.7,0) 
            {\includegraphics[width=0.15\linewidth]{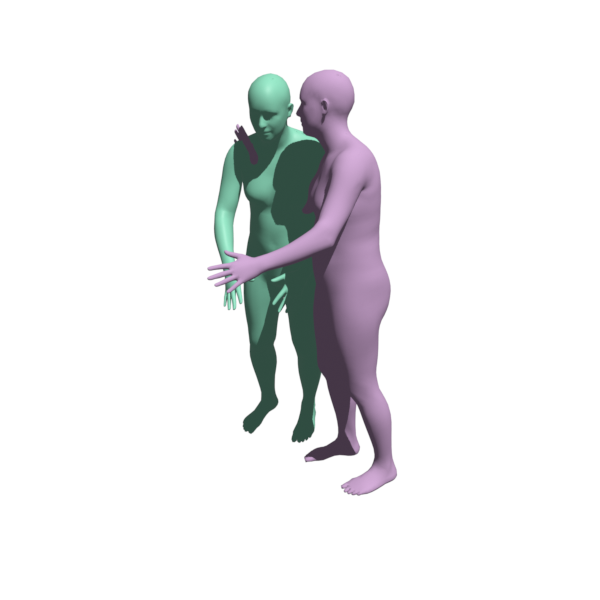}};
            \node (img23) at (5.7,0) 
            {\includegraphics[width=0.15\linewidth]{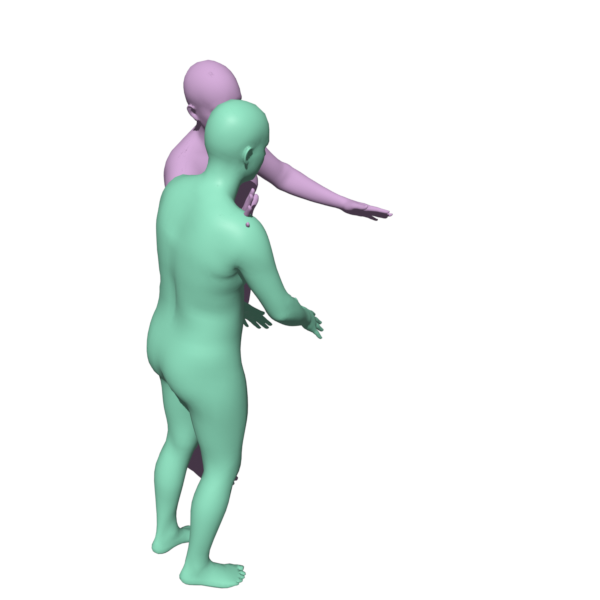}};
            \node (img24) at (8.4,0) 
            {\includegraphics[width=0.15\linewidth]{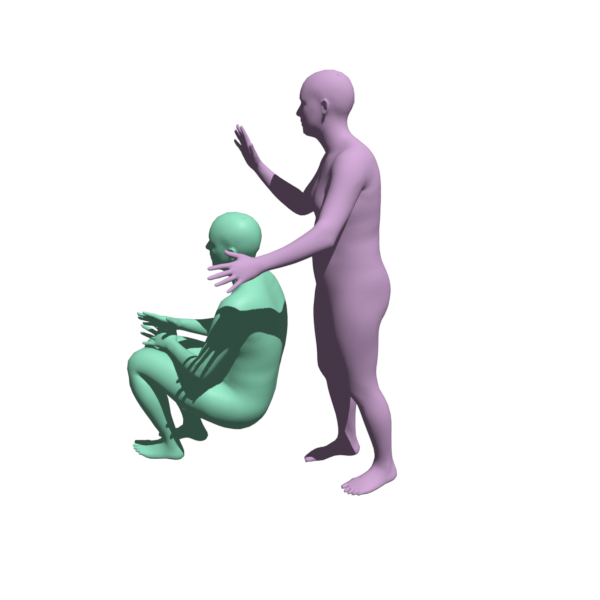}};
            \node (img25) at (10.4,0) 
            {\includegraphics[width=0.15\linewidth]{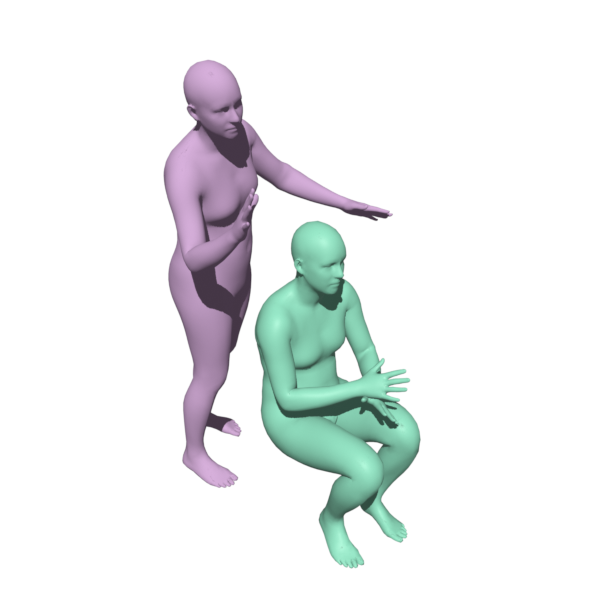}};
            \node (img26) at (13.1,0) 
            {\includegraphics[width=0.15\linewidth]{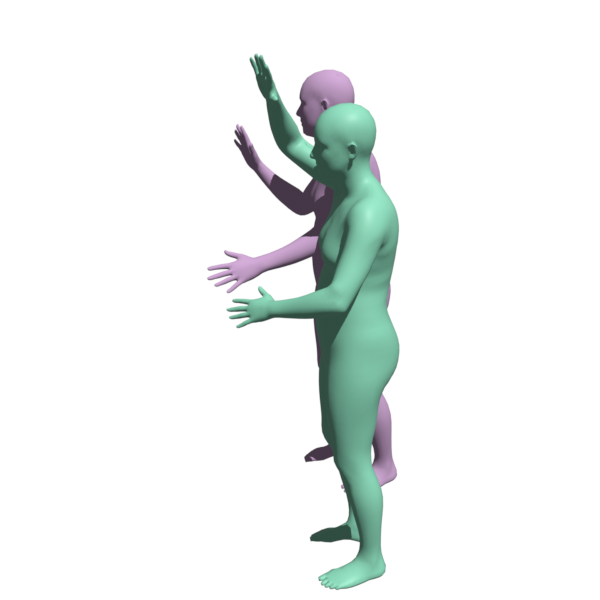}};
            \node (img27) at (15.1,0) 
            {\includegraphics[width=0.15\linewidth]{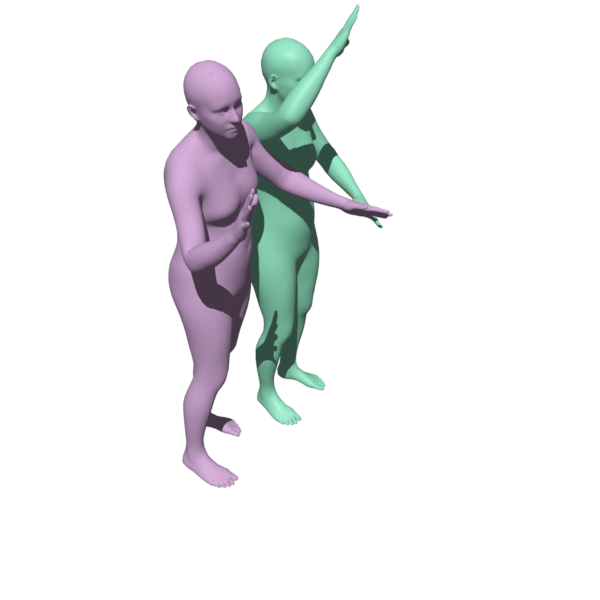}};
            \node (input)[align=center, rotate=90, xshift = -45pt, yshift = 35pt] at (img20.north){BUDDI};
\end{tikzpicture}
    }
    \caption{
    Examples of conditional sampling from our model and BUDDI~\cite{mueller2024buddi} for a set of \conditioning{conditioning pose}, labeled as \textit{dance}, from the Hi4D dataset. These \conditioning{conditioning poses} belong to the in-distribution benchmark. We show two views and four different samples, organized in four columns.  Our generated \reaction{reaction poses} respond naturally to the input, producing realistic human-human 3D interactions that are semantically consistent with the input poses. In contrast, BUDDI \reaction{reaction poses} lack contextual coherence and frequently display noticeable interpenetrations.
    }
    \label{fig:qualitative-results}
\vspace{-0.5cm}
\end{figure*}

In Figure~\ref{fig:qualitative-results}, we compare the expressivity of our model and BUDDI by sampling diverse \reaction{reaction poses} conditioned on a \conditioning{conditioning pose} from the Hi4D \textit{dance} category (See the Supplementary Material for more examples).
Results demonstrate that our model generates rich and semantically consistent responses. In contrast, BUDDI often lacks contextual coherence and shows interpenetrations.

Figures~\ref{fig:teaser} and \ref{fig:controlnet} showcase applications of our method, highlighting how our increased controllability enables easy 
control for text-conditioned image generative tools like FLUX.1-Depth-dev~\cite{flux2024}, which yields accurate pose control human synthesis. Please refer to the Supplementary Material for more examples.

\subsection{User Study}
To evaluate the perceptual plausibility of our generated interactions, we conducted a user study with 35 participants (19 female, 16 male), aged 18 to 58 years ($\mu = 32.2$ years, $\sigma = 11.4$ years), rating 3D dyadic scenes on a 5-point Likert scale (1 = Very Implausible, 5 = Very Plausible).

We designed the study around 10 base interactions, each presented under three conditions: 
ground truth, augmented (i.e. samples from Section \ref{sec:dataset}), and generated with our method.
Critically, the \conditioning{conditioning avatar} is held identical across all three conditions for a given interaction, ensuring that any difference in plausibility ratings is driven solely by the reacting avatar.
Each participant evaluated 30 scenes, distributed across three randomized blocks to prevent immediate cross-condition comparison.
Participants interacted with a WebGL viewer that allowed free camera rotation, panning, and zooming to resolve spatial ambiguities before rating.

Mean plausibility scores were $3.97 \pm 0.76$ for ground truth, $3.91 \pm 0.78$ for augmented poses, and $4.02 \pm 0.69$ for generated poses.
To test whether generated poses are perceived as equivalent to ground truth, rather than merely not different, we applied equivalence testing (TOST) with a margin of $\delta = 0.5$ points. 
The results confirm perceptual equivalence between generated and ground truth poses ($p < 0.0001$), between augmented and ground truth poses ($p < 0.0001$), and between generated and augmented poses ($p < 0.0001$).
These findings demonstrate that our model produces interactions that are indistinguishable in perceived plausibility from real captured data.

\section{Conclusions}
This work presented a novel approach for human-human interaction generation.
Our model enables the 3D collision-aware synthesis of humans in interaction, conditioned on one pose.
We introduced a new data augmentation strategy that leverages existing datasets to reduce the need for extensive data collection, thus reducing the dependency on costly motion capture systems and post-processing pipelines.
This democratizes the creation of high-quality human-human interaction data for a broader range of users.

We also demonstrate that traversing our latent space results in smooth transitions across diverse poses, which can become a powerful tool for artists to fine-tune their creations. Furthermore, our generated \reaction{reaction poses} consistently exhibit greater contextual coherence and fewer interpenetrations compared to those produced by BUDDI~\cite{mueller2024buddi}.

A potential open path for exploration is integrating our approach as a post-processing step in existing methods for tracking and reconstructing 3D humans \cite{sun2022bev,mueller2024buddi}, which would provide a greater variety and enhanced controllability of the output pose.
Additionally, incorporating our model into text-to-motion generative methods would enable the synthesis of realistic animations of humans in contact. 

Despite improving 3D interaction synthesis, our model currently handles only static poses. Since the trajectory leading to a pose often defines its semantic meaning, future work involves extending our data augmentation and architecture to incorporate time-dependent information for dynamic interactions. \new{Furthermore, since each augmented pose is treated as a quasi-static configuration, we currently do not employ Continuous Collision Detection (CCD) in our data generation stage. While our SDF-based approach effectively resolves static interpenetrations, interpreting the augmentation as a continuous deformation and applying CCD could better preserve first-contact configurations, particularly for thin structures such as fingers.} \new{Beyond physical and temporal constraints, our stochastic augmentation lacks explicit semantic limits, which could theoretically generate socially awkward interactions. However, our user studies did not identify any such problematic cases. Explicitly enforcing social plausibility remains an interesting area for further exploration.}

Additionally, while our $\capfix$ module resolves geometric interpenetration via volumetric proxies, it is not a full physics simulation. It does not account for secondary dynamics like muscle deformation, soft-body contact, or gravity-induced balance. In rare cases of extreme mesh entanglement, the module may converge to local minima. We thus view $\capfix$ as a geometric prerequisite for realism, rather than a replacement for high-fidelity physics solvers~\cite{zhang2023simulation}. \new{Finally, this module is explicitly designed to reduce inter-body collisions and does not include a separate self-contact loss, meaning the matrix is computed across the two bodies rather than within a single body. However, because our data augmentation stage successfully resolves self-penetrations using SDFs (as detailed in Section \ref{sec:collision-solving}), the model does not observe self-collisions during training. Consequently, the network rarely produces outputs with noticeable self-penetrations in practice. Incorporating an explicit self-collision term during this refinement stage remains a potential extension for future work.}

\section*{Acknowledgments}
This work has been partially funded by the Comunidad de Madrid through two initiatives. First, in the framework of the Multiannual Agreement with the Universidad Rey Juan Carlos in line of Action 1, "Estímulo a la investigación de jóvenes doctores", for the project "Captura de humanos restringida por sus entornos" (Acronym: CaptHuRe) with reference M2736. Second, the presented results are also part of the project "Inteligencia artificial para la industria 4.0: generación de datos, modelado avanzado optimización e interpretabilidad" (Acronym: IDEA-CM), with reference TEC-2024/COM-89, funded through the call for grants for collaborative R\&D projects under the modality of "Programas de Actividades de I+D en Tecnologías 2024", according to Order 3177/2024. 

\begin{center}
\includegraphics[width=0.9\linewidth]{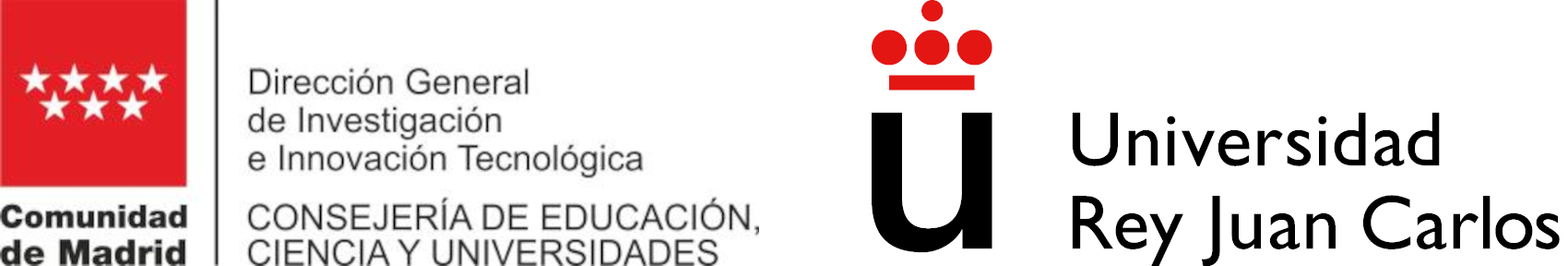}
\end{center}

\bibliographystyle{eg-alpha-doi}  
\bibliography{main}

@String(CVPR= {IEEE Conf. Comput. Vis. Pattern Recog.})

@String(ICCV= {Int. Conf. Comput. Vis.})

@String(ECCV= {Eur. Conf. Comput. Vis.})

@String(NIPS= {Adv. Neural Inform. Process. Syst.})

@String(CVPR  = {CVPR})

@String(ICCV  = {ICCV})

@String(ECCV  = {ECCV})

@String(NIPS  = {NeurIPS})

@InProceedings{ghosh2025duetgen,
            title={{DuetGen: Music Driven Two-Person Dance Generation via Hierarchical Masked Modeling}},
            author={Ghosh, Anindita and Zhou, Bing and Dabral, Rishabh and Wang, Jian and Golyanik, Vladislav and Theobalt, Christian and Slusallek, Philipp and Guo, Chuan},
            booktitle = {ACM SIGGRAPH Conference Proceedings},
            year={2025}
        }

@inproceedings{multiply,
      title={MultiPly: Reconstruction of Multiple People from Monocular Video in the Wild}, 
      author={Jiang, Zeren and Guo, Chen and Kaufmann, Manuel and Jiang, Tianjian and Valentin, Julien and Hilliges, Otmar and Song, Jie},
      booktitle = {Proceedings of the IEEE/CVF Conference on Computer Vision and Pattern Recognition (CVPR)},
      year      = {2024},
  }

@inproceedings{liu2025ponimator,
  author    = {Liu, Shaowei and Guo, Chuan and Zhou, Bing and Wang, Jian},
  title     = {Ponimator: Unfolding Interactive Pose for Versatile Human-Human Interaction Animation},
  booktitle   = ICCV,
  year      = {2025},
}

@inproceedings{javed2025intermask,
    title={{InterMask: 3D Human Interaction Generation via Collaborative Masked Modeling}},
    author={Muhammad Gohar Javed and Chuan Guo and Li Cheng and Xingyu Li},
    booktitle={The Thirteenth International Conference on Learning Representations},
    year={2025},
    pages={1--22},
    publisher={OpenReview.net},
    address={Singapore},
    url={https://openreview.net/forum?id=ZAyuwJYN8N}
}

@article{liang2024intergen,
    author = {Liang, Han and Zhang, Wenqian and Li, Wenxuan and Yu, Jingyi and Xu, Lan},
    title = {InterGen: Diffusion-Based Multi-human Motion Generation Under Complex Interactions},
    year = {2024},
    issue_date = {Sep 2024},
    publisher = {Kluwer Academic Publishers},
    address = {USA},
    volume = {132},
    number = {9},
    issn = {0920-5691},
    url = {https://doi.org/10.1007/s11263-024-02042-6},
    doi = {10.1007/s11263-024-02042-6},
    journal = {Int. J. Comput. Vision},
    month = mar,
    pages = {3463–3483},
    numpages = {21}
}

@inproceedings{fan2024freemotion,
    author="Fan, Ke
    and Tang, Junshu
    and Cao, Weijian
    and Yi, Ran
    and Li, Moran
    and Gong, Jingyu
    and Zhang, Jiangning
    and Wang, Yabiao
    and Wang, Chengjie
    and Ma, Lizhuang",
    editor="Leonardis, Ale{\v{s}}
    and Ricci, Elisa
    and Roth, Stefan
    and Russakovsky, Olga
    and Sattler, Torsten
    and Varol, G{\"u}l",
    title="FreeMotion: A Unified Framework for Number-Free Text-to-Motion Synthesis",
    booktitle="Computer Vision -- ECCV 2024",
    year="2025",
    publisher="Springer Nature Switzerland",
    address="Cham",
    pages="93--109",
    isbn="978-3-031-73242-3"
}

@inproceedings{shan2024multiperson,
    author="Shan, Mengyi
    and Dong, Lu
    and Han, Yutao
    and Yao, Yuan
    and Liu, Tao
    and Nwogu, Ifeoma
    and Qi, Guo-Jun
    and Hill, Mitch",
    editor="Leonardis, Ale{\v{s}}
    and Ricci, Elisa
    and Roth, Stefan
    and Russakovsky, Olga
    and Sattler, Torsten
    and Varol, G{\"u}l",
    title="Towards Open Domain Text-Driven Synthesis of Multi-person Motions",
    booktitle="Computer Vision -- ECCV 2024",
    year="2025",
    publisher="Springer Nature Switzerland",
    address="Cham",
    pages="67--86",
    isbn="978-3-031-73650-6"
}

@inproceedings{tanaka2023interaction,
    author={Tanaka, Mikihiro and Fujiwara, Kent},
    booktitle={2023 IEEE/CVF International Conference on Computer Vision (ICCV)}, 
    title={Role-aware Interaction Generation from Textual Description}, 
    year={2023},
    volume={},
    number={},
    pages={15953-15963},
    publisher={IEEE},
    address={Paris, France},
    doi={10.1109/ICCV51070.2023.01466}
}

@article{chopin2023interaction,
    author={Chopin, Baptiste and Tang, Hao and Otberdout, Naima and Daoudi, Mohamed and Sebe, Nicu},
    journal={IEEE Transactions on Multimedia}, 
    title={{Interaction Transformer for Human Reaction Generation}}, 
    year={2023},
    volume={25},
    number={},
    pages={8842-8854},
    doi={10.1109/TMM.2023.3242152}
}

@article{shuai2023reconstructing,
    author = {Shuai, Qing and Yu, Zhiyuan and Zhou, Zhize and Fan, Lixin and Yang, Haijun and Yang, Can and Zhou, Xiaowei},
    title = {Reconstructing Close Human Interactions from Multiple Views},
    year = {2023},
    issue_date = {December 2023},
    publisher = {Association for Computing Machinery},
    address = {New York, NY, USA},
    volume = {42},
    number = {6},
    issn = {0730-0301},
    url = {https://doi.org/10.1145/3618336},
    doi = {10.1145/3618336},
    journal = {ACM Trans. Graph.},
    month = dec,
    articleno = {273},
    numpages = {14}
}

@inproceedings{mueller2018ganerated,
    author={Mueller, Franziska and Bernard, Florian and Sotnychenko, Oleksandr and Mehta, Dushyant and Sridhar, Srinath and Casas, Dan and Theobalt, Christian},
    booktitle={2018 IEEE/CVF Conference on Computer Vision and Pattern Recognition}, 
    title={GANerated Hands for Real-Time 3D Hand Tracking from Monocular RGB}, 
    year={2018},
    volume={},
    number={},
    pages={49-59},
    publisher={IEEE},
    address={Salt Lake City, UT, USA},
    doi={10.1109/CVPR.2018.00013}
}

@inproceedings{sohn2015learning,
    author = {Sohn, Kihyuk and Yan, Xinchen and Lee, Honglak},
    title = {Learning structured output representation using deep conditional generative models},
    year = {2015},
    publisher = {MIT Press},
    address = {Cambridge, MA, USA},
    booktitle = {Proceedings of the 29th International Conference on Neural Information Processing Systems - Volume 2},
    pages = {3483–3491},
    numpages = {9},
    location = {Montreal, Canada},
    series = {NIPS'15}
}

@inproceedings{zhou2019continuity,
    author={Zhou, Yi and Barnes, Connelly and Lu, Jingwan and Yang, Jimei and Li, Hao},
    booktitle={2019 IEEE/CVF Conference on Computer Vision and Pattern Recognition (CVPR)}, 
    title={On the Continuity of Rotation Representations in Neural Networks}, 
    year={2019},
    volume={},
    number={},
    pages={5738-5746},
    publisher={IEEE},
    address={Long Beach, CA, USA},
    doi={10.1109/CVPR.2019.00589}
}

@inproceedings{vposer2019cvpr,
    author={Pavlakos, Georgios and Choutas, Vasileios and Ghorbani, Nima and Bolkart, Timo and Osman, Ahmed A. and Tzionas, Dimitrios and Black, Michael J.},
    booktitle={2019 IEEE/CVF Conference on Computer Vision and Pattern Recognition (CVPR)}, 
    title={Expressive Body Capture: 3D Hands, Face, and Body From a Single Image}, 
    year={2019},
    volume={},
    number={},
    pages={10967-10977},
    publisher={IEEE},
    address={Long Beach, CA, USA},
    doi={10.1109/CVPR.2019.01123}
}

@inproceedings{guzov24ireplica,
    author={Guzov, Vladimir and Chibane, Julian and Marin, Riccardo and He, Yannan and Saracoglu, Yunus and Sattler, Torsten and Pons-Moll, Gerard},
    booktitle={2024 International Conference on 3D Vision (3DV)}, 
    title={Interaction Replica: Tracking Human–Object Interaction and Scene Changes From Human Motion}, 
    year={2024},
    volume={},
    number={},
    pages={1006-1016},
    publisher={IEEE},
    address={Davos, Switzerland},
    doi={10.1109/3DV62453.2024.00072}
}

@inproceedings{hassan2021populating,
    author={Hassan, Mohamed and Ghosh, Partha and Tesch, Joachim and Tzionas, Dimitrios and Black, Michael J.},
    booktitle={2021 IEEE/CVF Conference on Computer Vision and Pattern Recognition (CVPR)}, 
    title={Populating 3D Scenes by Learning Human-Scene Interaction}, 
    year={2021},
    volume={},
    number={},
    pages={14703-14713},
    publisher={IEEE},
    address={Nashville, TN, USA},
    doi={10.1109/CVPR46437.2021.01447}
}

@inproceedings{jiang2024scaling,
    author={Jiang, Nan and Zhang, Zhiyuan and Li, Hongjie and Ma, Xiaoxuan and Wang, Zan and Chen, Yixin and Liu, Tengyu and Zhu, Yixin and Huang, Siyuan},
    booktitle={2024 IEEE/CVF Conference on Computer Vision and Pattern Recognition (CVPR)}, 
    title={Scaling Up Dynamic Human-Scene Interaction Modeling}, 
    year={2024},
    volume={},
    number={},
    pages={1737-1747},
    publisher={IEEE},
    address={Seattle, WA, USA},
    doi={10.1109/CVPR52733.2024.00171}
}

@inproceedings{ye2023slahmr,
    author={Ye, Vickie and Pavlakos, Georgios and Malik, Jitendra and Kanazawa, Angjoo},
    booktitle={2023 IEEE/CVF Conference on Computer Vision and Pattern Recognition (CVPR)}, 
    title={Decoupling Human and Camera Motion from Videos in the Wild}, 
    year={2023},
    volume={},
    number={},
    pages={21222-21232},
    publisher={IEEE},
    address={Vancouver, BC, Canada},
    doi={10.1109/CVPR52729.2023.02033}
}

@inproceedings{ugrinovic2024multiphys,
    author={Ugrinovic, Nicolas and Pan, Boxiao and Pavlakos, Georgios and Paschalidou, Despoina and Shen, Bokui and Sanchez-Riera, Jordi and Moreno-Noguer, Francesc and Guibas, Leonidas},
    booktitle={2024 IEEE/CVF Conference on Computer Vision and Pattern Recognition (CVPR)}, 
    title={MultiPhys: Multi-Person Physics-Aware 3D Motion Estimation}, 
    year={2024},
    volume={},
    number={},
    pages={2331-2340},
    publisher={IEEE},
    address={Seattle, WA, USA},
    doi={10.1109/CVPR52733.2024.00226}
}

@inproceedings{yin2023hi4d,
    author={Yin, Yifei and Guo, Chen and Kaufmann, Manuel and Zarate, Juan Jose and Song, Jie and Hilliges, Otmar},
    booktitle={2023 IEEE/CVF Conference on Computer Vision and Pattern Recognition (CVPR)}, 
    title={Hi4D: 4D Instance Segmentation of Close Human Interaction}, 
    year={2023},
    volume={},
    number={},
    pages={17016-17027},
    publisher={IEEE},
    address={Vancouver, BC, Canada},
    doi={10.1109/CVPR52729.2023.01632}
}

@inproceedings{fieraru2020three,
    author={Fieraru, Mihai and Zanfir, Mihai and Oneata, Elisabeta and Popa, Alin-Ionut and Olaru, Vlad and Sminchisescu, Cristian},
    booktitle={2020 IEEE/CVF Conference on Computer Vision and Pattern Recognition (CVPR)}, 
    title={Three-Dimensional Reconstruction of Human Interactions}, 
    year={2020},
    volume={},
    number={},
    pages={7212-7221},
    publisher={IEEE},
    address={Seattle, WA, USA},
    doi={10.1109/CVPR42600.2020.00724}
}

@inproceedings{mueller2024buddi,
    author={Müller, Lea and Ye, Vickie and Pavlakos, Georgios and Black, Michael and Kanazawa, Angjoo},
    booktitle={2024 IEEE/CVF Conference on Computer Vision and Pattern Recognition (CVPR)}, 
    title={Generative Proxemics: A Prior for 3D Social Interaction from Images}, 
    year={2024},
    volume={},
    number={},
    pages={9687-9697},
    publisher={IEEE},
    address={Seattle, WA, USA},
    doi={10.1109/CVPR52733.2024.00925}
}

@inproceedings{mueller2021tuch,
    author={Müller, Lea and Osman, Ahmed A. A. and Tang, Siyu and Huang, Chun-Hao P. and Black, Michael J.},
    booktitle={2021 IEEE/CVF Conference on Computer Vision and Pattern Recognition (CVPR)}, 
    title={On Self-Contact and Human Pose}, 
    year={2021},
    volume={},
    number={},
    pages={9985-9994},
    publisher={IEEE},
    address={Nashville, TN, USA},
    doi={10.1109/CVPR46437.2021.00986}
}

@inproceedings{hasson2019learning,
    author={Hasson, Yana and Varol, Gül and Tzionas, Dimitrios and Kalevatykh, Igor and Black, Michael J. and Laptev, Ivan and Schmid, Cordelia},
    booktitle={2019 IEEE/CVF Conference on Computer Vision and Pattern Recognition (CVPR)}, 
    title={Learning Joint Reconstruction of Hands and Manipulated Objects}, 
    year={2019},
    volume={},
    number={},
    pages={11799-11808},
    publisher={IEEE},
    address={Long Beach, CA, USA},
    doi={10.1109/CVPR.2019.01208}
}

@inproceedings{ye2023ghop,
    author={Ye, Yufei and Gupta, Abhinav and Kitani, Kris and Tulsiani, Shubham},
    booktitle={2024 IEEE/CVF Conference on Computer Vision and Pattern Recognition (CVPR)}, 
    title={G-HOP: Generative Hand-Object Prior for Interaction Reconstruction and Grasp Synthesis}, 
    year={2024},
    volume={},
    number={},
    pages={1911-1920},
    publisher={IEEE},
    address={Seattle, WA, USA},
    doi={10.1109/CVPR52733.2024.00187}
}

@article{loper2015smpl,
    author = {Loper, Matthew and Mahmood, Naureen and Romero, Javier and Pons-Moll, Gerard and Black, Michael J.},
    title = {SMPL: a skinned multi-person linear model},
    year = {2015},
    issue_date = {November 2015},
    publisher = {Association for Computing Machinery},
    address = {New York, NY, USA},
    volume = {34},
    number = {6},
    issn = {0730-0301},
    url = {https://doi.org/10.1145/2816795.2818013},
    doi = {10.1145/2816795.2818013},
    journal = {ACM Trans. Graph.},
    month = oct,
    articleno = {248},
    numpages = {16}
}

@inproceedings{sun2022bev,
    author={Sun, Yu and Liu, Wu and Bao, Qian and Fu, Yili and Mei, Tao and Black, Michael J.},
    booktitle={2022 IEEE/CVF Conference on Computer Vision and Pattern Recognition (CVPR)}, 
    title={Putting People in their Place: Monocular Regression of 3D People in Depth}, 
    year={2022},
    volume={},
    number={},
    pages={13233-13242},
    publisher={IEEE},
    address={New Orleans, LA, USA},
    doi={10.1109/CVPR52688.2022.01289}
}

@inproceedings{sun2021romp,
    author={Sun, Yu and Bao, Qian and Liu, Wu and Fu, Yili and Black, Michael J. and Mei, Tao},
    booktitle={2021 IEEE/CVF International Conference on Computer Vision (ICCV)}, 
    title={Monocular, One-stage, Regression of Multiple 3D People}, 
    year={2021},
    volume={},
    number={},
    pages={11159-11168},
    publisher={IEEE},
    address={Montreal, QC, Canada},
    doi={10.1109/ICCV48922.2021.01099}
}

@inproceedings{guo2022expi,
    author={Guo, Wen and Bie, Xiaoyu and Alameda-Pineda, Xavier and Moreno–Noguer, Francesc},
    booktitle={2022 IEEE/CVF Conference on Computer Vision and Pattern Recognition (CVPR)}, 
    title={Multi-Person Extreme Motion Prediction}, 
    year={2022},
    volume={},
    number={},
    pages={13043-13054},
    publisher={IEEE},
    address={New Orleans, LA, USA},
    doi={10.1109/CVPR52688.2022.01271}
}

@inproceedings{zheng2021deepmulticap,
    author={Zheng, Yang and Shao, Ruizhi and Zhang, Yuxiang and Yu, Tao and Zheng, Zerong and Dai, Qionghai and Liu, Yebin},
    booktitle={2021 IEEE/CVF International Conference on Computer Vision (ICCV)}, 
    title={DeepMultiCap: Performance Capture of Multiple Characters Using Sparse Multiview Cameras}, 
    year={2021},
    volume={},
    number={},
    pages={6219-6229},
    publisher={IEEE},
    address={Montreal, QC, Canada},
    doi={10.1109/ICCV48922.2021.00618}
}

@inproceedings{christen2022dgrasp,
    author={Christen, Sammy and Kocabas, Muhammed and Aksan, Emre and Hwangbo, Jemin and Song, Jie and Hilliges, Otmar},
    booktitle={2022 IEEE/CVF Conference on Computer Vision and Pattern Recognition (CVPR)}, 
    title={D-Grasp: Physically Plausible Dynamic Grasp Synthesis for Hand-Object Interactions}, 
    year={2022},
    volume={},
    number={},
    pages={20545-20554},
    publisher={IEEE},
    address={New Orleans, LA, USA},
    doi={10.1109/CVPR52688.2022.01992}
}

@inproceedings{xie2022chore,
    author = {Xie, Xianghui and Bhatnagar, Bharat Lal and Pons-Moll, Gerard},
    title = {CHORE: Contact, Human and Object Reconstruction from a Single RGB Image},
    year = {2022},
    isbn = {978-3-031-20085-4},
    publisher = {Springer-Verlag},
    address = {Berlin, Heidelberg},
    url = {https://doi.org/10.1007/978-3-031-20086-1_8},
    doi = {10.1007/978-3-031-20086-1_8},
    booktitle = {Computer Vision – ECCV 2022: 17th European Conference, Tel Aviv, Israel, October 23–27, 2022, Proceedings, Part II},
    pages = {125–145},
    numpages = {21},
    location = {Tel Aviv, Israel}
}

@inproceedings{taheri2021goal,
    author={Taheri, Omid and Choutas, Vasileios and Black, Michael J. and Tzionas, Dimitrios},
    booktitle={2022 IEEE/CVF Conference on Computer Vision and Pattern Recognition (CVPR)}, 
    title={GOAL: Generating 4D Whole-Body Motion for Hand-Object Grasping}, 
    year={2022},
    volume={},
    number={},
    pages={13253-13263},
    publisher={IEEE},
    address={New Orleans, LA, USA},
    doi={10.1109/CVPR52688.2022.01291}
}

@inproceedings{mihajlovic2022coap,
    author={Mihajlovic, Marko and Saito, Shunsuke and Bansal, Aayush and Zollhoefer, Michael and Tang, Siyu},
    booktitle={2022 IEEE/CVF Conference on Computer Vision and Pattern Recognition (CVPR)}, 
    title={COAP: Compositional Articulated Occupancy of People}, 
    year={2022},
    volume={},
    number={},
    pages={13191-13200},
    publisher={IEEE},
    address={New Orleans, LA, USA},
    doi={10.1109/CVPR52688.2022.01285}
}

@inproceedings{fieraru2021remips,
    author = {Fieraru, Mihai and Zanfir, Mihai and Szente, Teodor Alexandru and Bazavan, Eduard Gabriel and Olaru, Vlad and Sminchisescu, Cristian},
    title = {REMIPS: physically consistent 3D reconstruction of multiple interacting people under weak supervision},
    year = {2021},
    isbn = {9781713845393},
    publisher = {Curran Associates Inc.},
    address = {Red Hook, NY, USA},
    booktitle = {Proceedings of the 35th International Conference on Neural Information Processing Systems},
    articleno = {1483},
    numpages = {13},
    series = {NIPS '21}
}

@inproceedings{hassan2021stochastic,
    author={Hassan, Mohamed and Ceylan, Duygu and Villegas, Ruben and Saito, Jun and Yang, Jimei and Zhou, Yi and Black, Michael},
    booktitle={2021 IEEE/CVF International Conference on Computer Vision (ICCV)}, 
    title={Stochastic Scene-Aware Motion Prediction}, 
    year={2021},
    volume={},
    number={},
    pages={11354-11364},
    publisher={IEEE},
    address={Montreal, QC, Canada},
    doi={10.1109/ICCV48922.2021.01118}
}

@inproceedings{xu2024interx,
    author={Xu, Liang and Lv, Xintao and Yan, Yichao and Jin, Xin and Wu, Shuwen and Xu, Congsheng and Liu, Yifan and Zhou, Yizhou and Rao, Fengyun and Sheng, Xingdong and Liu, Yunhui and Zeng, Wenjun and Yang, Xiaokang},
    booktitle={2024 IEEE/CVF Conference on Computer Vision and Pattern Recognition (CVPR)}, 
    title={Inter-X: Towards Versatile Human-Human Interaction Analysis}, 
    year={2024},
    volume={},
    number={},
    pages={22260-22271},
    publisher={IEEE},
    address={Seattle, WA, USA},
    doi={10.1109/CVPR52733.2024.02101}
}

@inproceedings{zhao2023synthesizing,
    author={Zhao, Kaifeng and Zhang, Yan and Wang, Shaofei and Beeler, Thabo and Tang, Siyu},
    booktitle={2023 IEEE/CVF International Conference on Computer Vision (ICCV)}, 
    title={Synthesizing Diverse Human Motions in 3D Indoor Scenes}, 
    year={2023},
    volume={},
    number={},
    pages={14692-14703},
    publisher={IEEE},
    address={Paris, France},
    doi={10.1109/ICCV51070.2023.01354}
}

@inproceedings{zhao2022compositional,
    author="Zhao, Kaifeng
    and Wang, Shaofei
    and Zhang, Yan
    and Beeler, Thabo
    and Tang, Siyu",
    editor="Avidan, Shai
    and Brostow, Gabriel
    and Ciss{\'e}, Moustapha
    and Farinella, Giovanni Maria
    and Hassner, Tal",
    title="Compositional Human-Scene Interaction Synthesis with Semantic Control",
    booktitle="Computer Vision -- ECCV 2022",
    year="2022",
    publisher="Springer Nature Switzerland",
    address="Cham",
    pages="311--327",
    isbn="978-3-031-20068-7"
}

@inproceedings{jiang2023chairs,
    author={Jiang, Nan and Liu, Tengyu and Cao, Zhexuan and Cui, Jieming and Zhang, Zhiyuan and Chen, Yixin and Wang, He and Zhu, Yixin and Huang, Siyuan},
    booktitle={2023 IEEE/CVF International Conference on Computer Vision (ICCV)}, 
    title={Full-Body Articulated Human-Object Interaction}, 
    year={2023},
    volume={},
    number={},
    pages={9331-9342},
    publisher={IEEE},
    address={Paris, France},
    doi={10.1109/ICCV51070.2023.00859}
}

@article{gast2015optimization,
    title={Optimization integrator for large time steps},
    author={Gast, Theodore F and Schroeder, Craig and Stomakhin, Alexey and Jiang, Chenfanfu and Teran, Joseph M},
    journal={IEEE transactions on visualization and computer graphics},
    volume={21},
    number={10},
    pages={1103--1115},
    year={2015},
    publisher={IEEE}
}

@article{kane2000variational,
    author = {Kane, C. and Marsden, J. E. and Ortiz, M. and West, M.},
    title = {Variational integrators and the Newmark algorithm for conservative and dissipative mechanical systems},
    journal = {International Journal for Numerical Methods in Engineering},
    volume = {49},
    number = {10},
    pages = {1295-1325},
    doi = {https://doi.org/10.1002/1097-0207(20001210)49:10<1295::AID-NME993>3.0.CO;2-W},
    year = {2000}
}

@misc{flux2024,
    author={Black Forest Labs},
    title={FLUX},
    year={2024},
    howpublished={\url{https://github.com/black-forest-labs/flux}},
}

@inproceedings{mueller2024hsfm,
    author={Müller, Lea and Choi, Hongsuk and Zhang, Anthony and Yi, Brent and Malik, Jitendra and Kanazawa, Angjoo},
    booktitle={2025 IEEE/CVF Conference on Computer Vision and Pattern Recognition (CVPR)}, 
    title={Reconstructing People, Places, and Cameras}, 
    year={2025},
    volume={},
    number={},
    pages={21948-21958},
    publisher={IEEE},
    address={Nashville, TN, USA},
    doi={10.1109/CVPR52734.2025.02044}
}

@article{maluleke2024synergy,
    author={Maluleke, Vongani H. and M{\"u}ller, Lea and Rajasegaran, Jathushan and Pavlakos, Georgios and Ginosar, Shiry and Kanazawa, Angjoo and Malik, Jitendra},
    title={Synergy and Synchrony in Couple Dances},
    journal={arXiv preprint arXiv:2409.04440},
    year={2024},
    volume={abs/2409.04440},
    pages={1--11},
    doi={10.48550/arXiv.2409.04440},
    url={https://arxiv.org/abs/2409.04440}
}

@inproceedings{liu2006composition,
    author = {Liu, C. Karen and Hertzmann, Aaron and Popovi\'{c}, Zoran},
    title = {Composition of complex optimal multi-character motions},
    year = {2006},
    isbn = {3905673347},
    publisher = {Eurographics Association},
    address = {Goslar, DEU},
    booktitle = {Proceedings of the 2006 ACM SIGGRAPH/Eurographics Symposium on Computer Animation},
    pages = {215–222},
    numpages = {8},
    location = {Vienna, Austria},
    series = {SCA '06}
}

@inproceedings{ho2010spatial,
    author = {Ho, Edmond S. L. and Komura, Taku and Tai, Chiew-Lan},
    title = {Spatial relationship preserving character motion adaptation},
    year = {2010},
    isbn = {9781450302104},
    publisher = {Association for Computing Machinery},
    address = {New York, NY, USA},
    url = {https://doi.org/10.1145/1833349.1778770},
    doi = {10.1145/1833349.1778770},
    booktitle = {ACM SIGGRAPH 2010 Papers},
    articleno = {33},
    numpages = {8},
    location = {Los Angeles, California},
    series = {SIGGRAPH '10}
}

@inproceedings{kim2012tiling,
    author = {Kim, Manmyung and Hwang, Youngseok and Hyun, Kyunglyul and Lee, Jehee},
    title = {Tiling motion patches},
    year = {2012},
    isbn = {9783905674378},
    publisher = {Eurographics Association},
    address = {Goslar, DEU},
    booktitle = {Proceedings of the ACM SIGGRAPH/Eurographics Symposium on Computer Animation},
    pages = {117–126},
    numpages = {10},
    location = {Lausanne, Switzerland},
    series = {SCA '12}
}

@article{won2014generating,
    author = {Won, Jungdam and Lee, Kyungho and O'Sullivan, Carol and Hodgins, Jessica K. and Lee, Jehee},
    title = {Generating and ranking diverse multi-character interactions},
    year = {2014},
    issue_date = {November 2014},
    publisher = {Association for Computing Machinery},
    address = {New York, NY, USA},
    volume = {33},
    number = {6},
    issn = {0730-0301},
    url = {https://doi.org/10.1145/2661229.2661271},
    doi = {10.1145/2661229.2661271},
    journal = {ACM Trans. Graph.},
    month = nov,
    articleno = {219},
    numpages = {12}
}

@inproceedings{shum2007simulating,
    author = {Shum, Hubert P. H. and Komura, Taku and Yamazaki, Shuntaro},
    title = {Simulating competitive interactions using singly captured motions},
    booktitle = {Proceedings of the 13th {ACM} Symposium on Virtual Reality Software and Technology (VRST)},
    year = {2007},
    isbn = {9781595938633},
    publisher = {Association for Computing Machinery},
    address = {New York, NY, USA},
    url = {https://doi.org/10.1145/1315184.1315194},
    doi = {10.1145/1315184.1315194},
    pages = {65–72},
    numpages = {8},
    location = {Newport Beach, California},
    series = {VRST '07}
}

@article{shum2012simulating,
  author={Shum, Hubert P.H. and Komura, Taku and Yamazaki, Shuntaro},
  journal={IEEE Transactions on Visualization and Computer Graphics}, 
  title={Simulating Multiple Character Interactions with Collaborative and Adversarial Goals}, 
  year={2012},
  volume={18},
  number={5},
  pages={741-752},
  doi={10.1109/TVCG.2010.257}
}

@inproceedings{zhang2023simulation,
    author = {Zhang, Yunbo and Gopinath, Deepak and Ye, Yuting and Hodgins, Jessica and Turk, Greg and Won, Jungdam},
    title = {Simulation and Retargeting of Complex Multi-Character Interactions},
    year = {2023},
    isbn = {9798400701597},
    publisher = {Association for Computing Machinery},
    address = {New York, NY, USA},
    url = {https://doi.org/10.1145/3588432.3591491},
    doi = {10.1145/3588432.3591491},
    booktitle = {ACM SIGGRAPH 2023 Conference Proceedings},
    articleno = {65},
    numpages = {11},
    location = {Los Angeles, CA, USA},
    series = {SIGGRAPH '23}
}

\end{document}